\documentclass[11pt]{article}
\pdfoutput=1

\usepackage{url}
\usepackage{color}
\usepackage[letterpaper, margin=3cm]{geometry}
\usepackage{todonotes}
\usepackage{amsmath,amsfonts,amssymb}
\usepackage{bm}
\usepackage[frozencache]{minted}
\usepackage{algorithm}
\usepackage{algpseudocode}
\usepackage[pdftex,bookmarks=false,colorlinks=false, pdfstartview=FitV,linkcolor=blue,citecolor=blue,urlcolor=blue]{hyperref}

\newcommand{\ignore}[1]{}

\newcommand{\term}[1]{\textbf{#1}}
\newcommand{\argmax}[1]{\underset{#1}{\text{argmax }}}

\newcommand{\secref}[1]{\autoref{sec:#1}}
\newcommand{\figref}[1]{\autoref{fig:#1}}

\renewcommand{\algref}[1]{\autoref{alg:#1}}
\renewcommand{\eqref}[1]{\autoref{eq:#1}}
\newcommand{\newcite}[1]{\cite{#1}}

\newcommand{\sentend}{\langle\text{/s}\rangle}
\newcommand{\sentbegin}{\langle\text{s}\rangle}
\newcommand{\sentunk}{\langle\text{unk}\rangle}

\newcommand{\question}[1]{\footnote{\textit{Question: #1}}}

\newif\iffullbook
\fullbookfalse

\title{Neural Machine Translation and Sequence-to-sequence Models: A Tutorial}
\author{Graham Neubig\\Language Technologies Institute, Carnegie Mellon University}
\date{}

\begin{document}
\maketitle

% \begin{abstract}
% TODO
% \end{abstract}

  \section{Introduction}
  \label{sec:seq2seqintro}
  This tutorial introduces a new and powerful set of techniques variously called ``neural machine translation'' or ``neural sequence-to-sequence models''.
These techniques have been used in a number of tasks regarding the handling of human language, and can be a powerful tool in the toolbox of anyone who wants to model sequential data of some sort.
The tutorial assumes that the reader knows the basics of math and programming, but does not assume any particular experience with neural networks or natural language processing.
It attempts to explain the intuition behind the various methods covered, then delves into them with enough mathematical detail to understand them concretely, and culiminates with a suggestion for an implementation exercise, where readers can test that they understood the content in practice.

\subsection{Background}
\label{sec:intro:background}

Before getting into the details, it might be worth describing each of the terms that appear in the title ``Neural Machine Translation and Sequence-to-sequence Models''.
\term{Machine translation} is the technology used to translate between human language.
Think of the universal translation device showing up in sci-fi movies to allow you to communicate effortlessly with those that speak a different language, or any of the plethora of online translation web sites that you can use to assimilate content that is not in your native language.
This ability to remove language barriers, needless to say, has the potential to be very useful, and thus machine translation technology has been researched from shortly after the advent of digital computing.
\iffullbook
Because of this, machine translation has been a target technology for the use of computers since the advent of the computing era, and \secref{intro:history} describes a bit more about this long history.
\fi

We call the language input to the machine translation system the \term{source language}, and call the output language the \term{target language}.
Thus, machine translation can be described as the task of converting a \textit{sequence} of words in the source, and converting into a \textit{sequence} of words in the target.
The goal of the machine translation practitioner is to come up with an effective model that allows us to perform this conversion accurately over a broad variety of languages and content.

\begin{figure}
 \centering
 \includegraphics[width=10cm]{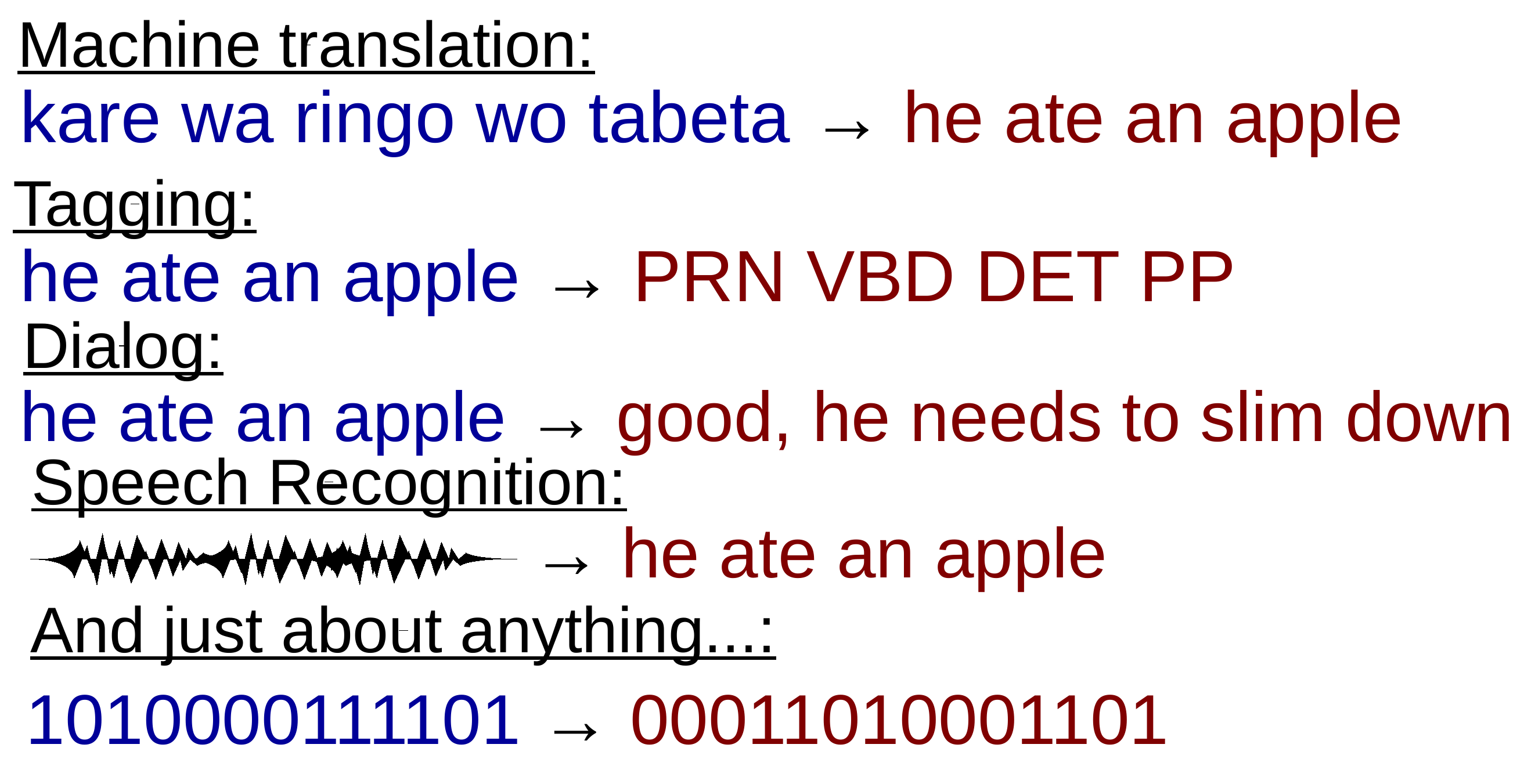}
 \caption{An example of sequence-to-sequence modeling tasks.}
 \label{fig:seq2seqmodels}
\end{figure}

The second part of the title, \term{sequence-to-sequence models}, refers to the broader class of models that include \textit{all} models that map one sequence to another.
This, of course, includes machine translation, but it also covers a broad spectrum of other methods used to handle other tasks as shown in \figref{seq2seqmodels}.
In fact, if we think of a computer program as something that takes in a sequence of input bits, then outputs a sequence of output bits, we could say that \textit{every single program is a sequence-to-sequence model} expressing some behavior (although of course in many cases this is not the most natural or intuitive way to express things).

The motivation for using machine translation as a representative of this larger class of sequence-to-sequence models is many-fold:
\begin{enumerate}
\item Machine translation is a widely-recognized and useful instance of sequence-to-sequence models, and allows us to use many intuitive examples demonstrating the difficulties encountered when trying to tackle these problems.
\item Machine translation is often one of the main driving tasks behind the development of new models, and thus these models tend to be tailored to MT first, then applied to other tasks.
\item However, there are also cases where MT has learned from other tasks as well, and introducing these tasks helps explain the techniques used in MT as well.
\end{enumerate}

\subsection{Structure of this Tutorial}
\label{sec:intro:outline}

This tutorial first starts out with a general mathematical definition of statistical techniques for machine translation in \secref{smt}.
The rest of this tutorial will sequentially describe techniques of increasing complexity, leading up to attentional models, which represent the current state-of-the-art in the field.

First, Sections \ref{sec:ngramlm}-\ref{sec:rnnlm} focus on \term{language models}, which calculate the probability of a target sequence of interest.
These models are not capable of performing translation or sequence transduction, but will provide useful preliminaries to understand sequence-to-sequence models.
\begin{itemize}
\item \secref{ngramlm} describes \term{$n$-gram language models}, simple models that calculate the probability of words based on their counts in a set of data.
It also describes how we evaluate how well these models are doing using measures such as \term{perplexity}.
\item \secref{lllm} describes \term{log-linear language models}, models that instead calculate the probability of the next word based on features of the context.
It describes how we can learn the parameters of the models through \term{stochastic gradient descent} -- calculating derivatives and gradually updating the parameters to increase the likelihood of the observed data.
\item \secref{nnlm} introduces the concept of \term{neural networks}, which allow us to combine together multiple pieces of information more easily than log-linear models, resulting in increased modeling accuracy.
It gives an example of \term{feed-forward neural language models}, which calculate the probability of the next word based on a few previous words using neural networks.
\item \secref{rnnlm} introduces \term{recurrent neural networks}, a variety of neural networks that have mechanisms to allow them to remember information over multiple time steps.
These lead to \term{recurrent neural network language models}, which allow for the handling of long-term dependencies that are useful when modeling language or other sequential data.
\end{itemize}

Finally, Sections \ref{sec:encdec} and \ref{sec:attention} describe actual sequence-to-sequence models capable of performing machine translation or other tasks.
\begin{itemize}
\item \secref{encdec} describes \term{encoder-decoder} models, which use a recurrent neural network to \textit{encode} the target sequence into a vector of numbers, and another network to \textit{decode} this vector of numbers into an output sentence.
It also describes \term{search algorithms} to generate output sequences based on this model.
\item \secref{attention} describes \term{attention}, a method that allows the model to focus on different parts of the input sentence while generating translations.
This allows for a more efficient and intuitive method of representing sentences, and is often more effective than its simpler encoder-decoder counterpart.
\end{itemize}

  \section{Statistical MT Preliminaries}
  \label{sec:smt}
  First, before talking about any specific models, this chapter describes the overall framework of \term{statistical machine translation} (SMT) \cite{brown93smt} more formally.

First, we define our task of machine translation as translating a source sentence $F=f_1,\ldots,f_J=f_1^{|F|}$ into a target sentence $E=e_1,\ldots,e_I=e_1^{|E|}$.%
\iffullbook
\footnote{Note for the time being, we are assuming that we translate each sentence independently, although we will discuss document-level translation in \secref{document}.}
\fi
Thus, any type of translation system can be defined as a function
\begin{equation}
\hat{E} = \text{mt}(F),
\end{equation}
which returns a translation hypothesis $\hat{E}$ given a source sentence $F$ as input.

\term{Statistical machine translation} systems are systems that perform translation by creating a probabilistic model for the probability of $E$ given $F$, $P(E \mid F;\theta)$, and finding the target sentence that maximizes this probability:
\begin{equation}
\hat{E} = \argmax{E} P(E \mid F;\theta),
\end{equation}
where $\theta$ are the parameters of the model specifying the probability distribution.
The parameters $\theta$ are learned from data consisting of aligned sentences in the source and target languages, which are called \term{parallel corpora} in technical terminology.%
\iffullbook
\footnote{Details about data can be found in \secref{data}.}
\fi
 Within this framework, there are three major problems that we need to handle appropriately in order to create a good translation system:
\begin{description}
\item[Modeling:] First, we need to decide what our model $P(E \mid F;\theta)$ will look like.
What parameters will it have, and how will the parameters specify a probability distribution?
\item[Learning:] Next, we need a method to learn appropriate values for parameters $\theta$ from training data.
\item[Search:] Finally, we need to solve the problem of finding the most probable sentence (solving ``argmax'').
This process of searching for the best hypothesis and is often called \term{decoding}.\footnote{This is based on the famous quote from Warren Weaver, likening the process of machine translation to decoding an encoded cipher.}
\end{description}
The remainder of the material here will focus on solving these problems.

  \section{$n$-gram Language Models}
  \label{sec:ngramlm}

While the final goal of a statistical machine translation system is to create a model of the target sentence $E$ given the source sentence $F$, $P(E \mid F)$, in this chapter we will take a step back, and attempt to create a \term{language model} of only the target sentence $P(E)$.
Basically, this model allows us to do two things that are of practical use.
\begin{description}
\item[Assess naturalness:]
Given a sentence $E$, this can tell us, does this look like an actual, natural sentence in the target language?
If we can learn a model to tell us this, we can use it to assess the fluency of sentences generated by an automated system to improve its results.
It could also be used to evaluate sentences generated by a human for purposes of grammar checking or error correction.
\item[Generate text:]
Language models can also be used to randomly generate text by sampling a sentence $E'$ from the target distribution: $E' \sim P(E)$.%
\footnote{$\sim$ means ``is sampled from''.}
Randomly generating samples from a language model can be interesting in itself -- we can see what the model ``thinks'' is a natural-looking sentences -- but it will be more practically useful in the context of the neural translation models described in the following chapters.
\end{description}

In the following sections, we'll cover a few methods used to calculate this probability $P(E)$.

\subsection{Word-by-word Computation of Probabilities}
\label{sec:ngramlm:wordbyword}

As mentioned above, we are interested in calculating the probability of a sentence $E=e_1^T$.
Formally, this can be expressed as
\begin{equation}
P(E) = P(|E|=T,e_1^T),
\label{eq:ngramlm:joint}
\end{equation}
the joint probability that the length of the sentence is ($|E|=T$), that the identity of the first word in the sentence is $e_1$, the identity of the second word in the sentence is $e_2$, up until the last word in the sentence being $e_T$.
Unfortunately, directly creating a model of this probability distribution is not straightforward,\footnote{Although it is possible, as shown by \term{whole-sentence language models} in \cite{rosenfeld01wholesentence}.} as the length of the sequence $T$ is not determined in advance, and there are a large number of possible combinations of words.\question{If $V$ is the size of the target vocabulary, how many are there for a sentence of length $T$?}

\begin{figure}[h]
 \centering
 \includegraphics[width=8.5cm]{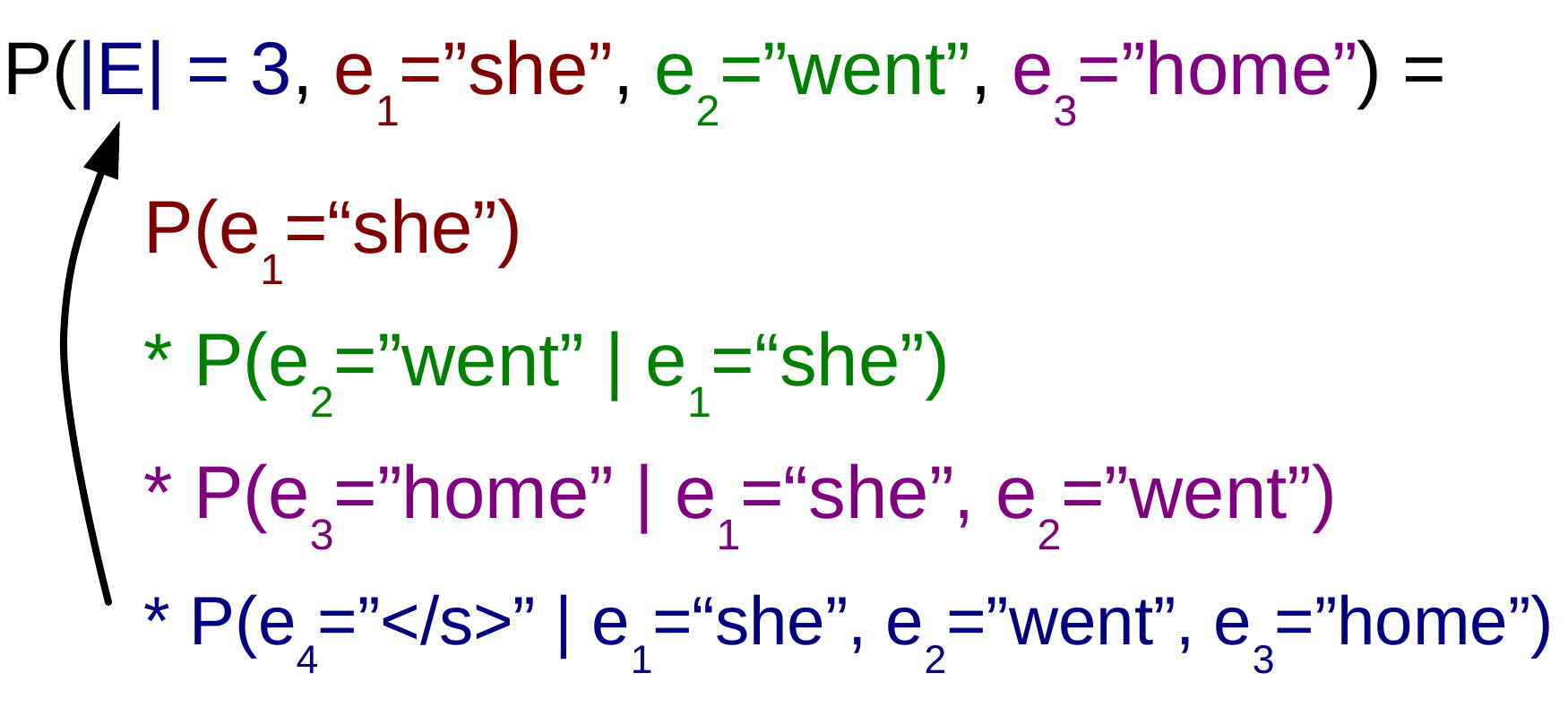}
 \caption{An example of decomposing language model probabilities word-by-word.}
 \label{fig:ngramlm:wordbyword}
\end{figure}

As a way to make things easier, it is common to re-write the probability of the full sentence as the product of single-word probabilities.
This takes advantage of the fact that a joint probability -- for example $P(e_1,e_2,e_3)$ -- can be calculated by multiplying together conditional probabilities for each of its elements.
In the example, this means that $P(e_1,e_2,e_3) = P(e_1) P(e_2 \mid e_1) P(e_3 \mid e_1,e_2)$.

\figref{ngramlm:wordbyword} shows an example of this incremental calculation of probabilities for the sentence ``she went home''.
Here, in addition to the actual words in the sentence, we have introduced an implicit \textit{sentence end} (``$\sentend$'') symbol, which we will indicate when we have terminated the sentence.
Stepping through the equation in order, this means we first calculate the probability of ``she'' coming at the beginning of the sentence, then the probability of ``went'' coming next in a sentence starting with ``she'', the probability of ``home'' coming after the sentence prefix ``she went'', and then finally the sentence end symbol ``$\sentend$'' after ``she went home''.
More generally, we can express this as the following equation:
\begin{equation}
P(E) = \prod_{t=1}^{T+1} P(e_t \mid e_1^{t-1})
\label{eq:ngramlm:prod}
\end{equation}
where $e_{T+1}=\sentend$.
So coming back to the sentence end symbol $\sentend$, the reason why we introduce this symbol is because it allows us to know when the sentence ends.
In other words, by examining the position of the $\sentend$ symbol, we can determine the $|E|=T$ term in our original LM joint probability in \eqref{ngramlm:joint}.
In this example, when we have $\sentend$ as the 4th word in the sentence, we know we're done and our final sentence length is 3.

Once we have the formulation in \eqref{ngramlm:prod}, the problem of language modeling now becomes a problem of calculating the next word given the previous words $P(e_t \mid e_1^{t-1})$.
This is much more manageable than calculating the probability for the whole sentence, as we now have a fixed set of items that we are looking to calculate probabilities for.
The next couple of sections will show a few ways to do so.

\subsection{Count-based $n$-gram Language Models}
\label{sec:ngramlm:countbased}

\begin{figure}
 \centering
 \includegraphics[width=10cm]{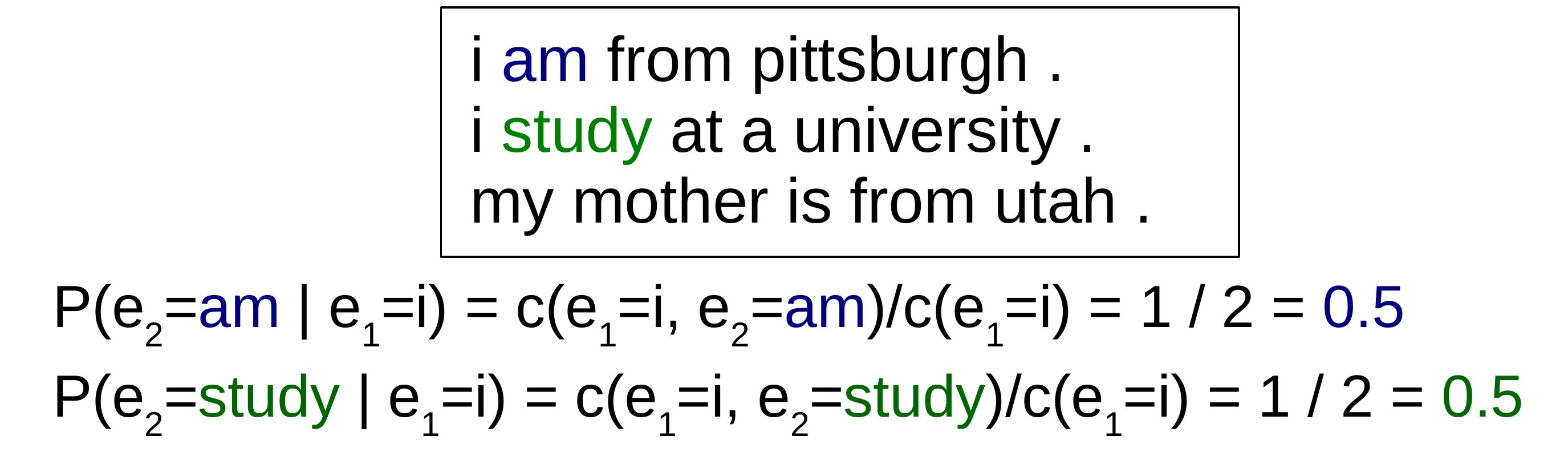}
 \caption{An example of calculating probabilities using maximum likelihood estimation.}
 \label{fig:ngramlm:countbasedmle}
\end{figure}

The first way to calculate probabilities is simple: prepare a set of training data from which we can count word strings, count up the number of times we have seen a particular string of words, and divide it by the number of times we have seen the context.
This simple method, can be expressed by the equation below, with an example shown in \figref{ngramlm:countbasedmle}
\begin{equation}
P_{\text{ML}}(e_t \mid e_1^{t-1}) = \frac{c_{\text{prefix}}(e_1^t)}{c_{\text{prefix}}(e_1^{t-1})}.
\label{eq:ngramlm:wordbyword}
\end{equation}
Here $c_{\text{prefix}}(\cdot)$ is the count of the number of times this particular word string appeared at the beginning of a sentence in the training data.
This approach is called \term{maximum likelihood estimation} (MLE, details later in this chapter), and is both simple and guaranteed to create a model that assigns a high probability to the sentences in training data.

However, let's say we want to use this model to assign a probability to a new sentence that we've never seen before.
For example, if we want to calculate the probability of the sentence ``i am from utah .'' based on the training data in the example.
This sentence is extremely similar to the sentences we've seen before, but unfortunately because the string ``i am from utah'' has not been observed in our training data, $c_{\text{prefix}}(\text{i},\text{am},\text{from},\text{utah})=0$, $P(e_4=\text{utah} \mid e_1=\text{i},e_2=\text{am},e_3=\text{from})$ becomes zero, and thus the probability of the whole sentence as calculated by \eqref{ngramlm:wordbyword} also becomes zero.
In fact, this language model will assign a probability of zero to every sentence that it hasn't seen before in the training corpus, which is not very useful, as the model loses ability to tell us whether a new sentence a system generates is natural or not, or generate new outputs.

To solve this problem, we take two measures.
First, instead of calculating probabilities from the beginning of the sentence, we set a fixed window of previous words upon which we will base our probability calculations, approximating the true probability.
If we limit our context to $n-1$ previous words, this would amount to:
\begin{equation}
P(e_t \mid e_1^{t-1}) \approx P_{\text{ML}}(e_t \mid e_{t-n+1}^{t-1}).
\end{equation}
Models that make this assumption are called \term{$n$-gram models}.
Specifically, when models where $n=1$ are called unigram models, $n=2$ bigram models, $n=3$ trigram models, and $n \ge 4$ four-gram, five-gram, etc.

The parameters $\theta$ of $n$-gram models consist of probabilities of the next word given $n-1$ previous words:
\begin{equation}
\theta_{e_{t-n+1}^{t}} = P(e_t \mid e_{t-n+1}^{t-1}),
\end{equation}
and in order to train an $n$-gram model, we have to learn these parameters from data.%
\question{How many parameters does an $n$-gram model with a particular $n$ have?}
In the simplest form, these parameters can be calculated using maximum likelihood estimation as follows:
\begin{equation}
\theta_{e_{t-n+1}^{t}} = P_{\text{ML}}(e_t \mid e_{t-n+1}^{t-1}) = \frac{c(e_{t-n+1}^t)}{c(e_{t-n+1}^{t-1})},
\label{eq:ngramlm:mle}
\end{equation}
where $c(\cdot)$ is the count of the word string anywhere in the corpus.
Sometimes these equations will reference $e_{t-n+1}$ where $t-n+1<0$. In this case, we assume that $e_{t-n+1}=\sentbegin$ where $\sentbegin$ is a special \textit{sentence start} symbol.

If we go back to our previous example and set $n=2$, we can see that while the string ``i am from utah .'' has never appeared in the training corpus, ``i am'', ``am from'', ``from utah'', ``utah .'', and ``. $\sentend$'' are all somewhere in the training corpus, and thus we can patch together probabilities for them and calculate a non-zero probability for the whole sentence.%
\question{What is this probability?}

However, we still have a problem: what if we encounter a two-word string that has never appeared in the training corpus?
In this case, we'll still get a zero probability for that particular two-word string, resulting in our full sentence probability also becoming zero.
$n$-gram models fix this problem by \term{smoothing} probabilities, combining the maximum likelihood estimates for various values of $n$.
In the simple case of smoothing unigram and bigram probabilities, we can think of a model that combines together the probabilities as follows:
\begin{equation}
P(e_t \mid e_{t-1}) = (1 - \alpha) P_{\text{ML}}(e_t \mid e_{t-1}) + \alpha P_{\text{ML}}(e_t),
\label{eq:ngramlm:interpolation}
\end{equation}
where $\alpha$ is a variable specifying how much probability mass we hold out for the unigram distribution.
As long as we set $\alpha > 0$, regardless of the context all the words in our vocabulary will be assigned some probability.
This method is called \term{interpolation}, and is one of the standard ways to make probabilistic models more robust to low-frequency phenomena.

If we want to use even more context -- $n=3$, $n=4$, $n=5$, or more -- we can recursively define our interpolated probabilities as follows:
\begin{equation}
P(e_t \mid e_{t-m+1}^{t-1}) = (1 - \alpha_m) P_{\text{ML}}(e_t \mid e_{t-m+1}^{t-1}) + \alpha_m P(e_t \mid e_{t-m+2}^{t-1}).
\label{eq:ngramlm:recursiveinterpolation}
\end{equation}
The first term on the right side of the equation is the maximum likelihood estimate for the model of order $m$, and the second term is the interpolated probability for all orders up to $m-1$.

There are also more sophisticated methods for smoothing, which are beyond the scope of this section, but summarized very nicely in \cite{chen96smoothing}.
\begin{description}
\item[Context-dependent smoothing coefficients:]
Instead of having a fixed $\alpha$, we condition the interpolation coefficient on the context: $\alpha_{e_{t-m+1}^{t-1}}$.
This allows the model to give more weight to higher order $n$-grams when there are a sufficient number of training examples for the parameters to be estimated accurately and fall back to lower-order $n$-grams when there are fewer training examples.
These context-dependent smoothing coefficients can be chosen using heuristics \cite{witten91zero} or learned from data \cite{neubig16emnlp}.
\item[Back-off:]
In \eqref{ngramlm:interpolation}, we interpolated together two probability distributions over the full vocabulary $V$.
In the alternative formulation of \term{back-off}, the lower-order distribution only is used to calculate probabilities for words that were given a probability of zero in the higher-order distribution.
Back-off is more expressive but also more complicated than interpolation, and the two have been reported to give similar results \cite{goodman01abitofprogress}.
\item[Modified distributions:]
It is also possible to use a different distribution than $P_{\text{ML}}$.
This can be done by subtracting a constant value from the counts before calculating probabilities, a method called \term{discounting}.
It is also possible to modify the counts of lower-order distributions to reflect the fact that they are used mainly as a fall-back for when the higher-order distributions lack sufficient coverage.
\end{description}
Currently, \term{Modified Kneser-Ney smoothing} (MKN; \cite{chen96smoothing}), is generally considered one of the standard and effective methods for smoothing $n$-gram language models.
MKN uses context-dependent smoothing coefficients, discounting, and modification of lower-order distributions to ensure accurate probability estimates.

\subsection{Evaluation of Language Models}
\label{sec:ngramlm:eval}

Once we have a language model, we will want to test whether it is working properly.
The way we test language models is, like many other machine learning models, by preparing three sets of data:
\begin{description}
\item[Training data] is used to train the parameters $\theta$ of the model.
\item[Development data] is used to make choices between alternate models, or to tune the \term{hyper-parameters} of the model. Hyper-parameters in the model above could include the maximum length of $n$ in the $n$-gram model or the type of smoothing method.
\item[Test data] is used to measure our final accuracy and report results.
\end{description}

For language models, we basically want to know whether the model is an accurate model of language, and there are a number of ways we can define this.
The most straight-forward way of defining accuracy is the \term{likelihood} of the model with respect to the development or test data.
The likelihood of the parameters $\theta$ with respect to this data is equal to the probability that the model assigns to the data.
For example, if we have a test dataset $\mathcal{E}_{\text{test}}$, this is:
\begin{equation}
P(\mathcal{E}_{\text{test}};\theta).
\label{eq:ngramlm:likelihood}
\end{equation}
We often assume that this data consists of several independent sentences or documents $E$, giving us
\begin{equation}
P(\mathcal{E}_{\text{test}};\theta) = \prod_{E \in \mathcal{E}_{\text{test}}} P(E;\theta).
\end{equation}

Another measure that is commonly used is \term{log likelihood}
\begin{equation}
\label{eq:ngramlm:loglikelihood}
\log P(\mathcal{E}_{\text{test}};\theta) = \sum_{E \in \mathcal{E}_{\text{test}}} \log P(E;\theta).
\end{equation}
The log likelihood is used for a couple reasons.
The first is because the probability of any particular sentence according to the language model can be a very small number, and the product of these small numbers can become a very small number that will cause numerical precision problems on standard computing hardware.
The second is because sometimes it is more convenient mathematically to deal in log space.
For example, when taking the derivative in gradient-based methods to optimize parameters (used in the next section), it is more convenient to deal with the sum in \eqref{ngramlm:loglikelihood} than the product in \eqref{ngramlm:likelihood}.

It is also common to divide the log likelihood by the number of words in the corpus
\begin{equation}
\text{length}(\mathcal{E}_{\text{test}}) = \sum_{E \in \mathcal{E}_{\text{test}}} |E|.
\end{equation}
This makes it easier to compare and contrast results across corpora of different lengths.

The final common measure of language model accuracy is \term{perplexity}, which is defined as the exponent of the average negative log likelihood per word
\begin{equation}
\text{ppl}(\mathcal{E}_{\text{test}};\theta) = e^{- (\log P(\mathcal{E}_{\text{test}};\theta)) / \text{length}(\mathcal{E}_{\text{test}})}.
\label{eq:ngramlm:perplexity}
\end{equation}
An intuitive explanation of the perplexity is ``how confused is the model about its decision?''
More accurately, it expresses the value ``if we randomly picked words from the probability distribution calculated by the language model at each time step, on average how many words would it have to pick to get the correct one?''
One reason why it is common to see perplexities in research papers is because the numbers calculated by perplexity are bigger, making the differences in models more easily perceptible by the human eye.%
\footnote{And, some cynics will say, making it easier for your research papers to get accepted.}

\subsection{Handling Unknown Words}
\label{sec:ngramlm:unk}

Finally, one important point to keep in mind is that some of the words in the test set $\mathcal{E}_{\text{test}}$ will not appear even once in the training set $\mathcal{E}_{\text{train}}$.
These words are called \term{unknown words}, and need to be handeled in some way.
Common ways to do this in language models include:
\begin{description}
\item[Assume closed vocabulary:]
Sometimes we can assume that there will be no new words in the test set.
For example, if we are calculating a language model over ASCII characters, it is reasonable to assume that all characters have been observed in the training set.
Similarly, in some speech recognition systems, it is common to simply assign a probability of zero to words that don't appear in the training data, which means that these words will not be able to be recognized.
\item[Interpolate with an unknown words distribution:]
As mentioned in \eqref{ngramlm:recursiveinterpolation}, we can interpolate between distributions of higher and lower order.
In the case of unknown words, we can think of this as a distribution of order ``0'', and define the 1-gram probability as the interpolation between the unigram distribution and unknown word distribution
\begin{equation}
P(e_t) = (1 - \alpha_1) P_{\text{ML}}(e_t) + \alpha_1 P_{\text{\text{unk}}}(e_t).
\label{eq:ngramlm:unkinterpolation}
\end{equation}
Here, $P_{\text{\text{unk}}}$ needs to be a distribution that assigns a probability to all words $V_{\text{all}}$, not just ones in our vocabulary $V$ derived from the training corpus.
This could be done by, for example, training a language model over characters that ``spells out'' unknown words in the case they don't exist in in our vocabulary.
Alternatively, as a simpler approximation that is nonetheless fairer than ignoring unknown words, we can guess the total number of words $|V_{\text{all}}|$ in the language where we are modeling, where $|V_{\text{all}}| > |V|$, and define $P_{\text{\text{unk}}}$ as a uniform distribution over this vocabulary: $P_{\text{\text{unk}}}(e_t) = 1/|V_{\text{all}}|$.
\item[Add an $\sentunk$ word:]
As a final method to handle unknown words we can remove some of the words in $\mathcal{E}_{\text{train}}$ from our vocabulary, and replace them with a special $\sentunk$ symbol representing unknown words.
One common way to do so is to remove \term{singletons}, or words that only appear once in the training corpus.
By doing this, we explicitly predict in which contexts we will be seeing an unknown word, instead of implicitly predicting it through interpolation like mentioned above.
Even if we predict the $\sentunk$ symbol, we will still need to estimate the probability of the actual word, so any time we predict $\sentunk$ at position $i$, we further multiply in the probability of $P_{\text{unk}}(e_t)$.
\end{description}

\subsection{Further Reading}
\label{sec:ngramlm:further}

To read in more detail about $n$-gram language models, \cite{goodman01abitofprogress} gives a very nice introduction and comprehensive summary about a number of methods to overcome various shortcomings of vanilla $n$-grams like the ones mentioned above.

There are also a number of extensions to $n$-gram models that may be nice for the interested reader.
\begin{description}
\item[Large-scale language modeling:]
Language models are an integral part of many commercial applications, and in these applications it is common to build language models using massive amounts of data harvested from the web for other sources.
To handle this data, there is research on efficient data structures \cite{heafield11kenlm,pauls11fasterandsmaller}, distributed parameter servers \cite{brants07largelanguagemodels}, and lossy compression algorithms \cite{talbot08randlm}.
\item[Language model adaptation:]
In many situations, we want to build a language model for specific speaker or domain.
Adaptation techniques make it possible to create large general-purpose models, then adapt these models to more closely match the target use case \cite{bellegarda04statistical}.
\item[Longer-distance language count-based models:]
As mentioned above, $n$-gram models limit their context to $n-1$, but in reality there are dependencies in language that can reach much farther back into the sentence, or even span across whole documents.
The recurrent neural network language models that we will introduce in \secref{rnnlm} are one way to handle this problem, but there are also non-neural approaches such as cache language models \cite{kuhn1990cache}, topic models \cite{blei03lda}, and skip-gram models \cite{goodman01abitofprogress}.
\item[Syntax-based language models:]
There are also models that take into account the syntax of the target sentence.
For example, it is possible to condition probabilities not on words that occur directly next to each other in the sentence, but those that are ``close'' syntactically \cite{shen08stringtodependency}.
\end{description}

\subsection{Exercise}
\label{sec:ngramlm:exercise}

The exercise that we will be doing in class will be constructing an $n$-gram LM with linear interpolation between various levels of $n$-grams.
We will write code to:
\begin{itemize}
\item Read in and save the training and testing corpora.
\item Learn the parameters on the training corpus by counting up the number of times each $n$-gram has been seen, and calculating maximum likelihood estimates according to \eqref{ngramlm:mle}.
\item Calculate the probabilities of the test corpus using linearly interpolation according to \eqref{ngramlm:interpolation} or \eqref{ngramlm:recursiveinterpolation}.
\end{itemize}
To handle unknown words, you can use the \textit{uniform distribution} method described in \secref{ngramlm:unk}, assuming that there are 10,000,000 words in the English vocabulary.
As a sanity check, it may be better to report the number of unknown words, and which portions of the per-word log-likelihood were incurred by the main model, and which portion was incurred by the unknown word probability $\log P_{\text{unk}}$.

In order to do so, you will first need data, and to make it easier to start out you can use some pre-processed data from the German-English translation task of the IWSLT evaluation campaign\footnote{\url{http://iwslt.org}} here: \url{http://phontron.com/data/iwslt-en-de-preprocessed.tar.gz}.

Potential improvements to the model include reading \cite{chen96smoothing} and implementing a better smoothing method, implementing a better method for handling unknown words, or implementing one of the more advanced methods in \secref{ngramlm:further}.

  \section{Log-linear Language Models}
  \label{sec:lllm}
  
This chapter will discuss another set of language models: \term{log-linear language models} \cite{rosenfeld96melm,chen00melm}, which take a very different approach than the count-based $n$-grams described above.%
\footnote{It should be noted that the cited papers call these \term{maximum entropy language models}.
This is because models in this chapter can be motivated in two ways:
\textit{log-linear models} that calculate un-normalized log-probability scores for each function and normalize them to probabilities,
and \textit{maximum-entropy models} that spread their probability mass as evenly as possible given the constraint that they must model the training data.
While the maximum-entropy interpretation is quite interesting theoretically and interested readers can reference \cite{berger96maxent} to learn more, the explanation as log-linear models is simpler conceptually, and thus we will use this description in this chapter.}

\subsection{Model Formulation}
\label{sec:lllm:formulation}

Like $n$-gram language models, log-linear language models still calculate the probability of a particular word $e_t$ given a particular context $e_{t-n+1}^{t-1}$.
However, their method for doing so is quite different from count-based language models, based on the following procedure.

\textbf{Calculating features:} Log-linear language models revolve around the concept of \term{features}.
In short, features are basically, ``something about the context that will be useful in predicting the next word''.
More formally, we define a feature function $\bm{\phi}(e_{t-n+1}^{t-1})$ that takes a context as input, and outputs a real-valued \term{feature vector} $\bm{x} \in \mathbb{R}^N$ that describe the context using $N$ different features.%
\footnote{Alternative formulations that define feature functions that also take the current word as input $\bm{\phi}(e_{t-n+1}^{t})$ are also possible, but in this book, to simplify the transition into neural language models described in \secref{nnlm}, we consider features over only the context.}

\begin{figure}[h]
 \centering
 \includegraphics[width=10cm]{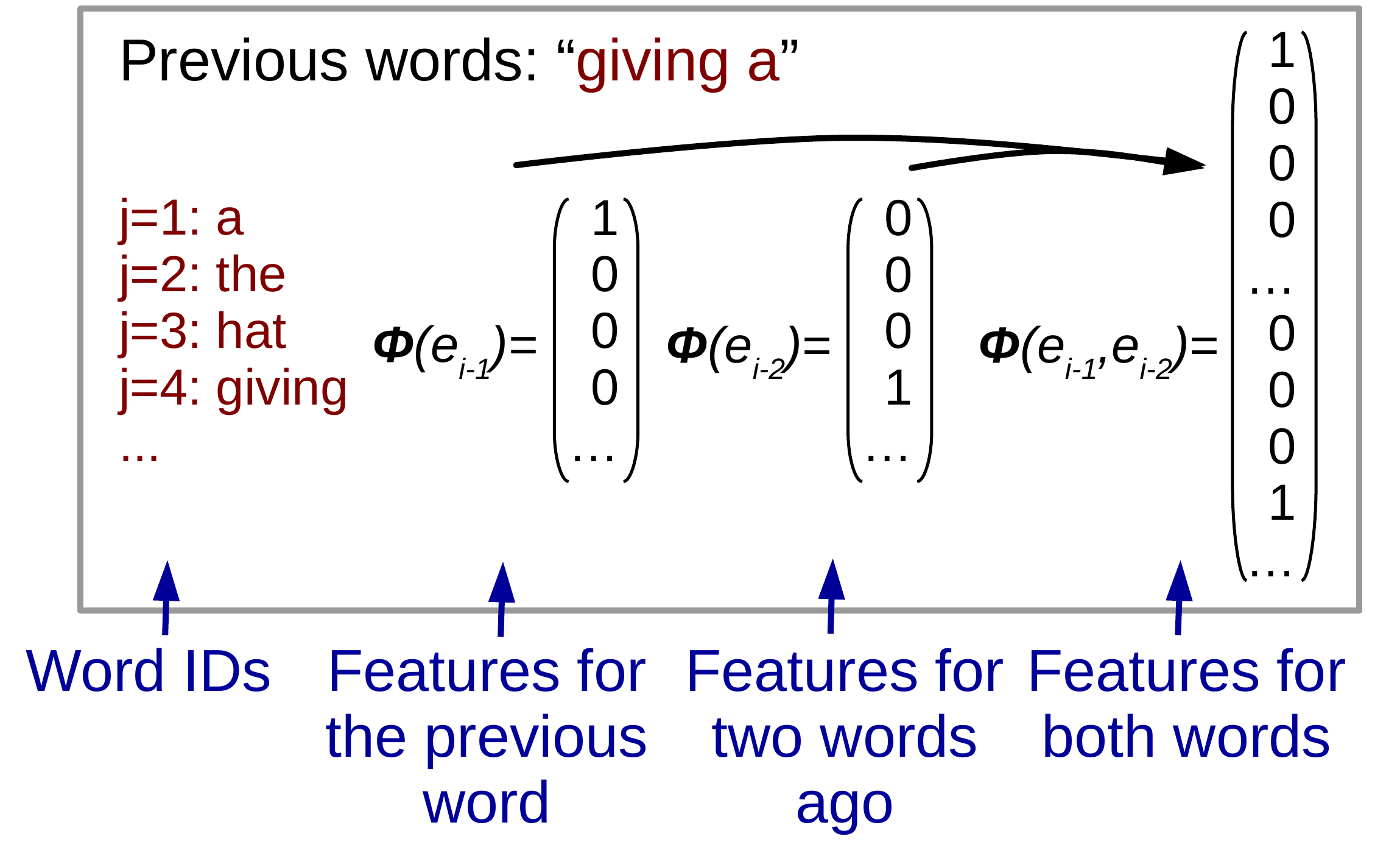}
 \caption{An example of feature values for a particular context.}
 \label{fig:lllm:features}
\end{figure}

For example, from our bi-gram models from the previous chapter, we know that ``the identity of the previous word'' is something that is useful in predicting the next word.
If we want to express the identity of the previous word as a real-valued vector, we can assume that each word in our vocabulary $V$ is associated with a word ID $j$, where $1 \le j \le |V|$.
Then, we define our feature function $\bm{\phi}(e_{t-n+1}^{t})$ to return a feature vector $\bm{x} = \mathbb{R}^{|V|}$, where if $e_{t-1}=j$, then the $j$th element is equal to one and the remaining elements in the vector are equal to zero.
This type of vector is often called a \term{one-hot vector}, an example of which is shown in \figref{lllm:features}(a).
For later user, we will also define a function $\text{onehot}(i)$ which returns a vector where only the $i$th element is one and the rest are zero (assume the length of the vector is the appropriate length given the context).

Of course, we are not limited to only considering one previous word.
We could also calculate one-hot vectors for both $e_{t-1}$ and $e_{t-2}$, then concatenate them together, which would allow us to create a model that considers the values of the two previous words.
In fact, there are many other types of feature functions that we can think of (more in \secref{lllm:features}), and the ability to flexibly define these features is one of the advantages of log-linear language models over standard $n$-gram models.

\textbf{Calculating scores:}
Once we have our feature vector, we now want to use these features to predict probabilities over our output vocabulary $V$.
In order to do so, we calculate a score vector $\bm{s} \in \mathbb{R}^{|V|}$ that corresponds to the likelihood of each word: words with higher scores in the vector will also have higher probabilities.
We do so using the model parameters $\theta$, which specifically come in two varieties: a \term{bias vector} $\bm{b} \in \mathbb{R}^{|V|}$, which tells us how likely each word in the vocabulary is overall, and a \term{weight matrix} $W=\mathbb{R}^{|V| \times N}$ which describes the relationship between feature values and scores. 
Thus, the final equation for calculating our scores for a particular context is:
\begin{equation}
\bm{s} = W \bm{x} + \bm{b}.
\label{eq:lllm:fullequation}
\end{equation}

\begin{figure}
 \centering
 \includegraphics[width=11cm]{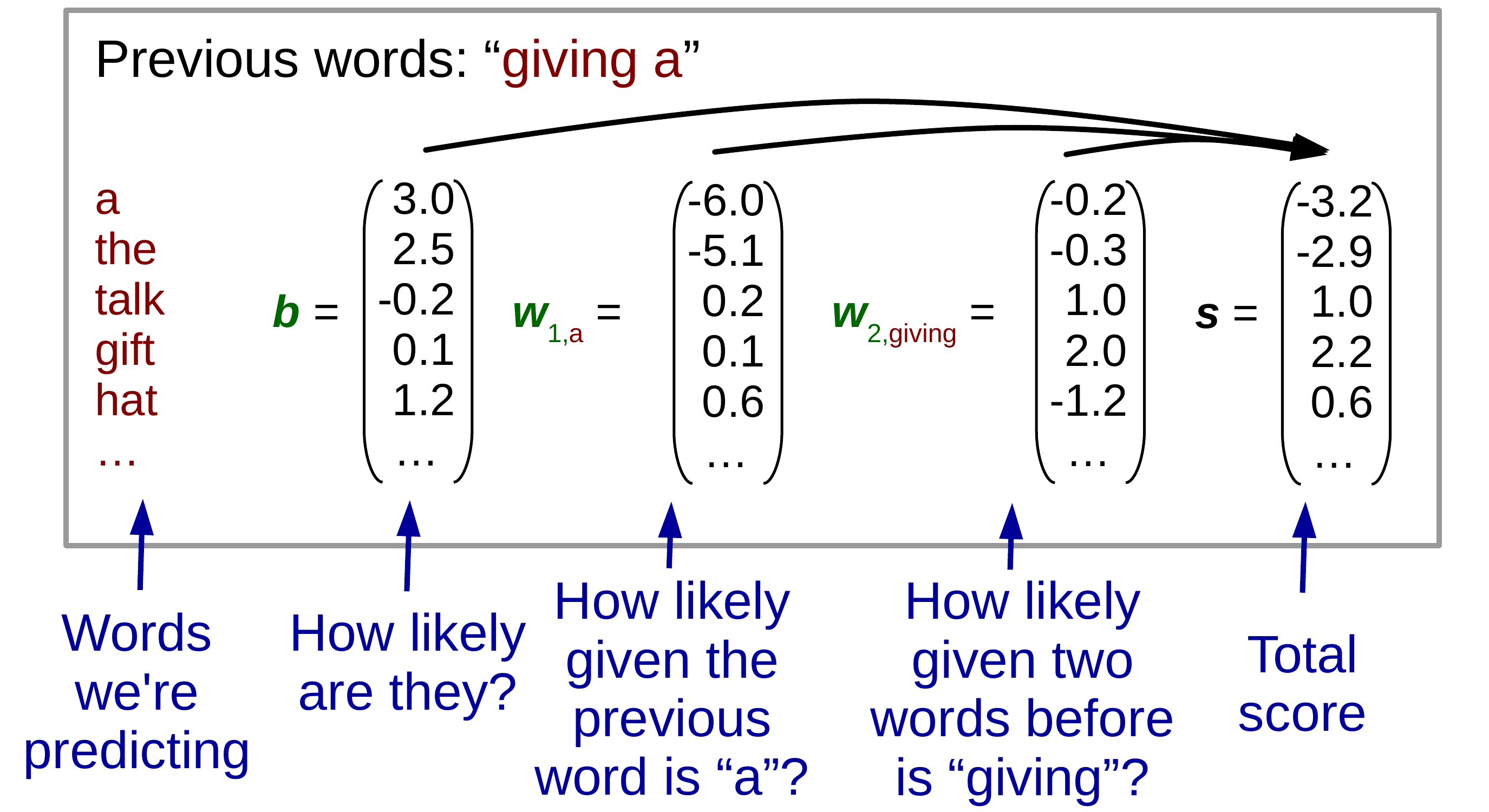}
 \caption{An example of the weights for a log linear model in a certain context.}
 \label{fig:lllm:loglinear}
\end{figure}

One thing to note here is that in the special case of one-hot vectors or other \textit{sparse} vectors where most of the elements are zero.
Because of this we can also think about \eqref{lllm:fullequation} in a different way that is numerically equivalent, but can make computation more efficient.
Specifically, instead of multiplying the large feature vector by the large weight matrix, we can add together the columns of the weight matrix for all \textit{active} (non-zero) features as follows:
\begin{equation}
\bm{s} = \sum_{\{j:x_j \ne 0\}} W_{\cdot,j} x_j + \bm{b},
\end{equation}
where $W_{\cdot,j}$ is the $j$th column of $W$.
This allows us to think of calculating scores as ``look up the vector for the features active for this instance, and add them together'', instead of writing them as matrix math.
An example calculation in this paradigm where we have two feature functions (one for the directly preceding word, and one for the word before that) is shown in \figref{lllm:loglinear}.

\textbf{Calculating probabilities:} 
It should be noted here that scores $\bm{s}$ are arbitrary real numbers, not probabilities: they can be negative or greater than one, and there is no restriction that they add to one.
Because of this, we run these scores through a function that performs the following transformation:
\begin{equation}
p_j = \frac{\text{exp}( s_j )}{\sum_{\tilde{j}} \text{exp}( s_{\tilde{j}} )}.
\label{eq:lllm:softmaxsum}
\end{equation}
By taking the exponent and dividing by the sum of the values over the entire vocabulary, these scores can be turned into probabilities that are between 0 and 1 and sum to 1.

This function is called the \term{softmax} function, and often expressed in vector form as follows:
\begin{equation}
\bm{p} = \text{softmax}(\bm{s}).
\end{equation}
Through applying this to the scores calculated in the previous section, we now have a way to go from features to language model probabilities.

\subsection{Learning Model Parameters}
\label{sec:lllm:learning}

Now, the only remaining missing link is how to acquire the parameters $\theta$, consisting of the weight matrix $W$ and bias $\bm{b}$.
Basically, the way we do so is by attempting to find parameters that fit the training corpus well.

To do so, we use standard machine learning methods for optimizing parameters.
First, we define a \term{loss function} $\ell(\cdot)$ -- a function expressing how poorly we're doing on the training data.
In most cases, we assume that this loss is equal to the \term{negative log likelihood}:
\begin{equation}
\ell(\mathcal{E}_{train},\bm{\theta}) = - \log P(\mathcal{E}_{train} \mid \bm{\theta}) = - \sum_{E \in \mathcal{E}_{train}} \log P(E \mid \bm{\theta}).
\end{equation}
We assume we can also define the loss on a per-word level:
\begin{equation}
\ell(e_{t-n+1}^t,\bm{\theta}) = \log P(e_t \mid e_{t-n+1}^{t-1}).
\end{equation}

Next, we optimize the parameters to reduce this loss.
While there are many methods for doing so, in recent years one of the go-to methods is \term{stochastic gradient descent} (SGD).
SGD is an iterative process where we randomly pick a single word $e_{t}$ (or mini-batch, discussed in \secref{nnlm}) and take a step to improve the likelihood with respect to $e_t$.
In order to do so, we first calculate the derivative of the loss with respect to each of the features in the full feature set $\bm{\theta}$:
\begin{equation}
\frac{d\ell(e_{t-n+1}^t,\bm{\theta})}{d\bm{\theta}}.
\end{equation}
We can then use this information to take a step in the direction that will reduce the loss according to the objective function
\begin{equation}
\bm{\theta} \leftarrow \bm{\theta} - \eta \frac{d\ell(e_{t-n+1}^t,\bm{\theta})}{d\bm{\theta}},
\end{equation}
where $\eta$ is our \term{learning rate}, specifying the amount with which we update the parameters every time we perform an update.
By doing so, we can find parameters for our model that reduce the loss, or increase the likelihood, on the training data.

This vanilla variety of SGD is quite simple and still a very competitive method for optimization in large-scale systems.
However, there are also a few things to consider to ensure that training remains stable:
\begin{description}
\item[Adjusting the learning rate:] SGD requires also requires us to carefully choose $\eta$: if $\eta$ is too big, training can become unstable and diverge, and if $\eta$ is too small, training may become incredibly slow or fall into bad local optima.
One way to handle this problem is \term{learning rate decay}: starting with a higher learning rate, then gradually reducing the learning rate near the end of training.
Other more sophisticated methods are listed below.
\item[Early stopping:]
It is common to use a held-out development set, measure our log-likelihood on this set, and save the model that has achieved the best log-likelihood on this held-out set.
This is useful in case the model starts to over-fit to the training set, losing its generalization capability, we can re-wind to this saved model.
As another method to prevent over-fitting and smooth convergence of training, it is common to measure log likelihood on a held-out development set, and when the log likelihood stops improving or starts getting worse, reduce the learning rate.
\item[Shuffling training order:]
One of the features of SGD is that it processes training data one at a time.
This is nice because it is simple and can be efficient, but it also causes problems if there is some bias in the order in which we see the data.
For example, if our data is a corpus of news text where news articles come first, then sports, then entertainment, there is a chance that near the end of training our model will see hundreds or thousands of entertainment examples in a row, resulting in the parameters moving to a space that favors these more recently seen training examples.
To prevent this problem, it is common (and highly recommended) to randomly shuffle the order with which the training data is presented to the learning algorithm on every pass through the data.
\end{description}

There are also a number of other update rules that have been proposed to improve gradient descent and make it more stable or efficient.
Some representative methods are listed below:
\begin{description}
\item[SGD with momentum \cite{rumelhart1986learning}:]
Instead of taking a single step in the direction of the current gradient, SGD with momentum keeps an exponentially decaying average of past gradients.
This reduces the propensity of simple SGD to ``jitter'' around, making optimization move more smoothly across the parameter space.
\item[AdaGrad \cite{duchi11adagrad}:]
AdaGrad focuses on the fact that some parameters are updated much more frequently than others.
For example, in the model above, columns of the weight matrix $W$ corresponding to infrequent context words will only be updated a few times for every pass through the corpus, while the bias $\bm{b}$ will be updated on every training example.
Based on this, AdaGrad dynamically adjusts the training rate $\eta$ for each parameter individually, with frequently updated (and presumably more stable) parameters such as $\bm{b}$ getting smaller updates, and infrequently updated parameters such as $W$ getting larger updates.
\item[Adam \cite{kingma14adam}:]
Adam is another method that computes learning rates for each parameter.
It does so by keeping track of exponentially decaying averages of the mean and variance of past gradients, incorporating ideas similar to both momentum and AdaGrad.
Adam is now one of the more popular methods for optimization, as it greatly speeds up convergence on a wide variety of datasets, facilitating fast experimental cycles.
However, it is also known to be prone to over-fitting, and thus, if high performance is paramount, it should be used with some caution and compared to more standard SGD methods.
\end{description}
\cite{ruder16overview} provides a good overview of these various methods with equations and notes a few other concerns when performing stochastic optimization.

\subsection{Derivatives for Log-linear Models}
\label{sec:lllm:derivatives}

Now, the final piece in the puzzle is the calculation of derivatives of the loss function with respect to the parameters.
To do so, first we step through the full loss function in one pass as below:
\begin{align}
\bm{x} & = \bm{\phi}(e_{t-m+1}^{t-1}) \\
\bm{s} & = \sum_{\{j:x_j != 0\}} W_{\cdot,j} x_j + \bm{b} \\
\bm{p} & = \text{softmax}( \bm{s} ) \\
\ell & = - \log \bm{p}_{e_t}.
\end{align}
And thus, using the chain rule to calculate
\begin{align}
\frac{d\ell(e_{t-n+1}^t, W, \bm{b})}{d\bm{b}} & = \frac{d\ell}{d\bm{p}} \frac{d\bm{p}}{d\bm{s}} \frac{d\bm{s}}{d\bm{b}} \\
\frac{d\ell(e_{t-n+1}^t, W, \bm{b})}{dW_{\cdot,j}} & = \frac{d\ell}{d\bm{p}} \frac{d\bm{p}}{d\bm{s}} \frac{d\bm{s}}{dW_{\cdot,j}}
\end{align}
we find that the derivative of the loss function for the bias and each column of the weight matrix is:
\begin{align}
\frac{d\ell(e_{t-n+1}^t, W, \bm{b})}{d\bm{b}} & = \bm{p} - \text{onehot}(e_t) \\
\frac{d\ell(e_{t-n+1}^t, W, \bm{b})}{dW_{\cdot,j}} & = x_j (\bm{p} - \text{onehot}(e_t))
\end{align}
Confirming these equations is left as a (highly recommended) exercise to the reader.
Hint: when performing this derivation, it is easier to work with the log probability $\log \bm{p}$ than working with $\bm{p}$ directly.

\subsection{Other Features for Language Modeling}
\label{sec:lllm:features}

One reason why log-linear models are nice is because they allow us to flexibly design features that we think might be useful for predicting the next word.
For example, these could include:
\begin{description}
\item[Context word features:] As shown in the example above, we can use the identity of $e_{t-1}$ or the identity of $e_{t-2}$.
\item[Context class:] Context words can be grouped into classes of similar words (using a method such as Brown clustering \cite{brown92classbased}), and instead of looking up a one-hot vector with a separate entry for every word, we could look up a one-hot vector with an entry for each class \cite{chen09shrinkingexponential}. Thus, words from the same class could share statistical strength, allowing models to generalize better.
\item[Context suffix features:] Maybe we want a feature that fires every time the previous word ends with ``...ing'' or other common suffixes. This would allow us to learn more generalized patterns about words that tend to follow progressive verbs, etc.
\item[Bag-of-words features:] Instead of just using the past $n$ words, we could use all previous words in the sentence. This would amount to calculating the one-hot vectors for every word in the previous sentence, and then instead of concatenating them simply summing them together. This would lose all information about what word is in what position, but could capture information about what words tend to co-occur within a sentence or document. 
\end{description}

It is also possible to combine together multiple features (for example $e_{t-1}$ is a particular word \textit{and} $e_{t-2}$ is another particular word).
This is one way to create a more expressive feature set, but also has a downside of greatly increasing the size of the feature space.
We discuss these features in more detail in \secref{nnlm:combination}.

\subsection{Further Reading}
\label{sec:lllm:furtherreading}

The language model in this section was basically a featurized version of an $n$-gram language model.
There are quite a few other varieties of linear featurized models including:
\begin{description}
\item[Whole-sentence language models:] These models, instead of predicting words one-by-one, predict the probability over the whole sentence then normalize \cite{rosenfeld01wholesentence}. This can be conducive to introducing certain features, such as a probability distribution over lengths of sentences, or features such as ``whether this sentence contains a verb''.
\item[Discriminative language models:] In the case that we want to use a language model to determine whether the output of a system is good or not, sometimes it is useful to train directly on this system output, and try to re-rank the outputs to achieve higher accuracy \cite{roark04discriminative}. Even if we don't have real negative examples, it can be possible to ``hallucinate'' negative examples that are still useful for training \cite{okanohara07pseudonegative}.
\end{description}

\subsection{Exercise}
\label{sec:lllm:exercise}

In the exercise for this chapter, we will construct a log-linear language model and evaluate its performance.
I highly suggest that you try to use the \texttt{NumPy} library to hold and perform calculations over feature vectors, as this will make things much easier.
If you have never used \texttt{NumPy} before, you can take a look at this tutorial to get started: \url{https://docs.scipy.org/doc/numpy-dev/user/quickstart.html}.

Writing the program will entail:
\begin{itemize}
\item Writing a function to read in the training and test corpora, and converting the words into numerical IDs.
\item Writing the feature function $\bm{\phi}(e_{t-n+1}^{t-1})$, which takes in a string and returns which features are active (for example, as a baseline these can be features with the identity of the previous two words).
\item Writing code to calculate the loss function.
\item Writing code to calculate gradients and perform stochastic gradient descent updates.
\item Writing (or re-using from the previous exercise) code to evaluate the language models.
\end{itemize}
Similarly to the $n$-gram language models, we will measure the per-word log likelihood and perplexity on our text corpus, and compare it to $n$-gram language models.
Handling unknown words will similarly require that you use the uniform distribution with 10,000,000 words in the English vocabulary.

Potential improvements to the model include designing better feature functions, adjusting the learning rate and measuring the results, and researching and implementing other types of optimizers such as AdaGrad or Adam.

  \section{Neural Networks and Feed-forward Language Models}
  \label{sec:nnlm}
  
In this chapter, we describe language models based on \term{neural networks}, a way to learn more sophisticated functions to improve the accuracy of our probability estimates with less feature engineering. 

\subsection{Potential and Problems with Combination Features}
\label{sec:nnlm:combination}

\begin{figure}[h]
 \centering
 \includegraphics[width=10cm]{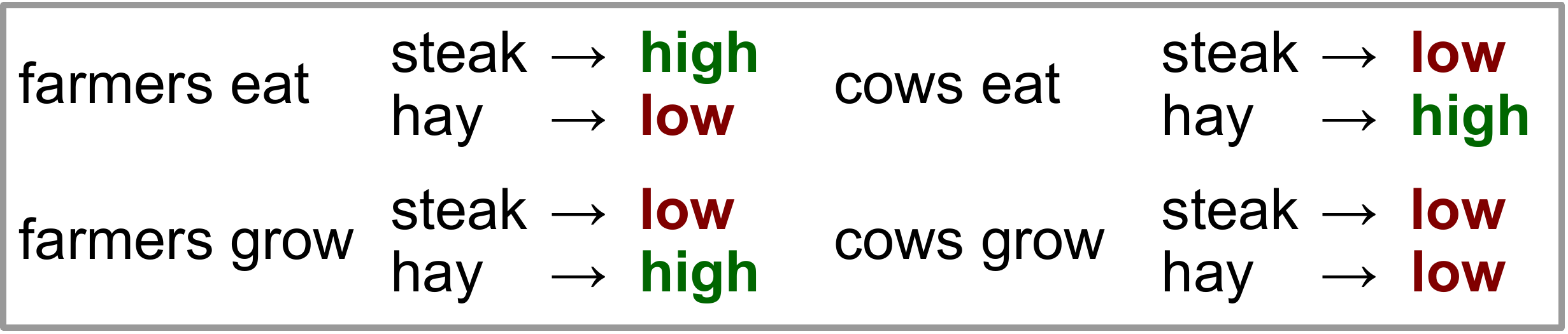}
 \caption{An example of the effect that combining multiple words can have on the probability of the next word.}
 \label{fig:nnlm:combinationexample}
\end{figure}

Before moving into the technical detail of neural networks, first let's take a look at a motivating example in \figref{nnlm:combinationexample}.
From the example, we can see $e_{t-2}=\text{``farmers''}$ is compatible with $e_{t}=\text{``hay''}$ (in the context ``farmers grow hay''), and $e_{t-1}=\text{``eat''}$ is also compatible (in the context ``cows eat hay'').
If we are using a log-linear model with one set of features dependent on $e_{t-1}$, and another set of features dependent on $e_{t-2}$, neither set of features can rule out the unnatural phrase ``farmers eat hay.''

One way we can fix this problem is by creating another set of features where we learn one vector for each pair of words $e_{t-2},e_{t-1}$.
If this is the case, our vector for the context $e_{t-2}=\text{``farmers''},e_{t-1}=\text{``eat''}$ could assign a low score to ``hay'', resolving this problem.
However, adding these combination features has one major disadvantage: it greatly expands the parameters: instead of $O(|V|^2)$ parameters for each pair $e_{i-1},e_i$, we need $O(|V|^3)$ parameters for each triplet $e_{i-2},e_{i-1},e_i$.
These numbers greatly increase the amount of memory used by the model, and if there are not enough training examples, the parameters may not be learned properly.

Because of both the importance of and difficulty in learning using these combination features, a number of methods have been proposed to handle these features, such as \term{kernelized support vector machines} \cite{cortes1995support} and \term{neural networks} \cite{rumelhart1988learning,goldberg15primer}.
Specifically in this section, we will cover neural networks, which are both flexible and relatively easy to train on large data, desiderata for sequence-to-sequence models.

\subsection{A Brief Overview of Neural Networks}
\label{sec:nnlm:nn}

% \begin{figure}
%  \centering
%  \includegraphics[width=12cm]{025-nnlm/figures/annneuron.pdf}
%  \caption[Two views of neural networks: one biological, and one mathematical.]{Two views of neural networks: one biological (on the left), and one mathematical (on the right).\footnotemark}
%  \label{fig:nnlm:annneuron}
% \end{figure}
% 
% Neural networks are a tool for machine learning that are expressive enough to learn these kinds of feature combinations.
% \figref{nnlm:annneuron} demonstrates the basic concept behind neural networks from two perspectives.
% The first perspective (on the left) is inspired by the network-like structure of the brain.
% The brain is made of neurons, which take stimuli from a variety of other neurons, and based on the stimuli moderate the stimuli that the provide to other neurons.
% \term{Artificial neural networks} are mathematical structures that mimic this behavior to some extent, as shown in the right of the figure.
% \footnotetext{This image is courtesy Joanna Ko\'{s}mider. Thanks to the author for creating it and releasing it to the public domain \url{https://commons.wikimedia.org/wiki/File:ANN_neuron.svg}}

\begin{figure}[h]
 \centering
 \includegraphics[width=4cm]{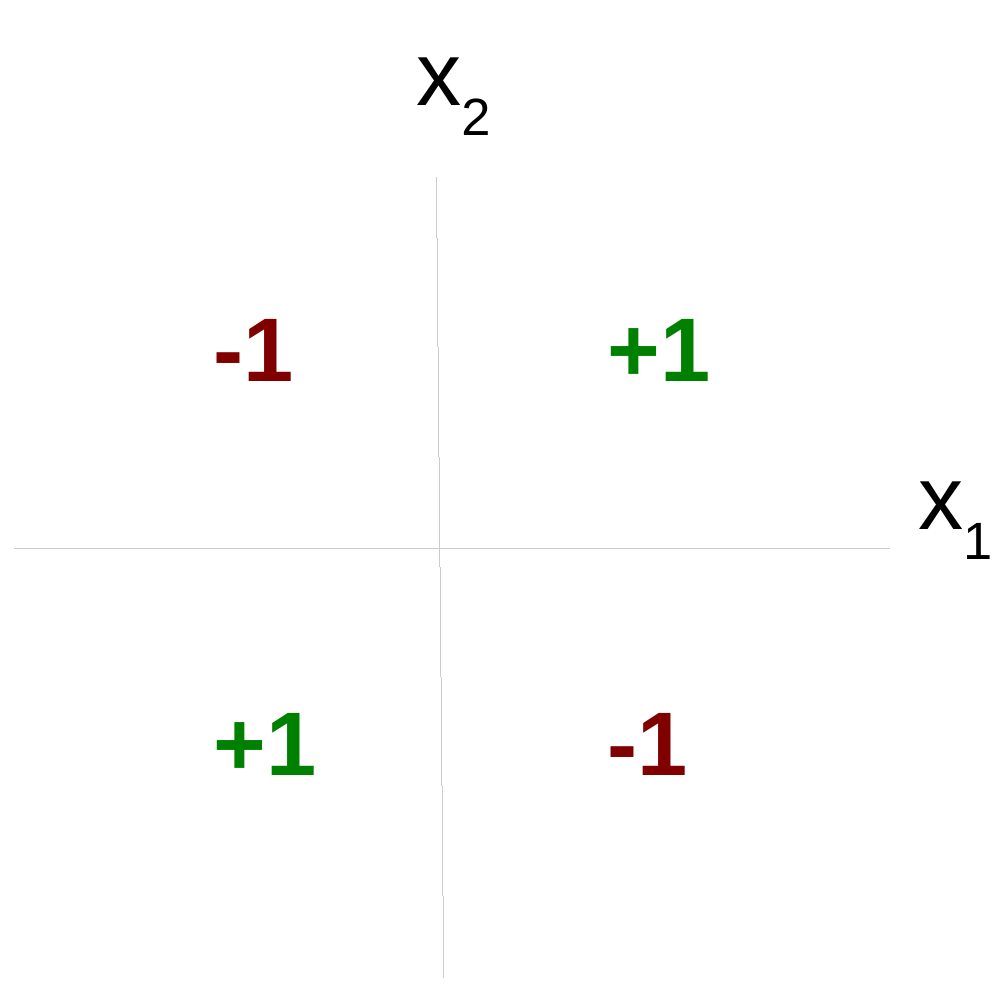}
 \caption{A function that cannot be solved by a linear transformation.}
 \label{fig:nnlm:unseparable}
\end{figure}

To understand neural networks in more detail, let's take a very simple example of a function that we cannot learn with a simple linear classifier like the ones we used in the last chapter: a function that takes an input $\bm{x} \in \{-1,1\}^2$ and outputs $y=1$ if both $x_1$ and $x_2$ are equal and $y=-1$ otherwise.
This function is shown in \figref{nnlm:unseparable}.

A first attempt at solving this function might define a linear model (like the log-linear models from the previous chapter) that solves this problem using the following form:
\begin{equation}
y = W\bm{x} + b.
\end{equation}
However, this class of functions is not powerful enough to represent the function at hand.\question{Prove this by trying to solve the system of equations.}

Thus, we turn to a slightly more complicated class of functions taking the following form:
\begin{align}
\bm{h} & = \text{step}(W_{xh} \bm{x} + \bm{b}_h) \nonumber \\
y & = \bm{w}_{hy} \bm{h} + b_y.
\label{eq:nnlm:stepmlp}
\end{align}
Computation is split into two stages: calculation of the \term{hidden layer}, which takes in input $\bm{x}$ and outputs a vector of hidden variables $\bm{h}$, and calculation of the \term{output layer}, which takes in $\bm{h}$ and calculates the final result $y$.
Both layers consist of an \term{affine transform}\footnote{A fancy name for a multiplication followed by an addition.} using weights $W$ and biases $\bm{b}$, followed by a $\text{step}(\cdot)$ function, which calculates the following:
\begin{equation}
\text{step}(x) = \begin{cases}
  1 & \mbox{if } x > 0, \\
  -1 & \mbox{otherwise}. \\
\end{cases}
\end{equation}
This function is one example of a class of neural networks called \term{multi-layer perceptrons} (MLPs).
In general, MLPs consist one or more hidden layers that consist of an affine transform followed by a non-linear function (such as the step function used here), culminating in an output layer that calculates some variety of output.

\begin{figure}
 \centering
 \includegraphics[width=16cm]{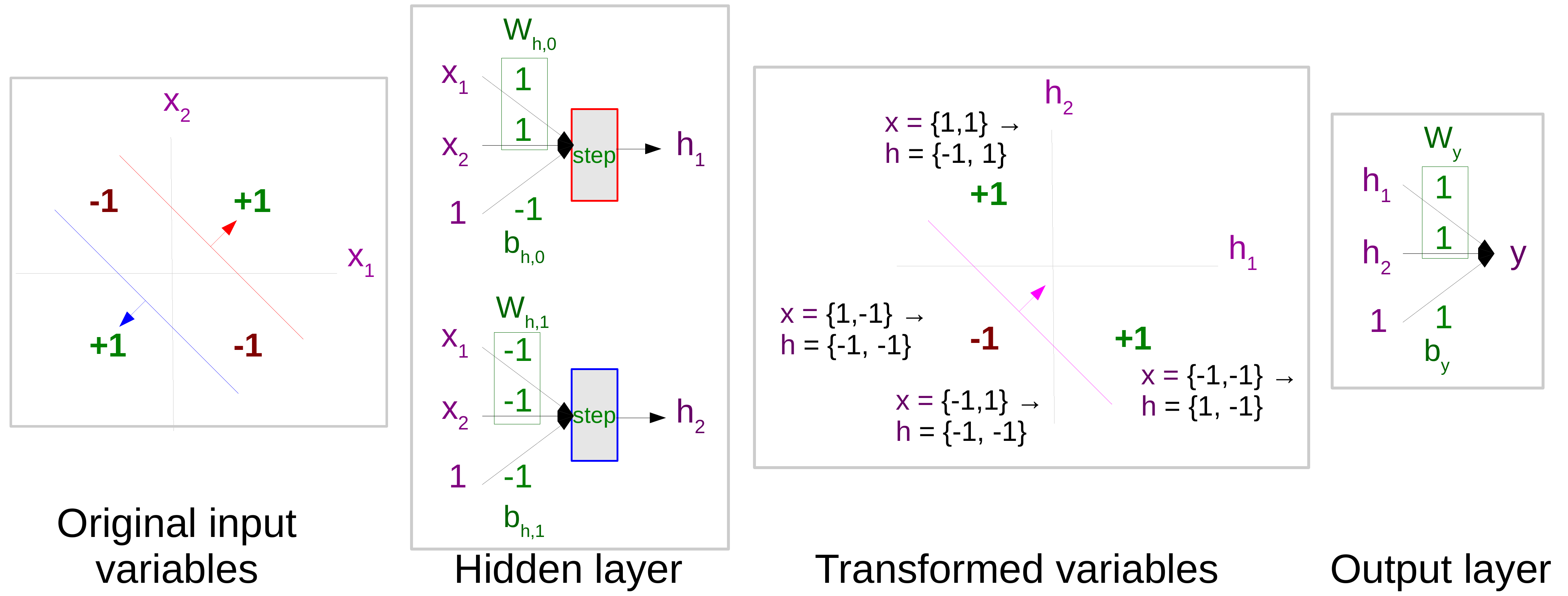}
 \caption{A simple neural network that represents the nonlinear function of \figref{nnlm:unseparable}.}
 \label{fig:nnlm:neuralnet}
\end{figure}

\figref{nnlm:neuralnet} demonstrates why this type of network does a better job of representing the non-linear function of \figref{nnlm:unseparable}.
In short, we can see that the first hidden layer \textit{transforms} the input $\bm{x}$ into a hidden vector $\bm{h}$ in a different space that is more conducive for modeling our final function.
Specifically in this case, we can see that $\bm{h}$ is now in a space where we can define a linear function (using $\bm{w}_y$ and $b_y$) that correctly calculates the desired output $y$.

As mentioned above, MLPs are one specific variety of neural network.
More generally, neural networks can be thought of as a chain of functions (such as the affine transforms and step functions used above, but also including many, many others) that takes some input and calculates some desired output.
The power of neural networks lies in the fact that chaining together a variety of simpler functions makes it possible to represent more complicated functions in an easily trainable, parameter-efficient way.
In fact, the simple single-layer MLP described above is a \term{universal function approximator} \cite{hornik1989multilayer}, which means that it can approximate any function to arbitrary accuracy if its hidden vector $\bm{h}$ is large enough.

We will see more about training in \secref{nnlm:training} and give some more examples of how these can be more parameter efficient in the discussion of neural network language models in \secref{nnlm:nnlm}.

\subsection{Training Neural Networks}
\label{sec:nnlm:training}

Now that we have a model in \eqref{nnlm:stepmlp}, we would like to train its parameters $W_{mh}$, $\bm{b}_h$, $\bm{w}_{hy}$, and $b_y$.
To do so, remembering our gradient-based training methods from the last chapter, we need to define the loss function $\ell(\cdot)$, calculate the derivative of the loss with respect to the parameters, then take a step in the direction that will reduce the loss.
For our loss function, let's use the \term{squared-error loss}, a commonly used loss function for regression problems which measures the difference between the calculated value $y$ and correct value $y^{*}$ as follows
\begin{equation}
\ell(y^{*}, y) = (y^{*} - y)^2.
\end{equation}

\begin{figure}
 \centering
 \includegraphics[width=16cm]{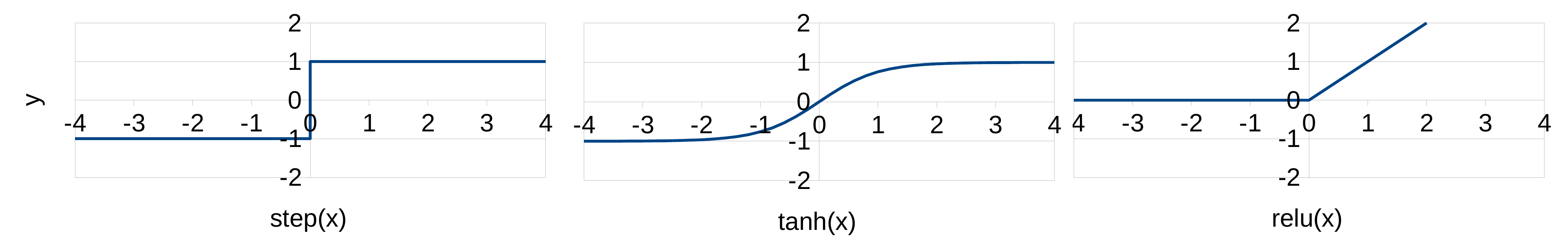}
 \caption{Types of non-linearities.}
 \label{fig:nnlm:nonlinearities}
\end{figure}

Next, we need to calculate derivatives.
Here, we run into one problem: the $\text{step}(\cdot)$ function is not very derivative friendly, with its derivative being:
\begin{equation}
\frac{d\text{step}(x)}{dx} = \begin{cases}
  \text{undefined} & \mbox{if } x = 0, \\
  0 & \mbox{otherwise}. \\
\end{cases}
\end{equation}
Because of this, it is more common to use other non-linear functions, such as the hyperbolic tangent (tanh) function.
The tanh function, as shown in \figref{nnlm:nonlinearities}, looks very much like a softened version of the step function that has a continuous gradient everywhere, making it more conducive to training with gradient-based methods.
There are a number of other alternatives as well, the most popular of which being the rectified linear unit (RelU)
\begin{equation}
\text{RelU}(x) = \begin{cases}
  x & \mbox{if } x > 0, \\
  0 & \mbox{otherwise}. \\
\end{cases}
\end{equation}
shown in the left of \figref{nnlm:nonlinearities}.
In short, RelUs solve the problem that the tanh function gets ``saturated'' and has very small gradients when the absolute value of input $x$ is very large ($x$ is a large negative or positive number).
Empirical results have often shown it to be an effective alternative to tanh, including for the language modeling task described in this chapter \cite{vaswani13neurallm}.

So let's say we swap in a tanh non-linearity instead of the step function to our network, we can now proceed to calculate derivatives like we did in \secref{lllm:derivatives}.
First, we perform the full calculation of the loss function:
\begin{align}
\bm{h}' & = W_{xh}\bm{x} + \bm{b}_h \nonumber \\
\bm{h}  & = \tanh(\bm{h}')        \nonumber \\
y       & = \bm{w}_{hy}\bm{h} + b_y \nonumber \\
\ell    & = (y^{*} - y)^2.       
\label{eq:nnlm:forward}
\end{align}
Then, again using the chain rule, we calculate the derivatives of each set of parameters:
\begin{align}
\frac{d\ell}{db_y}      & = \frac{d\ell}{dy} \frac{dy}{db_y}     \nonumber \\
\frac{d\ell}{d\bm{w}_{hy}} & = \frac{d\ell}{dy} \frac{dy}{d\bm{w}_{hy}}   \nonumber \\
\frac{d\ell}{d\bm{b}_h} & = \frac{d\ell}{dy} \frac{dy}{d\bm{h}} \frac{d\bm{h}}{d\bm{h}'} \frac{d\bm{h}'}{d\bm{b}_h}  \nonumber \\
\frac{d\ell}{dW_{xh}}      & = \frac{d\ell}{dy} \frac{dy}{d\bm{h}} \frac{d\bm{h}}{d\bm{h}'} \frac{d\bm{h}'}{dW_{xh}}.
\label{eq:nnlm:backward}
\end{align}

We could go through all of the derivations above by hand and precisely calculate the gradients of all parameters in the model.
Interested readers are free to do so, but even for a simple model like the one above, it is quite a lot of work and error prone.
For more complicated models, like the ones introduced in the following chapters, this is even more the case.

\begin{figure}
 \centering
 \includegraphics[width=8cm]{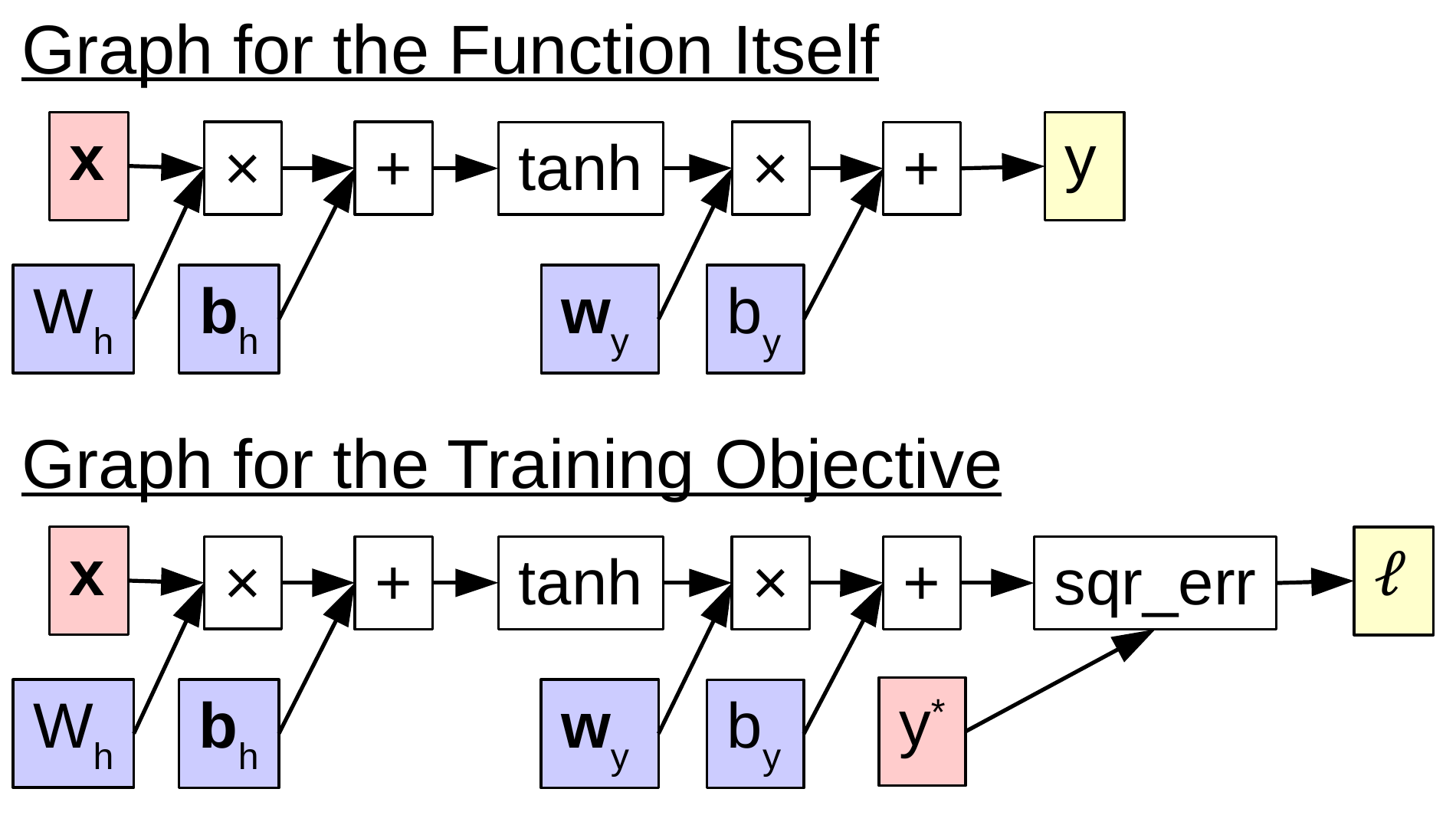}
 \caption{Computation graphs for the function itself, and the loss function.}
 \label{fig:nnlm:computationgraph}
\end{figure}

Fortunately, when we actually implement neural networks on a computer, there is a very useful tool that saves us a large portion of this pain: \term{automatic differentiation} (autodiff) \cite{wengert1964asa,griewank1991automatic}.
To understand automatic differentiation, it is useful to think of our computation in \eqref{nnlm:forward} as a data structure called a \term{computation graph}, two examples of which are shown in \figref{nnlm:computationgraph}.
In these graphs, each node represents either an input to the network or the result of one computational operation, such as a multiplication, addition, tanh, or squared error.
The first graph in the figure calculates the function of interest itself and would be used when we want to make predictions using our model, and the second graph calculates the loss function and would be used in training.

Automatic differentiation is a two-step dynamic programming algorithm that operates over the second graph and performs:
\begin{itemize}
\item \term{Forward calculation}, which traverses the nodes in the graph in topological order, calculating the actual result of the computation as in \eqref{nnlm:forward}.
\item \term{Back propagation}, which traverses the nodes in reverse topological order, calculating the gradients as in \eqref{nnlm:backward}.
\end{itemize}
The nice thing about this formulation is that while the overall function calculated by the graph can be relatively complicated, as long as it can be created by combining multiple simple nodes for which we are able to calculate the function $f(x)$ and derivative $f'(x)$, we are able to use automatic differentiation to calculate its derivatives using this dynamic program without doing the derivation by hand.

Thus, to implement a general purpose training algorithm for neural networks, it is necessary to implement these two dynamic programs, as well as the atomic forward function and backward derivative computations for each type of node that we would need to use.
While this is not trivial in itself, there are now a plethora of toolkits that either perform general-purpose auto-differentiation \cite{bendtsen1996fadbad,hogan2014fast}, or auto-differentiation specifically tailored for machine learning and neural networks \cite{abadi2016tensorflow,bergstra2010theano,collobert2002torch,tokui2015chainer,neubig2017dynet}.
These implement the data structures, nodes, back-propogation, and parameter optimization algorithms needed to train neural networks in an efficient and reliable way, allowing practitioners to get started with designing their models.
In the following sections, we will take this approach, taking a look at how to create our models of interest in a toolkit called \texttt{DyNet},\footnote{\url{http://github.com/clab/dynet}} which has a programming interface that makes it relatively easy to implement the sequence-to-sequence models covered here.%
\footnote{It is also developed by the author of these materials, so the decision might have been a wee bit biased.}

\subsection{An Example Implementation}
\label{sec:nnlm:nnexample}

\begin{figure}
  \centering
  \begin{minted}[mathescape,
                 linenos,
                 numbersep=5pt,
                 gobble=2,
                 frame=lines,
                 framesep=2mm]{python}
  import dynet as dy
  import random
  # Parameters of the model and training
  HIDDEN_SIZE = 20
  NUM_EPOCHS = 20
  # Define the model and SGD optimizer
  model = dy.Model()
  W_xh_p = model.add_parameters((HIDDEN_SIZE, 2))
  b_h_p = model.add_parameters(HIDDEN_SIZE)
  W_hy_p = model.add_parameters((1, HIDDEN_SIZE))
  b_y_p = model.add_parameters(1)
  trainer = dy.SimpleSGDTrainer(model)
  # Define the training data, consisting of (x,y) tuples
  data = [([1,1],1), ([-1,1],-1), ([1,-1],-1), ([-1,-1],1)]
  # Define the function we would like to calculate
  def calc_function(x):
    dy.renew_cg()
    w_xh = dy.parameter(w_xh_p)
    b_h = dy.parameter(b_h_p)
    W_hy = dy.parameter(W_hy_p)
    b_y = dy.parameter(b_y_p)
    x_val = dy.inputVector(x)
    h_val = dy.tanh(w_xh * x_val + b_h)
    y_val = W_hy * h_val + b_y
    return y_val
  # Perform training
  for epoch in range(NUM_EPOCHS):
    epoch_loss = 0
    random.shuffle(data)
    for x, ystar in data:
      y = calc_function(x)
      loss = dy.squared_distance(y, dy.scalarInput(ystar))
      epoch_loss += loss.value()
      loss.backward()
      trainer.update()
    print("Epoch %d: loss=%f" % (epoch, epoch_loss))
  # Print results of prediction
  for x, ystar in data:
    y = calc_function(x)
    print("%r -> %f" % (x, y.value()))
  \end{minted}
  \caption{An example of training a neural network for a multi-layer perceptron using the toolkit \texttt{DyNet}.}
  \label{fig:nnlm:dynetexample}
\end{figure}

\figref{nnlm:dynetexample} shows an example of implementing the above neural network in \texttt{DyNet}, which we'll step through line-by-line.
Lines 1-2 import the necessary libraries.
Lines 4-5 specify parameters of the models: the size of the hidden vector $\bm{h}$ and the number of epochs (passes through the data) for which we'll perform training.
Line 7 initializes a \texttt{DyNet} model, which will store all the parameters we are attempting to learn.
Lines 8-11 initialize parameters $W_{xh}$, $\bm{b}_h$, $\bm{w}_{hy}$, and $b_y$ to be the appropriate size so that dimensions in the equations for \eqref{nnlm:forward} match.
Line 12 initializes a ``trainer'', which will update the parameters in the model according to an update strategy (here we use simple stochastic gradient descent, but trainers for AdaGrad, Adam, and other strategies also exist).
Line 14 creates the training data for the function in \figref{nnlm:unseparable}.

Lines 16-25 define a function that takes input $\bm{x}$ and creates a computation graph to calculate \eqref{nnlm:forward}.
First, line 17 creates a new computation graph to hold the computation for this particular training example.
Lines 18-21 take the parameters (stored in the model) and adds them to the computation graph as \texttt{DyNet} variables for this particular training example.
Line 22 takes a Python list representing the current input and puts it into the computation graph as a \texttt{DyNet} variable.
Line 23 calculates the hidden vector $\bm{h}$, Line 24 calculates the value $y$, and Line 25 returns it. 

Lines 27-36 perform training for \texttt{NUM\_EPOCHS} passes over the data (one pass through the training data is usually called an ``epoch'').
Line 28 creates a variable to keep track of the loss for this epoch for later reporting.
Line 29 shuffles the data, as recommended in \secref{lllm:learning}.
Lines 30-35 perform stochastic gradient descent, looping over each of the training examples.
Line 31 creates a computation for the function itself, and Line 32 adds computation for the loss function.
Line 33 runs the forward calculation to calculate the loss and adds it to the loss for this epoch.
Line 34 runs back propagation, and Line 35 updates the model parameters.
At the end of the epoch, we print the loss for the epoch in Line 36 to make sure that the loss is going down and our model is converging.

Finally, at the end of training in Lines 38-40, we print the output results.
In an actual scenario, this would be done on a separate set of test data.

\subsection{Neural-network Language Models}
\label{sec:nnlm:nnlm}

Now that we have the basics down, it is time to apply neural networks to language modeling \cite{nakamura90wordcategory,bengio06nnlm}.
A feed-forward neural network language model is very much like the log-linear language model that we mentioned in the previous section, simply with the addition of one or more non-linear layers before the output.

First, let's recall the tri-gram log-linear language model.
In this case, assume we have two sets of features expressing the identity of $e_{t-1}$ (represented as $W^{(1)}$) and $e_{t-2}$ (as $W^{(2)}$), the equation for the log-linear model looks like this:
\begin{align}
\bm{s} & = W^{(1)}_{\cdot,e_{t-1}} + W^{(2)}_{\cdot,e_{t-2}} + \bm{b} \nonumber \\
\bm{p} & = \text{softmax}(\bm{s}),
\end{align}
where we add the appropriate columns from the weight matricies to the bias to get the score, then take the softmax to turn it into a probability.

\begin{figure}
 \centering
 \includegraphics[width=12cm]{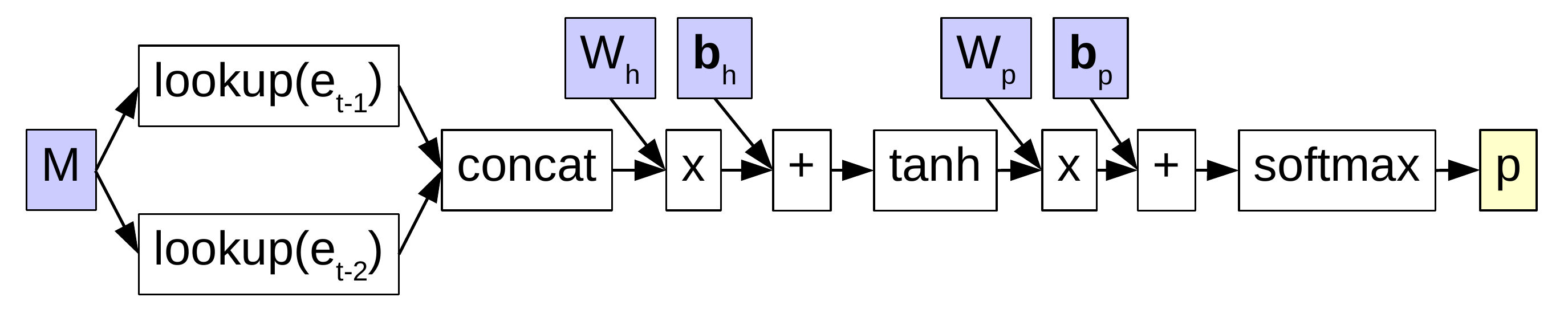}
 \caption{A computation graph for a tri-gram feed-forward neural language model.}
 \label{fig:nnlm:ffcomp}
\end{figure}

Compared to this, a tri-gram neural network model with a single layer is structured as shown in \figref{nnlm:ffcomp} and described in equations below: 
\begin{align}
\bm{m} & = \text{concat}(M_{\cdot,e_{t-2}},M_{\cdot,e_{t-1}}) \nonumber \\
\bm{h} & = \text{tanh}(W_{mh} \bm{m} + \bm{b}_h) \nonumber \\
\bm{s} & = W_{hs} \bm{h} + \bm{b}_s \nonumber \\
\bm{p} & = \text{softmax}(\bm{s})
\label{eq:nnlm:nnlm}
\end{align}

In the first line, we obtain a vector $\bm{m}$ representing the context $e_{i-n+1}^{i-1}$ (in the particular case above, we are handling a tri-gram model so $n=3$).
Here, $M$ is a matrix with $|V|$ columns, and $L_m$ rows, where each column corresponds to an $L_m$-length vector representing a single word in the vocabulary.
This vector is called a \term{word embedding} or a \term{word representation}, which is a vector of real numbers corresponding to particular words in the vocabulary.%
\footnote{For the purposes of the model in this chapter, these vectors can basically be viewed as one set of tunable parameters in the neural language model, but there has also been a large amount of interest in learning these vectors for use in other tasks. Some methods are outlined in \secref{nnlm:further}.}
The interesting thing about expressing words as vectors of real numbers is that each element of the vector could reflect a different aspect of the word.
For example, there may be an element in the vector determining whether a particular word under consideration could be a noun, or another element in the vector expressing whether the word is an animal, or another element that expresses whether the word is countable or not.\footnote{In reality, it is rare that single elements in the vector have such an intuitive meaning unless we impose some sort of constraint, such as sparsity constraints \cite{murphy12sparseembedding}.}
\figref{nnlm:dynetlookup} shows an example of how to define parameters that allow you to look up a vector in \texttt{DyNet}.

\begin{figure}
  \centering
  \begin{minted}[mathescape,
                 linenos,
                 numbersep=5pt,
                 gobble=2,
                 frame=lines,
                 framesep=2mm]{python}
  # Define the lookup parameters at model definition time
  # VOCAB_SIZE is the number of words in the vocabulary
  # EMBEDDINGS_SIZE is the length of the word embedding vector
  M_p = model.add_lookup_parameters((VOCAB_SIZE, EMBEDDING_SIZE))
  # Load the parameters into the computation graph 
  M = dy.lookup(M_p)
  # And look up the vector for word i
  m = M[i]
  \end{minted}
  \caption{Code for looking things up in \texttt{DyNet}.}
  \label{fig:nnlm:dynetlookup}
\end{figure}

The vector $\bm{m}$ then results from the concatenation of the word vectors for all of the words in the context, so $|\bm{m}| = L_m*(n-1)$.
Once we have this $\bm{m}$, we run the vectors through a hidden layer to obtain vector $\bm{h}$.
By doing so, the model can learn combination features that reflect information regarding multiple words in the context.
This allows the model to be expressive enough to represent the more difficult cases in \figref{nnlm:combinationexample}.
For example, given a context is ``cows eat'', and some elements of the vector $M_{\cdot,\text{cows}}$ identify the word as a ``large farm animal'' (e.g. ``cow'', ``horse'', ``goat''), while some elements of $M_{\cdot,\text{eat}}$ corresponds to ``eat'' and all of its relatives (``consume'', ``chew'', ``ingest''), then we could potentially learn a unit in the hidden layer $\bm{h}$ that is active when we are in a context that represents ``things farm animals eat''.

Next, we calculate the score vector for each word: $\bm{s} \in \mathbb{R}^{|V|}$. 
This is done by performing an affine transform of the hidden vector $\bm{h}$ with a weight matrix $W_{hs} \in \mathbb{R}^{|V| \times |\bm{h}|}$ and adding a bias vector $\bm{b}_s \in \mathbb{R}^{|V|}$.
Finally, we get a probability estimate $\bm{p}$ by running the calculated scores through a softmax function, like we did in the log-linear language models.
For training, if we know $e_t$ we can also calculate the loss function as follows, similarly to the log-linear model:
\begin{equation}
\ell = -\log(p_{e_t}).
\end{equation}
\texttt{DyNet} has a convenience function that, given a score vector $\bm{s}$, will calculate the negative log likelihood loss:
\begin{figure}[h]
  \centering
  \begin{minted}[mathescape,
                 linenos,
                 numbersep=5pt,
                 gobble=2,
                 frame=lines,
                 framesep=2mm]{python}
  loss = dy.pickneglogsoftmax(s, e_t)
  \end{minted}
\end{figure}

% TODO: discuss geometric interpretation
% One interesting way to think of the matrix $W_s$ is as an \term{output embedding} matrix, which is a counterpart to the \term{input embedding} $M$ \cite{bilinear,outputembedding,socherembeddingpaper}.
% Each row of $W_s$ is a length $|\bm{h}|$ vector corresponding to a single word in the vocabulary.
% So, we can think of the $i$th word in vocabulary $V$ as a point in space represented by the $i$th row of $W_s$, and it is the job of our network to find an $\bm{h}$ that is close to this point.

The reasons why the neural network formulation is nice becomes apparent when we compare this to $n$-gram language models in \secref{ngramlm}:
\begin{description}
\item[Better generalization of contexts:]
$n$-gram language models treat each word as its own discrete entity.
By using input embeddings $M$, it is possible to group together similar words so they behave similarly in the prediction of the next word.
In order to do the same thing, $n$-gram models would have to explicitly learn word classes and using these classes effectively is not a trivial problem \cite{brown92classbased}.
\item[More generalizable combination of words into contexts:]
In an $n$-gram language model, we would have to remember parameters for all combinations of $\{\text{cow}, \text{horse}, \text{goat}\} \times \{\text{consume}, \text{chew}, \text{ingest}\}$ to represent the context ``things farm animals eat''.
This would be quadratic in the number of words in the class, and thus learning these parameters is difficult in the face of limited training data.
Neural networks handle this problem by learning nodes in the hidden layer that can represent this quadratic combination in a feature-efficient way.
\item[Ability to skip previous words:]
$n$-gram models generally fall back sequentially from longer contexts (e.g. ``the two previous words $e_{t-2}^{t-1}$'') to shorter contexts (e.g. ``the previous words $e_{t-1}$''), but this doesn't allow them to ``skip'' a word and only reference for example, ``the word two words ago $e_{t-2}$''.
Log-linear models and neural networks can handle this skipping naturally.
\end{description}

\subsection{Further Reading}
\label{sec:nnlm:further}

In addition to the methods described above, there are a number of extensions to neural-network language models that are worth discussing.

\begin{description}
\item[Softmax approximations:]
One problem with the training of log-linear or neural network language models is that at every training example, they have to calculate the large score vector $\bm{s}$, then run a softmax over it to get probabilities.
As the vocabulary size $|V|$ grows larger, this can become quite time-consuming.
As a result, there are a number of ways to reduce training time.
One example are methods that sample a subset of the vocabulary $V' \in V$ where $|V'| \ll V$, then calculate the scores and approximate the loss over this smaller subset.
Examples of these include methods that simply try to get the true word $e_t$ to have a higher score (by some margin) than others in the subsampled set \cite{collobert11natural} and more probabilistically motivated methods, such as \term{importance sampling} \cite{bengio2008adaptive} or \term{noise-contrastive estimation} (NCE; \cite{mnih12nce}).
Interestingly, for other objective functions such as linear regression and special variety of softmax called the \term{spherical softmax}, it is possible to calculate the objective function in ways that do not scale linearly with the vocabulary size \cite{vincent2015efficient}.
\item[Other softmax structures:]
Another interesting trick to improve training speed is to create a softmax that is structured so that its loss functions can be computed efficiently.
One way to do so is the class-based softmax \cite{goodman2001classes}, which assigns each word $e_t$ to a class $c_t$, then divides computation into two steps: predicting the probability of class $c_t$ given the context, then predicting the probability of the word $e_t$ given the class and the current context $P(e_t \mid c_t,e_{t-n+1}^{t-1})P(c_t \mid e_{t-n+1}^{t-1})$.
The advantage of this method is that we only need to calculate scores for the correct class $c_t$ out of $|C|$ classes, then the correct word $e_t$ out of the vocabulary for class $c_t$, which is size $|V_{c_t}|$.
Thus, our computational complexity becomes $O(|C|+|V_{c_t}|)$ instead of $O(|V|)$.\question{What is the ideal class size to achieve the best computational efficiency?}
The hierarchical softmax \cite{mikolov13distributed} takes this a step further by predicting words along a binary-branching tree, which results in a computational complexity of $O({\log}_2 |V|)$.
\item[Other models to learn word representations:]
As mentioned in \secref{nnlm:nnlm}, we learn word embeddings $M$ as a by-product of training our language models.
One very nice feature of word representations is that language models can be trained purely on raw text, but the resulting representations can capture semantic or syntactic features of the words, and thus can be used to effectively improve down-stream tasks that don't have a lot of annotated data, such as part-of-speech tagging or parsing \cite{turian2010wordrepresentations}.%
\footnote{Manning (2015) called word embeddings the ``Sriracha sauce of NLP'', because you can add them to anything to make it better \url{http://nlp.stanford.edu/~manning/talks/NAACL2015-VSM-Compositional-Deep-Learning.pdf}}
Because of their usefulness, there have been an extremely large number of approaches proposed to learn different varieties of word embeddings,%
\footnote{So many that Daum\'{e} III (2016) called word embeddings the ``Sriracha sauce of NLP: it sounds like a good idea, you add too much, and now you're crying'' \url{https://twitter.com/haldaume3/status/706173575477080065}}
from early work based on distributional similarity and dimensionality reduction \cite{schutze1993word,turney2010frequency} to more recent models based on predictive models similar to language models \cite{turian2010wordrepresentations,mikolov2013efficient}, with the general current thinking being that predictive models are the more effective and flexible of the two \cite{baroni2014dontcount}.%
The most well-known methods are the continuous-bag-of-words and skip-gram models implemented in the software \texttt{word2vec},%
\footnote{\url{https://code.google.com/archive/p/word2vec/}}
which define simple objectives for predicting words using the immediately surrounding context or vice-versa.
\texttt{word2vec} uses a sampling-based approach and parallelization to easily scale up to large datasets, which is perhaps the primary reason for its popularity.
One thing to note is that these methods are not language models in themselves, as they do not calculate a probability of the sentence $P(E)$, but many of the parameter estimation techniques can be shared.
\end{description}

\subsection{Exercise}
\label{sec:nnlm:exercise}

In the exercise for this chapter, we will use \texttt{DyNet} to construct a feed-forward language model and evaluate its performance.

Writing the program will entail:
\begin{itemize}
\item Writing a function to read in the data and (turn it into numerical IDs).
\item Writing a function to calculate the loss function by looking up word embeddings, then running them through a multi-layer perceptron, then predicting the result.
\item Writing code to perform training using this function.
\item Writing evaluation code that measures the perplexity on a held-out data set.
\end{itemize}
Language modeling accuracy should be measured in the same way as previous exercises and compared with the previous models.

Potential improvements to the model include tuning the various parameters of the model. 
How big should $\bm{h}$ be?
Should we add additional hidden layers?
What optimizer with what learning rate should we use?
What happens if we implement one of the more efficient versions of the softmax explained in \secref{nnlm:further}?

  \section{Recurrent Neural Network Language Models}
  \label{sec:rnnlm}
  
The neural-network models presented in the previous chapter were essentially more powerful and generalizable versions of $n$-gram models.
In this section, we talk about language models based on recurrent neural networks (RNNs), which have the additional ability to capture long-distance dependencies in language.

\subsection{Long Distance Dependencies in Language}
\label{sec:rnnlm:dependencies}

\begin{figure}[h]
 \centering
 \includegraphics[width=9cm]{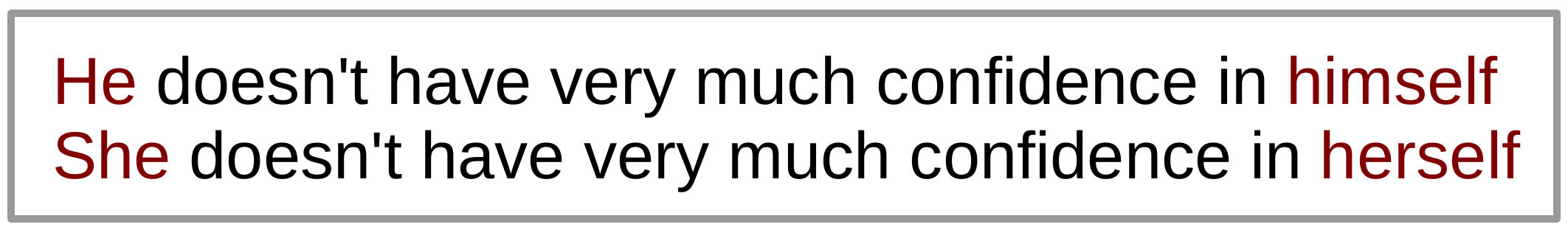}
 \caption{An example of long-distance dependencies in language.}
 \label{fig:rnnlm:longdistance}
\end{figure}

Before speaking about RNNs in general, it's a good idea to think about the various reasons a model with a limited history would not be sufficient to properly model all phenomena in language.

One example of a long-range \text{grammatical constraint} is shown in \figref{rnnlm:longdistance}.
In this example, there is a strong constraint that the starting ``he'' or ``her'' and the final ``himself'' or ``herself'' must match in gender.
Similarly, based on the subject of the sentence, the conjugation of the verb will change.
These sorts of dependencies exist regardless of the number of intervening words, and models with a finite history $e_{i-n+1}^{i-1}$, like the one mentioned in the previous chapter, will never be able to appropriately capture this.
These dependencies are frequent in English but even more prevalent in languages such as Russian, which has a large number of forms for each word, which must match in case and gender with other words in the sentence.%
\footnote{See \url{https://en.wikipedia.org/wiki/Russian\_grammar} for an overview.}

Another example where long-term dependencies exist is in \term{selectional preferences} \cite{resnik1997selectional}.
In a nutshell, selectional preferences are basically common sense knowledge of ``what will do what to what''.
For example, ``I ate salad with a fork'' is perfectly sensible with ``a fork'' being a tool, and ``I ate salad with my friend'' also makes sense, with ``my friend'' being a companion.
On the other hand, ``I ate salad with a backpack'' doesn't make much sense because a backpack is neither a tool for eating nor a companion.
These selectional preference violations lead to nonsensical sentences and can also span across an arbitrary length due to the fact that subjects, verbs, and objects can be separated by a great distance.

Finally, there are also dependencies regarding the \term{topic} or \term{register} of the sentence or document.
For example, it would be strange if a document that was discussing a technical subject suddenly started going on about sports -- a violation of topic consistency.
It would also be unnatural for a scientific paper to suddenly use informal or profane language -- a lack of consistency in register.

These and other examples describe why we need to model long-distance dependencies to create workable applications.

\subsection{Recurrent Neural Networks}
\label{sec:rnnlm:recurrent}

\begin{figure}
 \centering
 \includegraphics[width=15cm]{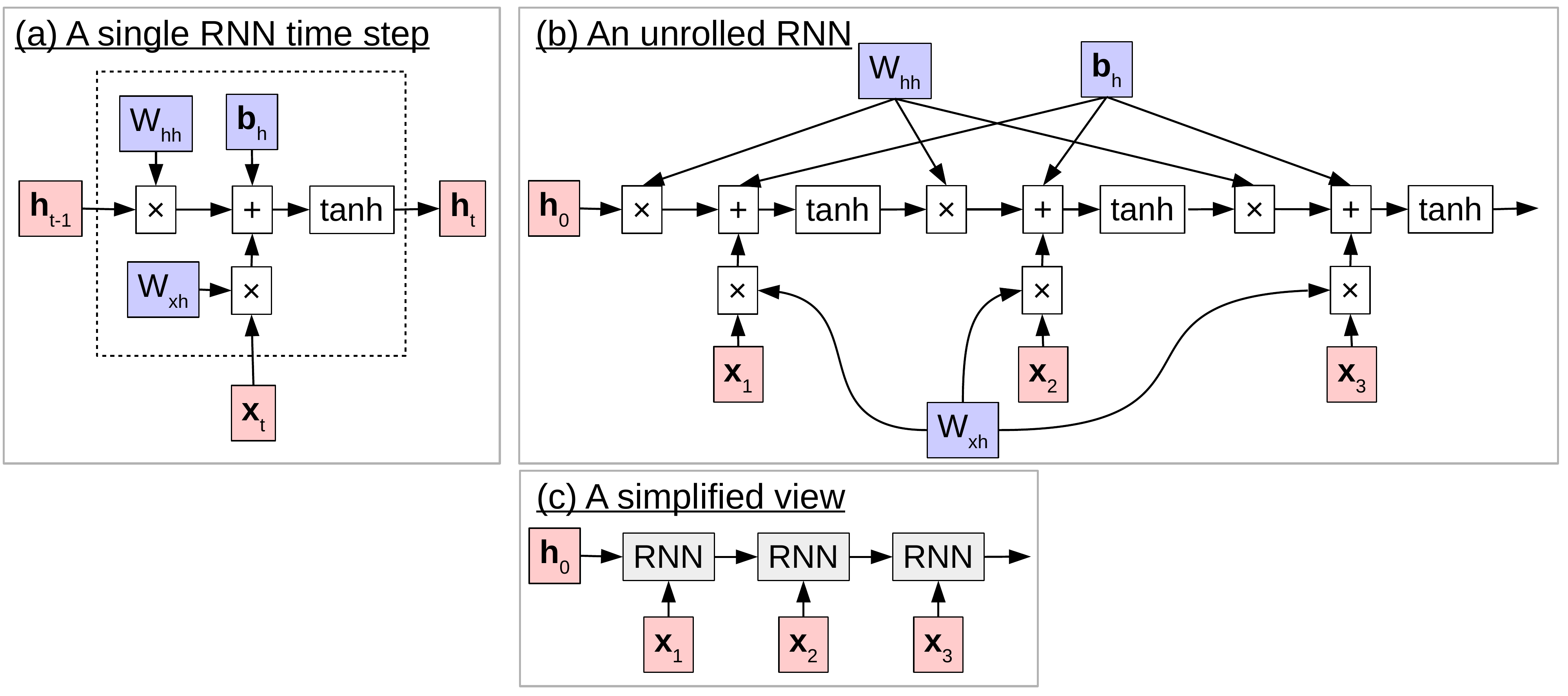}
 \caption{Examples of computation graphs for neural networks. (a) shows a single time step. (b) is the unrolled network. (c) is a simplified version of the unrolled network, where gray boxes indicate a function that is parameterized (in this case by $W_{xh}$, $W_{hh}$, and $\bm{b}_h$).}
 \label{fig:rnnlm:rnn}
\end{figure}

\term{Recurrent neural networks} (RNNs; \cite{elman90rnn}) are a variety of neural network that makes it possible to model these long-distance dependencies.
The idea is simply to add a connection that references the previous hidden state $\bm{h}_{t-1}$ when calculating hidden state $\bm{h}$, written in equations as:
\begin{equation}
\bm{h}_t = \begin{cases}
  \text{tanh}(W_{xh} \bm{x}_t + W_{hh} \bm{h}_{t-1} + \bm{b}_h) & t \ge 1, \\
  \bm{0} & \mbox{otherwise}. \\
\end{cases}
\label{eq:rnnlm:rnn}
\end{equation}
As we can see, for time steps $t\ge1$, the only difference from the hidden layer in a standard neural network is the addition of the connection $W_{hh} \bm{h}_{t-1}$ from the hidden state at time step $t-1$ connecting to that at time step $t$. 
As this is a recursive equation that uses $\bm{h}_{t-1}$ from the previous time step.
This single time step of a recurrent neural network is shown visually in the computation graph shown in \figref{rnnlm:rnn}(a).

When performing this visual display of RNNs, it is also common to ``unroll'' the neural network in time as shown in \figref{rnnlm:rnn}(b), which makes it possible to explicitly see the information flow between multiple time steps.
From unrolling the network, we can see that we are still dealing with a standard computation graph in the same form as our feed-forward networks, on which we can still do forward computation and backward propagation, making it possible to learn our parameters. 
It also makes clear that the recurrent network has to start somewhere with an initial hidden state $\bm{h}_0$.
This initial state is often set to be a vector full of zeros, treated as a parameter $\bm{h}_{\text{init}}$ and learned, or initialized according to some other information (more on this in \secref{encdec}).

Finally, for simplicity, it is common to abbreviate the whole recurrent neural network step with a single block ``RNN'' as shown in \figref{rnnlm:rnn}.
In this example, the boxes corresponding to RNN function applications are gray, to show that they are internally parameterized with $W_{xh}$, $W_{hh}$, and $\bm{b}_h$.
We will use this convention in the future to represent parameterized functions.

RNNs make it possible to model long distance dependencies because they have the ability to pass information between timesteps.
For example, if some of the nodes in $\bm{h}_{t-1}$ encode the information that ``the subject of the sentence is male'', it is possible to pass on this information to $\bm{h}_{t}$, which can in turn pass it on to $\bm{h}_{t+1}$ and on to the end of the sentence.
This ability to pass information across an arbitrary number of consecutive time steps is the strength of recurrent neural networks, and allows them to handle the long-distance dependencies described in \secref{rnnlm:dependencies}.

Once we have the basics of RNNs, applying them to language modeling is (largely) straight-forward \cite{mikolov10rnnlm}.
We simply take the feed-forward language model of \eqref{nnlm:nnlm} and enhance it with a recurrent connection as follows:
\begin{align}
\bm{m}_t & = M_{\cdot,e_{t-1}} \nonumber \\
\bm{h}_t & = \begin{cases}
  \text{tanh}(W_{mh} \bm{m}_t + W_{hh} \bm{h}_{t-1} + \bm{b}_h) & t \ge 1, \\
  \bm{0} & \mbox{otherwise}. \\
\end{cases} \nonumber \\
\bm{p}_t & = \text{softmax}(W_{hs} \bm{h}_t + b_s).
\label{eq:rnnlm:rnnlm}
\end{align}
One thing that should be noted is that compared to the feed-forward language model, we are only feeding in the previous word instead of the two previous words.
The reason for this is because (if things go well) we can expect that information about $e_{t-2}$ and all previous words are already included in $\bm{h}_{t-1}$, making it unnecessary to feed in this information directly.

Also, for simplicity of notation, it is common to abbreviate the equation for $\bm{h}_t$ with a function $\text{RNN}(\cdot)$, following the simplified view of drawing RNNs in \figref{rnnlm:rnn}(c):
\begin{align}
\bm{m}_t & = M_{\cdot,e_{t-1}} \nonumber \\
\bm{h}_t & = \text{RNN}(\bm{m}_t, \bm{h}_{t-1}) \nonumber \\
\bm{p}_t & = \text{softmax}(W_{hs} \bm{h}_t + b_s).
\label{eq:rnnlm:rnnlmabbrv}
\end{align}

\subsection{The Vanishing Gradient and Long Short-term Memory}
\label{sec:rnnlm:lstm}

\begin{figure}
 \centering
 \includegraphics[width=10cm]{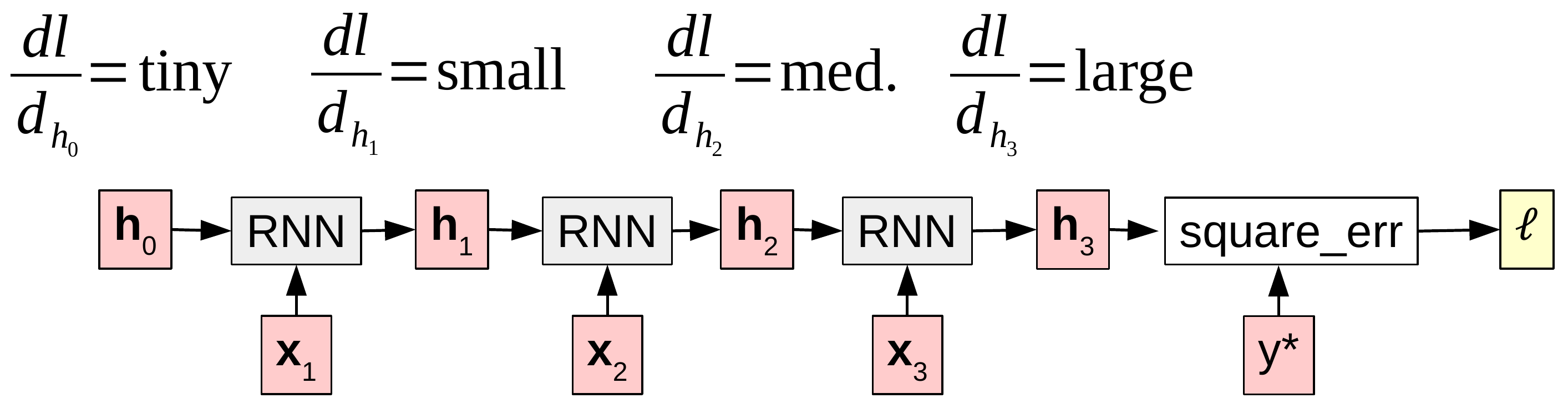}
 \caption{An example of the vanishing gradient problem.}
 \label{fig:rnnlm:vanishing}
\end{figure}

However, while the RNNs in the previous section are conceptually simple, they also have problems: the \term{vanishing gradient} problem and the closely related cousin, the \term{exploding gradient} problem.

A conceptual example of the vanishing gradient problem is shown in \figref{rnnlm:vanishing}.
In this example, we have a recurrent neural network that makes a prediction after several times steps, a model that could be used to classify documents or perform any kind of prediction over a sequence of text.
After it makes its prediction, it gets a loss that it is expected to back-propagate over all time steps in the neural network.
However, at each time step, when we run the back propagation algorithm, the gradient gets smaller and smaller, and by the time we get back to the beginning of the sentence, we have a gradient so small that it effectively has no ability to have a significant effect on the parameters that need to be updated.
The reason why this effect happens is because unless $\frac{d\bm{h}_{t-1}}{d\bm{h}_t}$ is exactly one, it will tend to either diminish or amplify the gradient $\frac{d\ell}{d\bm{h}_t}$, and when this diminishment or amplification is done repeatedly, it will have an exponential effect on the gradient of the loss.%
\footnote{This is particularly detrimental in the case where we receive a loss only once at the end of the sentence, like the example above.
One real-life example of such a scenario is document classification, and because of this, RNNs have been less successful in this task than other methods such as convolutional neural networks, which do not suffer from the vanishing gradient problem \cite{kim2014cnntextcat,lei2015moldingcnns}.
It has been shown that pre-training an RNN as a language model before attempting to perform classification can help alleviate this problem to some extent \cite{dai2015semisupervised}.
}

One method to solve this problem, in the case of diminishing gradients, is the use of a neural network architecture that is specifically designed to ensure that the derivative of the recurrent function is exactly one.
A neural network architecture designed for this very purpose, which has enjoyed quite a bit of success and popularity in a wide variety of sequential processing tasks, is the \term{long short-term memory} (LSTM; \cite{hochreiter97lstm}) neural network architecture.
The most fundamental idea behind the LSTM is that in addition to the standard hidden state $\bm{h}$ used by most neural networks, it also has a \term{memory cell} $\bm{c}$, for which the gradient $\frac{d\bm{c}_{t}}{d\bm{c}_{t-1}}$ is exactly one.
Because this gradient is exactly one, information stored in the memory cell does not suffer from vanishing gradients, and thus LSTMs can capture long-distance dependencies more effectively than standard recurrent neural networks.

\begin{figure}
 \centering
 \includegraphics[width=10cm]{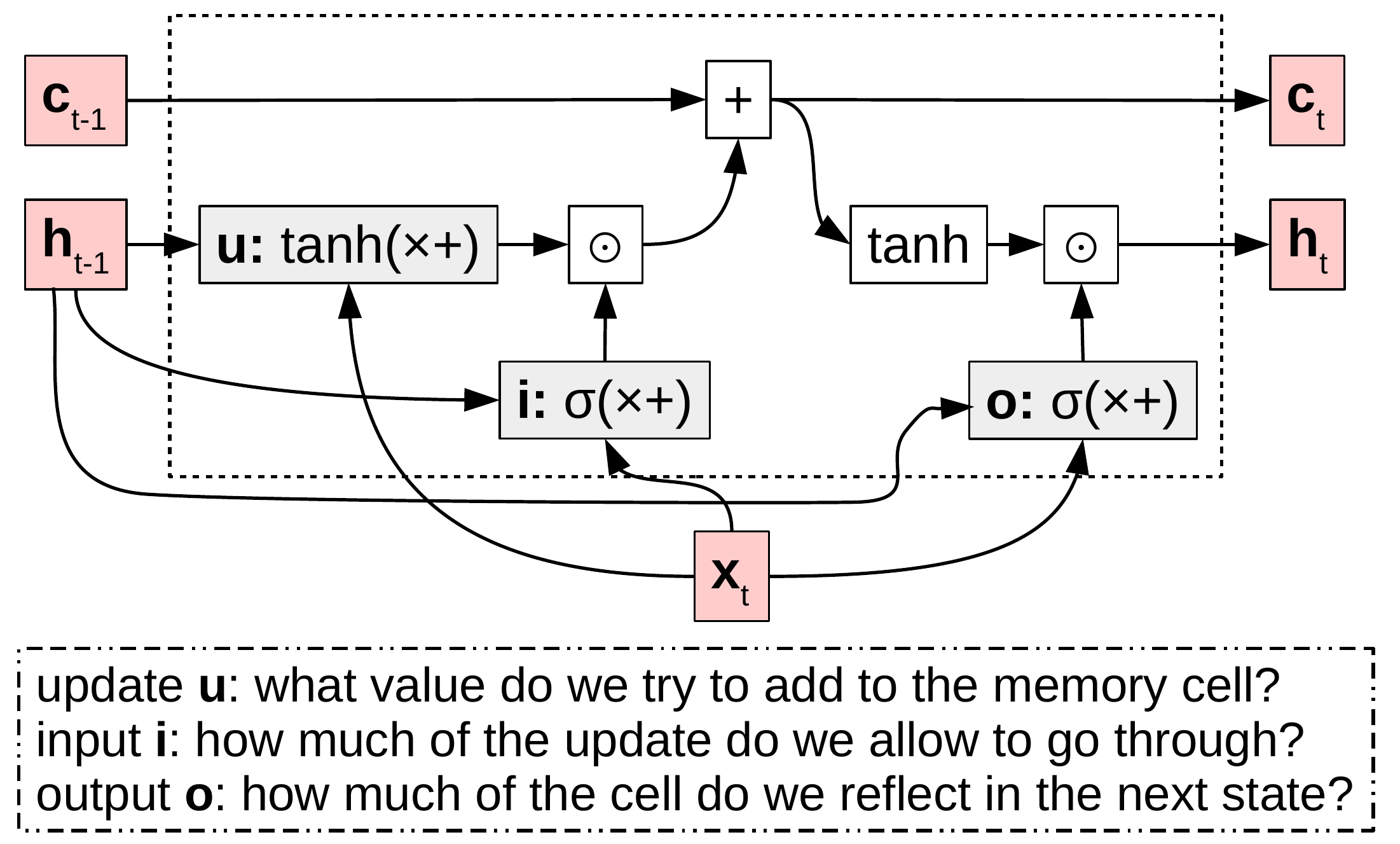}
 \caption{A single time step of long short-term memory (LSTM). The information flow between the $\bm{h}$ and cell $\bm{c}$ is modulated using parameterized input and output gates.}
 \label{fig:rnnlm:lstm}
\end{figure}

So how do LSTMs do this?
To understand this, let's take a look at the LSTM architecture in detail, as shown in \figref{rnnlm:lstm} and the following equations:
\begin{align}
\bm{u}_t & = \tanh(W_{xu} \bm{x}_t + W_{hu} h_{t-1} + \bm{b}_u) \label{eq:rnnlm:lstmupdate} \\
\bm{i}_t & = \sigma(W_{xi} \bm{x}_t + W_{hi} h_{t-1} + \bm{b}_i) \label{eq:rnnlm:lstminput} \\
% \bm{f}_t & = \sigma(W_{xf} \bm{x}_t + W_{hf} h_{t-1} + \bm{b}_f)   \nonumber \\
\bm{o}_t & = \sigma(W_{xo} \bm{x}_t + W_{ho} h_{t-1} + \bm{b}_o) \label{eq:rnnlm:lstmoutput} \\
\bm{c}_t & = \bm{i}_t \odot \bm{u}_t + \bm{c}_{t-1} \label{eq:rnnlm:lstmcell} \\
\bm{h}_t & = \bm{o}_t \odot \text{tanh}( \bm{c}_t ) \label{eq:rnnlm:lstmhidden}.
\end{align}
Taking the equations one at a time: \eqref{rnnlm:lstmupdate} is the update, which is basically the same as the RNN update in \eqref{rnnlm:rnn}; it takes in the input and hidden state, performs an affine transform and runs it through the tanh non-linearity.

\eqref{rnnlm:lstminput} and \eqref{rnnlm:lstmoutput} are the \term{input gate} and \term{output gate} of the LSTM respectively.
The function of ``gates'', as indicated by their name, is to either allow information to pass through or block it from passing.
Both of these gates perform an affine transform followed by the \term{sigmoid function}, also called the \term{logistic function}%
\footnote{
  To be more accurate, the sigmoid function is actually any mathematical function having an s-shaped curve, so the tanh function is also a type of sigmoid function.
  The logistic function is also a slightly broader class of functions $f(x) = \frac{L}{1+\exp(-k(x-x_0))}$.
  However, in the machine learning literature, the ``sigmoid'' is usually used to refer to the particular variety in \eqref{rnnlm:sigmoid}.
}
\begin{equation}
\sigma(x) = \frac{1}{1+\exp(-x)}, \label{eq:rnnlm:sigmoid}
\end{equation}
which squashes the input between 0 (which $\sigma(x)$ will approach as $x$ becomes more negative) and 1 (which $\sigma(x)$ will approach as $x$ becomes more positive).
The output of the sigmoid is then used to perform a componentwise multiplication
\begin{align}
\bm{z} & = \bm{x} \odot \bm{y} \nonumber \\
z_i & = x_i * y_i \nonumber
\end{align}
with the output of another function.
This results in the ``gating'' effect: if the result of the sigmoid is close to one for a particular vector position, it will have little effect on the input (the gate is ``open''), and if the result of the sigmoid is close to zero, it will block the input, setting the resulting value to zero (the gate is ``closed'').

\eqref{rnnlm:lstmcell} is the most important equation in the LSTM, as it is the equation that implements the intuition that $\frac{d\bm{c}_t}{d\bm{c}_{t-1}}$ must be equal to one, which allows us to conquer the vanishing gradient problem.
This equation sets $\bm{c}_t$ to be equal to the update $\bm{u}_t$ modulated by the input gate $\bm{i}_t$ plus the cell value for the previous time step $\bm{c}_{t-1}$.
Because we are directly adding $\bm{c}_{t-1}$ to $\bm{c}_t$, if we consider only this part of \eqref{rnnlm:lstmcell}, we can easily confirm that the gradient will indeed be one.%
\footnote{In actuality $\bm{i}_t \odot \bm{u}_t$ is also affected by $\bm{c}_{t-1}$, and thus $\frac{d\bm{c}_t}{d\bm{c}_{t-1}}$ is not exactly one, but the effect is relatively indirect.
Especially for vector elements with $\bm{i}_t$ close to zero, the effect will be minimal.}

Finally, \eqref{rnnlm:lstmhidden} calculates the next hidden state of the LSTM.
This is calculated by using a tanh function to scale the cell value between -1 and 1, then modulating the output using the output gate value $\bm{o}_t$.
This will be the value actually used in any downstream calculation, such as the calculation of language model probabilities.
\begin{equation}
\bm{p}_t = \text{softmax}(W_{hs} \bm{h}_t + b_s).
\end{equation}

\subsection{Other RNN Variants}
\label{sec:rnnlm:rnnvariants}

Because of the importance of recurrent neural networks in a number of applications, many variants of these networks exist. 
One modification to the standard LSTM that is used widely (in fact so widely that most people who refer to ``LSTM'' are now referring to this variant) is the addition of a \term{forget gate} \cite{gers2000learning}.
The equations for the LSTM with a forget gate are shown below:
\begin{align}
\bm{u}_t & = \tanh(W_{xu} \bm{x}_t + W_{hu} h_{t-1} + \bm{b}_u)   \nonumber \\
\bm{i}_t & = \sigma(W_{xi} \bm{x}_t + W_{hi} h_{t-1} + \bm{b}_i)   \nonumber \\
\bm{f}_t & = \sigma(W_{xf} \bm{x}_t + W_{hf} h_{t-1} + \bm{b}_f)   \label{eq:rnnlm:lstmforget} \\
\bm{o}_t & = \sigma(W_{xo} \bm{x}_t + W_{ho} h_{t-1} + \bm{b}_o)   \nonumber \\
\bm{c}_t & = \bm{i}_t \odot \bm{u}_t + \bm{f}_t \odot \bm{c}_{t-1} \label{eq:rnnlm:lstmcellwithforget} \\
\bm{h}_t & = \bm{o}_t \odot \text{tanh}(\bm{c}_t). \nonumber
\end{align}
Compared to the standard LSTM, there are two changes.
First, in \eqref{rnnlm:lstmforget}, we add an additional gate, the forget gate.
Second, in \eqref{rnnlm:lstmcellwithforget}, we use the gate to modulate the passing of the previous cell $\bm{c}_{t-1}$ to the current cell $\bm{c}_{t}$.
This forget gate is useful in that it allows the cell to easily clear its memory when justified: for example, let's say that the model has remembered that it has seen a particular word strongly correlated with another word, such as ``he'' and ``himself'' or ``she'' and ``herself'' in the example above.
In this case, we would probably like the model to remember ``he'' until it is used to predict ``himself'', then forget that information, as it is no longer relevant.
Forget gates have the advantage of allowing this sort of fine-grained information flow control, but they also come with the risk that if $\bm{f}_t$ is set to zero all the time, the model will forget everything and lose its ability to handle long-distance dependencies.
Thus, at the beginning of neural network training, it is common to initialize the bias $\bm{b}_f$ of the forget gate to be a somewhat large value (e.g. 1), which will make the neural net start training without using the forget gate, and only gradually start forgetting content after the net has been trained to some extent.

While the LSTM provides an effective solution to the vanishing gradient problem, it is also rather complicated (as many readers have undoubtedly been feeling).
One simpler RNN variant that has nonetheless proven effective is the \term{gated recurrent unit} (GRU; \cite{chung2014gru}), expressed in the following equations:
\begin{align}
\bm{r}_t & = \sigma(W_{xr} \bm{x}_t + W_{hr} h_{t-1} + \bm{b}_r)  \label{eq:rnnlm:grureset} \\
\bm{z}_t & = \sigma(W_{xz} \bm{x}_t + W_{hz} h_{t-1} + \bm{b}_z)  \label{eq:rnnlm:gruupdate} \\
\tilde{\bm{h}}_t & = \tanh(W_{xh} \bm{x}_t + W_{hh} (\bm{r}_t \odot \bm{h}_{t-1}) + \bm{b}_h) \label{eq:rnnlm:grucandidate} \\
\bm{h}_t & = (1 - \bm{z}_t) \bm{h}_{t-1} + \bm{z}_t \tilde{\bm{h}}_t. \label{eq:rnnlm:gruhidden}
\end{align}
The most characteristic element of the GRU is \eqref{rnnlm:gruhidden}, which interpolates between a candidate for the updated hidden state $\tilde{\bm{h}}_t$ and the previous state $\tilde{\bm{h}}_{t-1}$.
This interpolation is modulated by an \term{update gate} $\bm{z}_t$ (\eqref{rnnlm:gruupdate}), where if the update gate is close to one, the GRU will use the new candidate hidden value, and if the update is close to zero, it will use the previous value.
The candidate hidden state is calculated by \eqref{rnnlm:grucandidate}, which is similar to a standard RNN update but includes an additional modulation of the hidden state input by a \term{reset gate} $\bm{r}_t$ calculated in \eqref{rnnlm:grureset}.
Compared to the LSTM, the GRU has slightly fewer parameters (it performs one less parameterized affine transform) and also does not have a separate concept of a ``cell''. Thus, GRUs have been used by some to conserve memory or computation time.

\begin{figure}[h]
 \centering
 \includegraphics[width=11cm]{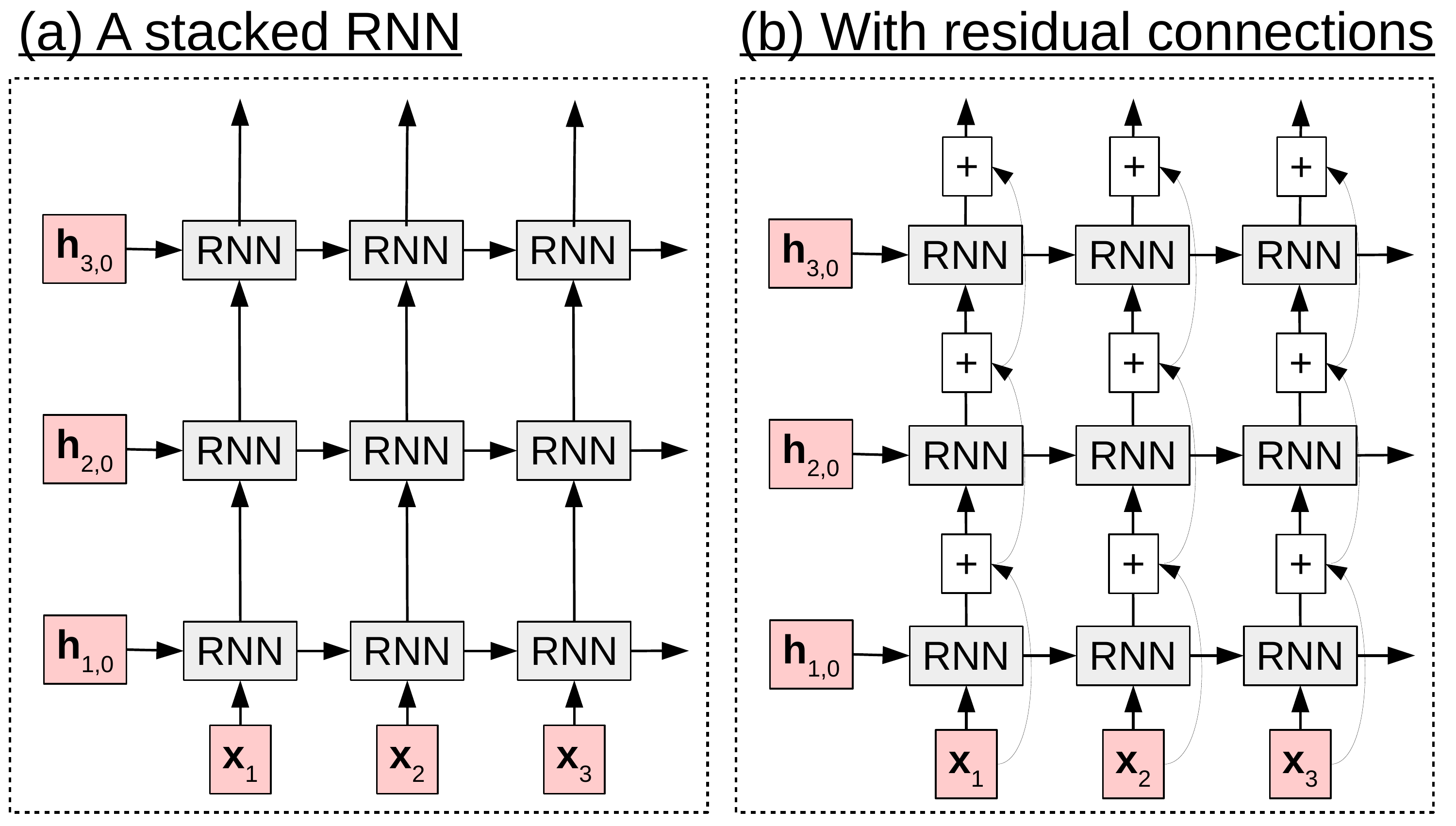}
 \caption{An example of (a) stacked RNNs and (b) stacked RNNs with residual connections.}
 \label{fig:rnnlm:stackedrnn}
\end{figure}

One other important modification we can do to RNNs, LSTMs, GRUs, or really any other neural network layer is simple but powerful: stack multiple layers on top of each other (\term{stacked RNNs} \figref{rnnlm:stackedrnn}(a)).
For example, in a 3-layer stacked RNN, the calculation at time step $t$ would look as follows:
\begin{align}
\bm{h}_{1,t} & = \text{RNN}_1(\bm{x}_t, \bm{h}_{1,t-1}) \nonumber \\
\bm{h}_{2,t} & = \text{RNN}_2(\bm{h}_{1,t}, \bm{h}_{2,t-1}) \nonumber \\
\bm{h}_{3,t} & = \text{RNN}_3(\bm{h}_{2,t}, \bm{h}_{3,t-1}), \nonumber
\end{align}
where $\bm{h}_{n,t}$ is the hidden state for the $n$th layer at time step $t$, and $\text{RNN}(\cdot)$ is an abbreviation for the RNN equation in \eqref{rnnlm:rnn}.
Similarly, we could substitute this function for $\text{LSTM}(\cdot)$, $\text{GRU}(\cdot)$, or any other recurrence step.
The reason why stacking multiple layers on top of each other is useful is for the same reason that non-linearities proved useful in the standard neural networks introduced in \secref{nnlm}: they are able to progressively extract more abstract features of the current words or sentences.
For example, \cite{shi2016nmtsyntax} find evidence that in a two-layer stacked LSTM, the first layer tends to learn granular features of words such as part of speech tags, while the second layer learns more abstract features of the sentence such as voice or tense.

While stacking RNNs has potential benefits, it also has the disadvantage that it suffers from the vanishing gradient problem in the vertical direction, just as the standard RNN did in the horizontal direction.
That is to say, the gradient will be back-propagated from the layer close to the output ($\text{RNN}_3$) to the layer close to the input ($\text{RNN}_1$), and the gradient may vanish in the process, causing the earlier layers of the network to be under-trained.
A simple solution to this problem, analogous to what the LSTM does for vanishing gradients over time, is \term{residual networks} (\figref{rnnlm:stackedrnn}(b)) \cite{he2016residual}.
The idea behind these networks is simply to add the output of the previous layer directly to the result of the next layer as follows:
\begin{align}
\bm{h}_{1,t} & = \text{RNN}_1(\bm{x}_t, \bm{h}_{1,t-1}) + \bm{x}_t \nonumber \\
\bm{h}_{2,t} & = \text{RNN}_2(\bm{h}_{1,t}, \bm{h}_{2,t-1}) + \bm{h}_{1,t} \nonumber \\
\bm{h}_{3,t} & = \text{RNN}_3(\bm{h}_{2,t}, \bm{h}_{3,t-1}) + \bm{h}_{2,t}. \nonumber
\end{align}
As a result, like the LSTM, there is no vanishing of gradients due to passing through the $\text{RNN}(\cdot)$ function, and even very deep networks can be learned effectively.

% Highway networks \cite{}.
% Peephole connections \cite{}.

\subsection{Online, Batch, and Minibatch Training}
\label{sec:rnnlm:minibatch}

As the observant reader may have noticed, the previous sections have gradually introduced more and more complicated models; we started with a simple linear model, added a hidden layer, added recurrence, added LSTM, and added more layers of LSTMs.
While these more expressive models have the ability to model with higher accuracy, they also come with a cost: largely expanded parameter space (causing more potential for overfitting) and more complicated operations (causing much greater potential computational cost).
This section describes an effective technique to improve the stability and computational efficiency of training these more complicated networks, \term{minibatching}.

Up until this point, we have used the stochastic gradient descent learning algorithm introduced in \secref{lllm:learning} that performs updates according to the following iterative process. This type of learning, which performs updates a single example at a time is called \term{online learning}.
\begin{algorithm}
\caption{A fully online training algorithm}
\label{alg:rnnlm:online}
\begin{algorithmic}[1]
\Procedure{Online}{}
\For{several epochs of training}
\For{each training example in the data}
\State Calculate gradients of the loss
\State Update the parameters according to this gradient
\EndFor
\EndFor
\EndProcedure
\end{algorithmic}
\end{algorithm}

In contrast, we can also think of a \term{batch learning} algorithm, which treats the entire data set as a single unit, calculates the gradients for this unit, then only performs update after making a full pass through the data.
\begin{algorithm}
\caption{A batch learning algorithm}
\label{alg:rnnlm:batch}
\begin{algorithmic}[1]
\Procedure{Batch}{}
\For{several epochs of training}
\For{each training example in the data}
\State Calculate and accumulate gradients of the loss
\EndFor
\State Update the parameters according to the accumulated gradient
\EndFor
\EndProcedure
\end{algorithmic}
\end{algorithm}

These two update strategies have trade-offs.
\begin{itemize}
\item Online training algorithms usually find a relatively good solution more quickly, as they don't need to make a full pass through the data before performing an update.
\item However, at the end of training, batch learning algorithms can be more stable, as they are not overly influenced by the most recently seen training examples.
\item Batch training algorithms are also more prone to falling into local optima; the randomness in online training algorithms often allows them to bounce out of local optima and find a better global solution.
\end{itemize}

Minibatching is a happy medium between these two strategies.
Basically, minibatched training is similar to online training, but instead of processing a single training example at a time, we calculate the gradient for $n$ training examples at a time.
In the extreme case of $n=1$, this is equivalent to standard online training, and in the other extreme where $n$ equals the size of the corpus, this is equivalent to fully batched training.
In the case of training language models, it is common to choose minibatches of $n=1$ to $n=128$ sentences to process at a single time.
As we increase the number of training examples, each parameter update becomes more informative and stable, but the amount of time to perform one update increases, so it is common to choose an $n$ that allows for a good balance between the two.

\begin{figure}
 \centering
 \includegraphics[width=12cm]{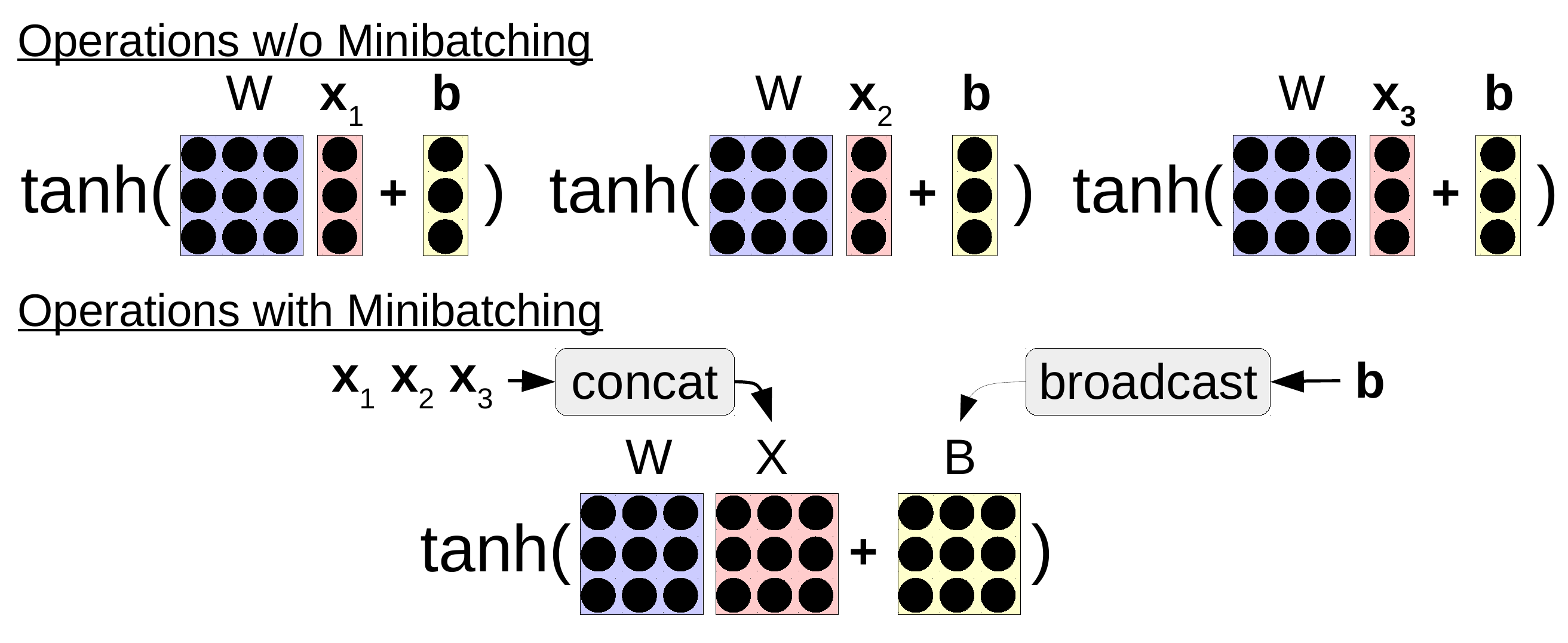}
 \caption{An example of combining multiple operations together when minibatching.}
 \label{fig:rnnlm:minibatching}
\end{figure}

One other major advantage of minibatching is that by using a few tricks, it is actually possible to make the simultaneous processing of $n$ training examples significantly faster than processing $n$ different examples separately.
Specifically, by taking multiple training examples and grouping similar operations together to be processed simultaneously, we can realize large gains in computational efficiency due to the fact that modern hardware (particularly GPUs, but also CPUs) have very efficient vector processing instructions that can be exploited with appropriately structured inputs.
As shown in \figref{rnnlm:minibatching}, common examples of this in neural networks include grouping together matrix-vector multiplies from multiple examples into a single matrix-matrix multiply or performing an element-wise operation (such as $\tanh$) over multiple vectors at the same time as opposed to processing single vectors individually.
Luckily, in DyNet, the library we are using, this is relatively easy to do, as much of the machinery for each elementary operation is handled automatically.
We'll give an example of the changes that we need to make when implementing an RNN language model below.

\begin{figure}[t!]
 \centering
 \includegraphics[width=10cm]{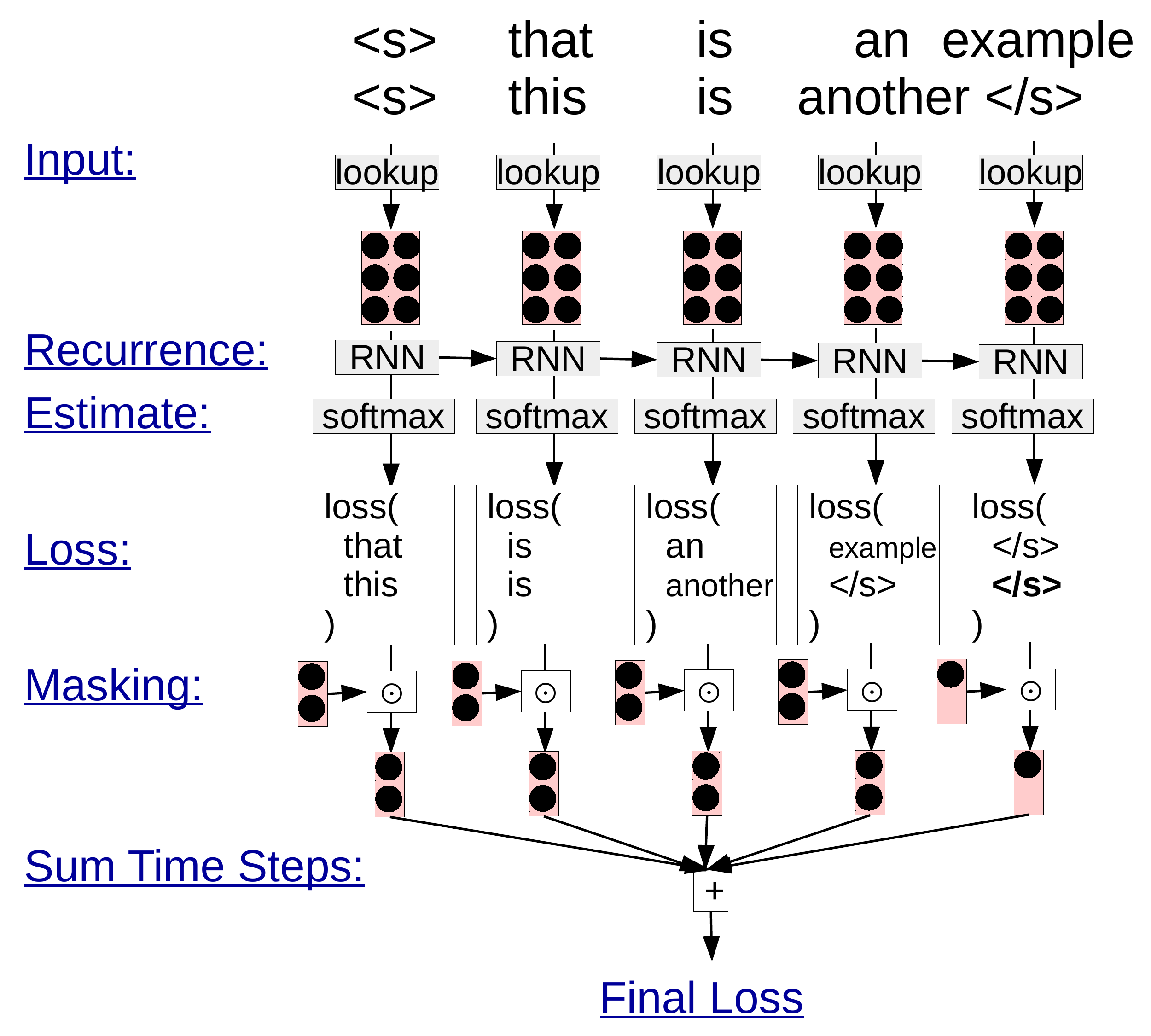}
 \caption{An example of minibatching in an RNN language model.}
 \label{fig:rnnlm:minibatchingsequences}
\end{figure}

The basic idea in the batched RNN language model (\figref{rnnlm:minibatchingsequences}) is that instead of processing a single sentence, we process multiple sentences at the same time.
So, instead of looking up a single word embedding, we look up multiple word embeddings (in DyNet, this is done by replacing the \texttt{lookup} function with the \texttt{lookup\_batch} function, where we pass in an array of word IDs instead of a single word ID).
We then run these batched word embeddings through the RNN and softmax as normal, resulting in two separate probability distributions over words in the first and second sentences.
We then calculate the loss for each word (again in DyNet, replacing the \texttt{pickneglogsoftmax} function with the \texttt{pickneglogsoftmax\_batch} function and pass word IDs).
We then sum together the losses and use this as the loss for our entire sentence.

One sticking point, however, is that we may need to create batches with sentences of different sizes, also shown in the figure.
In this case, it is common to perform \term{sentence padding} and \term{masking} to make sure that sentences of different lengths are treated properly.
Padding works by simply adding the ``end-of-sentence'' symbol to the shorter sentences until they are of the same length as the longest sentence in the batch.
Masking works by multiplying all loss functions calculated over these padded symbols by zero, ensuring that the losses for sentence end symbols don't get counted twice for the shorter sentences.

By taking these two measures, it becomes possible to process sentences of different lengths, but there is still a problem: if we perform lots of padding on sentences of vastly different lengths, we'll end up wasting a lot of computation on these padded symbols.
To fix this problem, it is also common to sort the sentences in the corpus by length before creating mini-batches to ensure that sentences in the same mini-batch are approximately the same size.

\subsection{Further Reading}
\label{sec:rnnlm:further}

Because of the prevalence of RNNs in a number of tasks both on natural language and other data, there is significant interest in extensions to them.
The following lists just a few other research topics that people are handling:

\begin{description}
\item[What can recurrent neural networks learn?:]
RNNs are surprisingly powerful tools for language, and thus many people have been interested in what exactly is going on inside them.
\cite{karpathy2015visualizing} demonstrate ways to visualize the internal states of LSTM networks, and find that some nodes are in charge of keeping track of length of sentences, whether a parenthesis has been opened, and other salietn features of sentences.
\cite{li2015visualizing} show ways to analyze and visualize which parts of the input are contributing to particular decisions made by an RNN-based model, by back-propagating information through the network.
\item[Other RNN architectures:]
There are also quite a few other recurrent network architectures.
\cite{greff15lstmsearchspace} perform an interesting study where they ablate various parts of the LSTM and attempt to find the best architecture for particular tasks.
\cite{zoph2016neural} take it a step further, explicitly training the model to find the best neural network architecture.
\end{description}

\subsection{Exercise}
\label{sec:rnnlm:exercise}

In the exercise for this chapter, we will construct a recurrent neural network language model using LSTMs.

Writing the program will entail:
\begin{itemize}
\item Writing a function such as \texttt{lstm\_step} or \texttt{gru\_step} that takes the input of the previous time step and updates it according to the appropriate equations. For reference, in DyNet, the componentwise multiply and sigmoid functions are \texttt{dy.cmult} and \texttt{dy.logistic} respectively.
\item Adding this function to the previous neural network language model and measuring the effect on the held-out set.
\item Ideally, implement mini-batch training by using the functionality implemented in DyNet, \texttt{lookup\_batch} and \texttt{pickneglogsoftmax\_batch}.
\end{itemize}
Language modeling accuracy should be measured in the same way as previous exercises and compared with the previous models.

Potential improvements to the model include:
Measuring the speed/stability improvements achieved by mini-batching.
Comparing the differences between recurrent architectures such as RNN, GRU, or LSTM.

  \section{Neural Encoder-Decoder Models}
  \label{sec:encdec}
  
From \secref{ngramlm} to \secref{rnnlm}, we focused on the language modeling problem of calculating the probability $P(E)$ of a sequence $E$.
In this section, we return to the statistical machine translation problem (mentioned in \secref{smt}) of modeling the probability $P(E \mid F)$ of the output $E$ given the input $F$.

\subsection{Encoder-decoder Models}
\label{sec:encdec:encdec}

The first model that we will cover is called an \term{encoder-decoder} model \cite{chrisman1991learning,forcada1997recursive,kalchbrenner13rnntm,sutskever14sequencetosequence}.
The basic idea of the model is relatively simple: we have an RNN language model, but before starting calculation of the probabilities of $E$, we first calculate the initial state of the language model using another RNN over the source sentence $F$.
The name ``encoder-decoder'' comes from the idea that the first neural network running over $F$ ``encodes'' its information as a vector of real-valued numbers (the hidden state), then the second neural network used to predict $E$ ``decodes'' this information into the target sentence.

\begin{figure}[h]
 \centering
 \includegraphics[width=12cm]{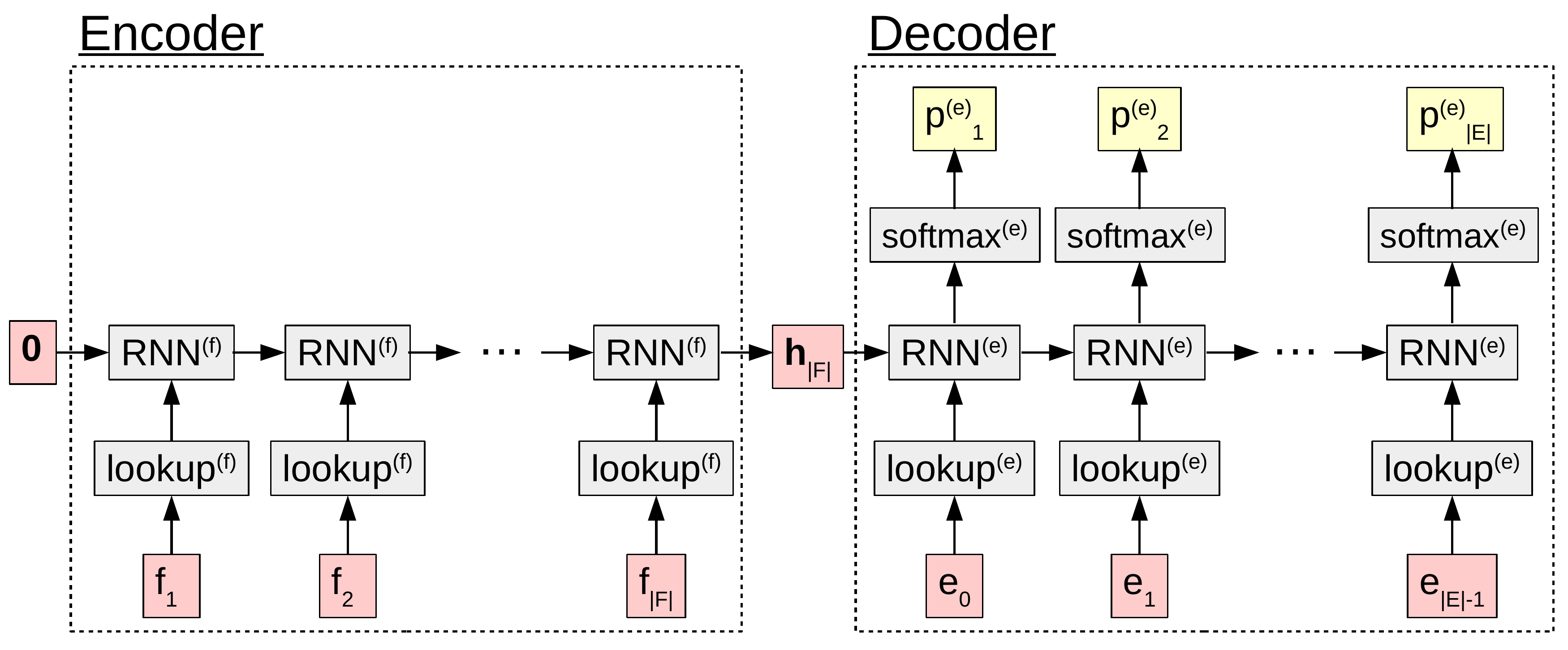}
 \caption{A computation graph of the encoder-decoder model.}
 \label{fig:encdec:encdec}
\end{figure}

If the encoder is expressed as $\text{RNN}^{(f)}(\cdot)$, the decoder is expressed as $\text{RNN}^{(e)}(\cdot)$, and we have a softmax that takes $\text{RNN}^{(e)}$'s hidden state at time step $t$ and turns it into a probability, then our model is expressed as follows (also shown in \figref{encdec:encdec}):
\begin{align}
\bm{m}^{(f)}_t & = M^{(f)}_{\cdot,f_t} \nonumber \\
\bm{h}^{(f)}_t & = \begin{cases}
  \text{RNN}^{(f)}(\bm{m}^{(f)}_t, \bm{h}^{(f)}_{t-1}) & t \ge 1, \\
  \bm{0} & \mbox{otherwise}. \\
\end{cases} \nonumber \\
\bm{m}^{(e)}_t & = M^{(e)}_{\cdot,e_{t-1}} \nonumber \\
\bm{h}^{(e)}_t & = \begin{cases}
  \text{RNN}^{(e)}(\bm{m}^{(e)}_t, \bm{h}^{(e)}_{t-1}) & t \ge 1, \\
  \bm{h}^{(f)}_{|F|} & \mbox{otherwise}. \\
\end{cases} \nonumber \\
\bm{p}^{(e)}_t & = \text{softmax}(W_{hs} \bm{h}^{(e)}_t + b_s)
\label{eq:encdec:softmax}
\end{align}
In the first two lines, we look up the embedding $\bm{m}^{(f)}_t$ and calculate the encoder hidden state $\bm{h}^{(f)}_t$ for the $t$th word in the source sequence $F$.
We start with an empty vector $\bm{h}^{(f)}_0 = \bm{0}$, and by $\bm{h}^{(f)}_{|F|}$, the encoder has seen all the words in the source sentence. Thus, this hidden state should theoretically be able to encode all of the information in the source sentence.

In the decoder phase, we predict the probability of word $e_t$ at each time step.
First, we similarly look up $\bm{m}^{(e)}_t$, but this time use the previous word $e_{t-1}$, as we must condition the probability of $e_t$ on the previous word, not on itself.
Then, we run the decoder to calculate $\bm{h}^{(e)}_t$.
This is very similar to the encoder step, with the important difference that $\bm{h}^{(e)}_0$ is set to the final state of the encoder $\bm{h}^{(f)}_{|F|}$, allowing us to condition on $F$.
Finally, we calculate the probability $\bm{p}^{(e)}_t$ by using a softmax on the hidden state $\bm{h}^{(e)}_t$.

While this model is quite simple (only 5 lines of equations), it gives us a straightforward and powerful way to model $P(E \mid F)$.
In fact, \cite{sutskever14sequencetosequence} have shown that a model that follows this basic pattern is able to perform translation with similar accuracy to heavily engineered systems specialized to the machine translation task (although it requires a few tricks over the simple encoder-decoder that we'll discuss in later sections: beam search (\secref{encdec:generation}), a different encoder (\secref{encdec:encoders}),
\iffullbook
and ensembling (\secref{systemcomb})).
\else
and ensembling (\secref{encdec:ensembling})).
\fi

\subsection{Generating Output}
\label{sec:encdec:generation}

At this point, we have only mentioned how to create a probability model $P(E \mid F)$ and haven't yet covered how to actually generate translations from it, which we will now cover in the next section.
In general, when we generate output we can do so according to several criteria:
\begin{description}
\item[Random Sampling:]
Randomly select an output $E$ from the probability distribution $P(E \mid F)$.
This is usually denoted $\hat{E} \sim P(E \mid F)$.
\item[1-best Search:]
Find the $E$ that maximizes $P(E \mid F)$, denoted $\hat{E} = \argmax{E} P(E \mid F)$.
\item[n-best Search:]
Find the $n$ outputs with the highest probabilities according to $P(E \mid F)$.
\end{description}
Which of these methods we will choose will depend on our application, so we will discuss some use cases along with the algorithms themselves.

\subsubsection{Random Sampling}

First, \term{random sampling} is useful in cases where we may want to get a variety of outputs for a particular input.
One example of a situation where this is useful would be in a sequence-to-sequence model for a dialog system, where we would prefer the system to not always give the same response to a particular user input to prevent monotony.
Luckily, in models like the encoder-decoder above, it is simple to exactly generate samples from the distribution $P(E \mid F)$ using a method called \term{ancestral sampling}.
Ancestral sampling works by sampling variable values one at a time, gradually conditioning on more context, so at time step $t$, we will sample a word from the distribution $P(e_t \mid \hat{e}_{1}^{t-1})$.
In the encoder-decoder model, this means we simply have to calculate $\bm{p}_t$ according to the previously sampled inputs, leading to the simple generation algorithm in \algref{encdec:sample}.

One thing to note is that sometimes we also want to know the probability of the sentence that we sampled.
For example, given a sentence $\hat{E}$ generated by the model, we might want to know how certain the model is in its prediction.
During the sampling process, we can calculate $P(\hat{E} \mid F) = \prod_{t}^{|\hat{E}|} P(\hat{e}_t \mid F,\hat{E}_1^{t-1})$ incrementally by stepping along and multiplying together the probabilities of each sampled word.
However, as we remember from the discussion of probability vs. log probability in \secref{ngramlm:eval}, using probabilities as-is can result in very small numbers that cause numerical precision problems on computers.
Thus, when calculating the full-sentence probability it is more common to instead add together log probabilities for each word, which avoids this problem.

\begin{algorithm}
\caption{Generating random samples from a neural encoder-decoder}
\label{alg:encdec:sample}
\begin{algorithmic}[1]
\Procedure{Sample}{}
\For{$t$ from 1 to $|F|$}
\State Calculate $\bm{m}^{(f)}_t$ and $\bm{h}^{(f)}_t$
\EndFor
\State Set $\hat{e}_0=$``$\sentbegin$'' and $t \leftarrow 0$
\While{$\hat{e}_t \ne $``$\sentend$''}
\State $t \leftarrow t + 1$
\State Calculate $\bm{m}^{(e)}_t$, $\bm{h}^{(e)}_t$, and $\bm{p}^{(e)}_t$ from $\hat{e}_{t-1}$
\State Sample $\hat{e}_t$ according to $\bm{p}^{(e)}_t$
\EndWhile
\EndProcedure
\end{algorithmic}
\end{algorithm}

\subsubsection{Greedy 1-best Search}

Next, let's consider the problem of generating a 1-best result.
This variety of generation is useful in machine translation, and most other applications where we simply want to output the translation that the model thought was best.
The simplest way of doing so is \term{greedy search}, in which we simply calculate $\bm{p}_t$ at every time step, select the word that gives us the highest probability, and use it as the next word in our sequence.
In other words, this algorithm is exactly the same as \algref{encdec:sample}, with the exception that on Line 9, instead of sampling $\hat{e}_t$ randomly according to $\bm{p}^{(e)}_t$, we instead choose the max: $\hat{e}_t = \argmax{i} p^{(e)}_{t,i}$.

\begin{figure}
 \centering
 \includegraphics[width=8cm]{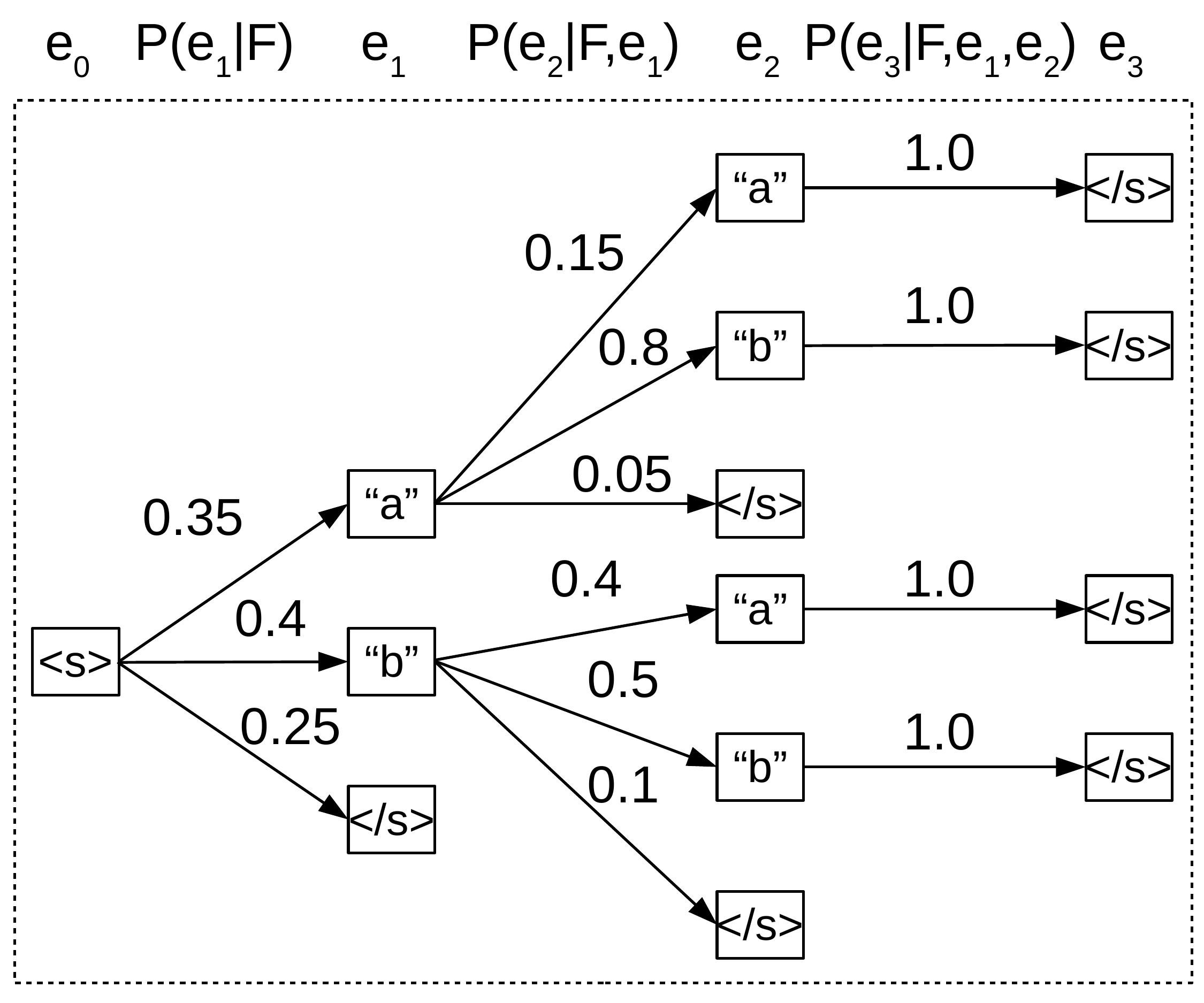}
 \caption{A search graph where greedy search fails.}
 \label{fig:encdec:searchgraph}
\end{figure}

Interestingly, while ancestral sampling exactly samples outputs from the distribution according to $P(E \mid F)$, greedy search is not guaranteed to find the translation with the highest probability.
An example of a case in which this is true can be found in the graph in \figref{encdec:searchgraph}, which is an example of search graph with a vocabulary of $\{\text{a}, \text{b}, \sentend\}$.\footnote{In reality, we will never have a probability of exactly $P(e_t=\sentend \mid F,e_1^{t-1})=1.0$, but for illustrative purposes, we show this here.}
As an exercise, I encourage readers to find the true 1-best (or $n$-best) sentence according to the probability $P(E \mid F)$ and the probability of the sentence found according to greedy search and confirm that these are different.

\begin{figure}
 \centering
 \includegraphics[width=8cm]{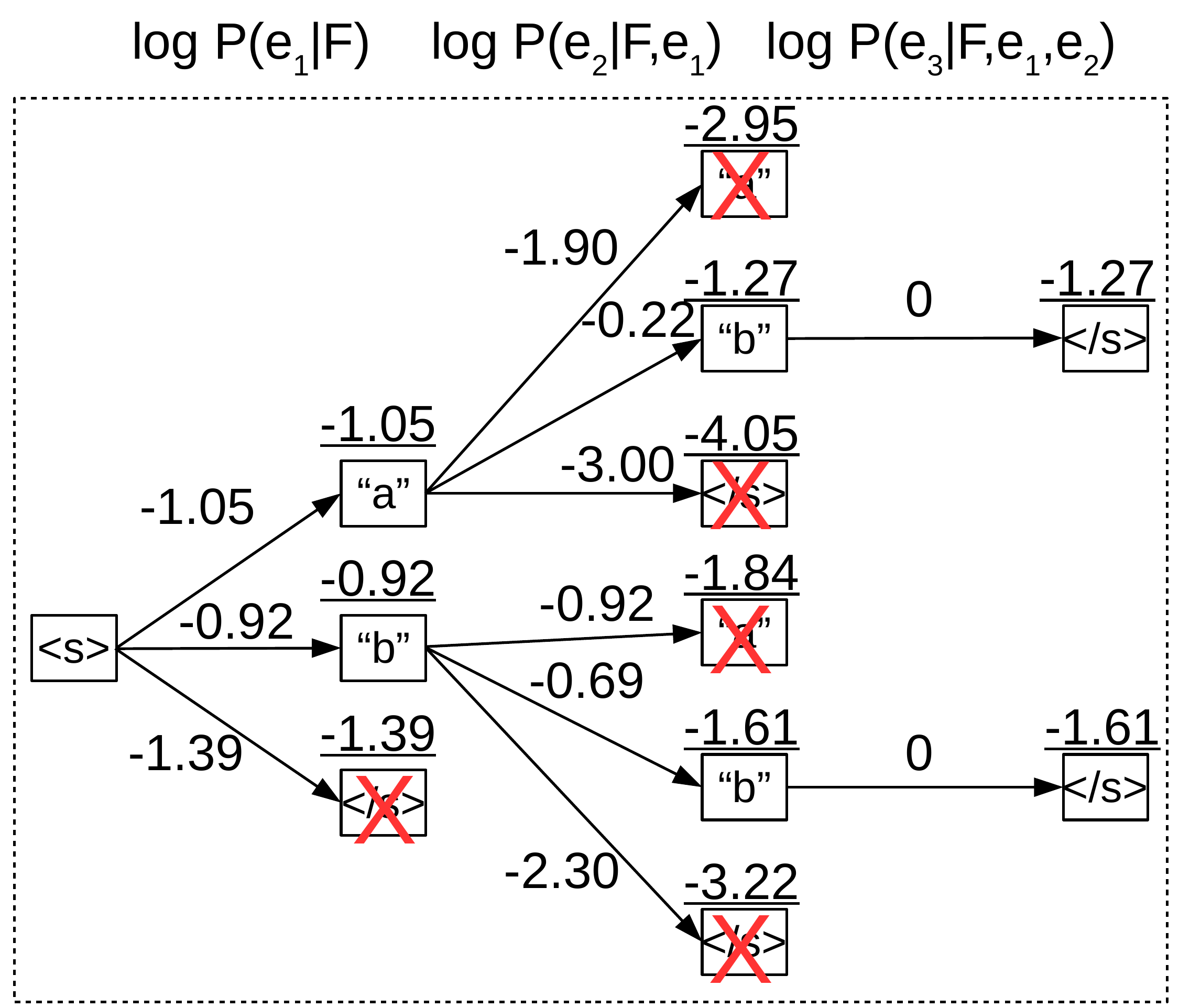}
 \caption{An example of beam search with $b=2$. Numbers next to arrows are log probabilities for a single word $\log P(e_t \mid F,e_1^{t-1})$, while numbers above nodes are log probabilities for the entire hypothesis up until this point.}
 \label{fig:encdec:beamsearch}
\end{figure}

\subsubsection{Beam Search}
\label{sec:encdec:beamsearch}

One way to solve this problem is through the use of \term{beam search}.
Beam search is similar to greedy search, but instead of considering only the one best hypothesis, we consider $b$ best hypotheses at each time step, where $b$ is the ``width'' of the beam.
An example of beam search where $b=2$ is shown in \figref{encdec:beamsearch} (note that we are using log probabilities here because they are more conducive to comparing hypotheses over the entire sentence, as mentioned before).
In the first time step, we expand hypotheses for $e_1$ corresponding to all of the three words in the vocabulary, then keep the top two (``b'' and ``a'') and delete the remaining one (``$\sentend$'').
In the second time step, we expand hypotheses for $e_2$ corresponding to the continuation of the first hypotheses for all words in the vocabulary, temporarily creating $b*|V|$ active hypotheses.
These active hypotheses are also pruned down to the $b$ active hypotheses (``a b'' and ``b b'').
This process of calculating scores for $b*|V|$ continuations of active hypotheses, then pruning back down to the top $b$, is continued until the end of the sentence.

One thing to be careful about when generating sentences using models, such as neural machine translation, where $P(E \mid F) = \prod_t^{|E|} P(e_t \mid F,e_1^{t-1})$ is that they tend to prefer shorter sentences.
This is because every time we add another word, we multiply in another probability, reducing the probability of the whole sentence.
As we increase the beam size, the search algorithm gets better at finding these short sentences, and as a result, beam search with a larger beam size often has a significant \term{length bias} towards these shorter sentences.

There have been several attempts to fix this length bias problem.
For example, it is possible to put a prior probability on the length of the sentence given the length of the previous sentence $P( |E| \mid |F| )$, and multiply this with the standard sentence probability $P(E \mid F)$ at decoding time \cite{eriguchi2016treetosequence}:
\begin{equation}
\hat{E} = \argmax{E} \log P(|E| \mid |F|) + \log P(E \mid F).
\end{equation}
This prior probability can be estimated from data, and \newcite{eriguchi2016treetosequence} simply estimate this using a multinomial distribution learned on the training data:
\begin{equation}
P(|E| \mid |F|) = \frac{c(|E|,|F|)}{c(|F|)}.
\end{equation}
A more heuristic but still widely used approach normalizes the log probability by the length of the target sentence, effectively searching for the sentence that has the highest average log probability per word \cite{cho14properties}:
\begin{equation}
\hat{E} = \argmax{E} \log P(E \mid F) / |E|.
\end{equation}

\subsection{Other Ways of Encoding Sequences}
\label{sec:encdec:encoders}

In \secref{encdec:encdec}, we described a model that works by encoding sequences linearly, one word at a time from left to right.
However, this may not be the most natural or effective way to turn the sentence $F$ into a vector $\bm{h}$.
In this section, we'll discuss a number of different ways to perform encoding that have been reported to be effective in the literature.

\subsubsection{Reverse and Bidirectional Encoders}

\begin{figure}
 \centering
 \includegraphics[width=8cm]{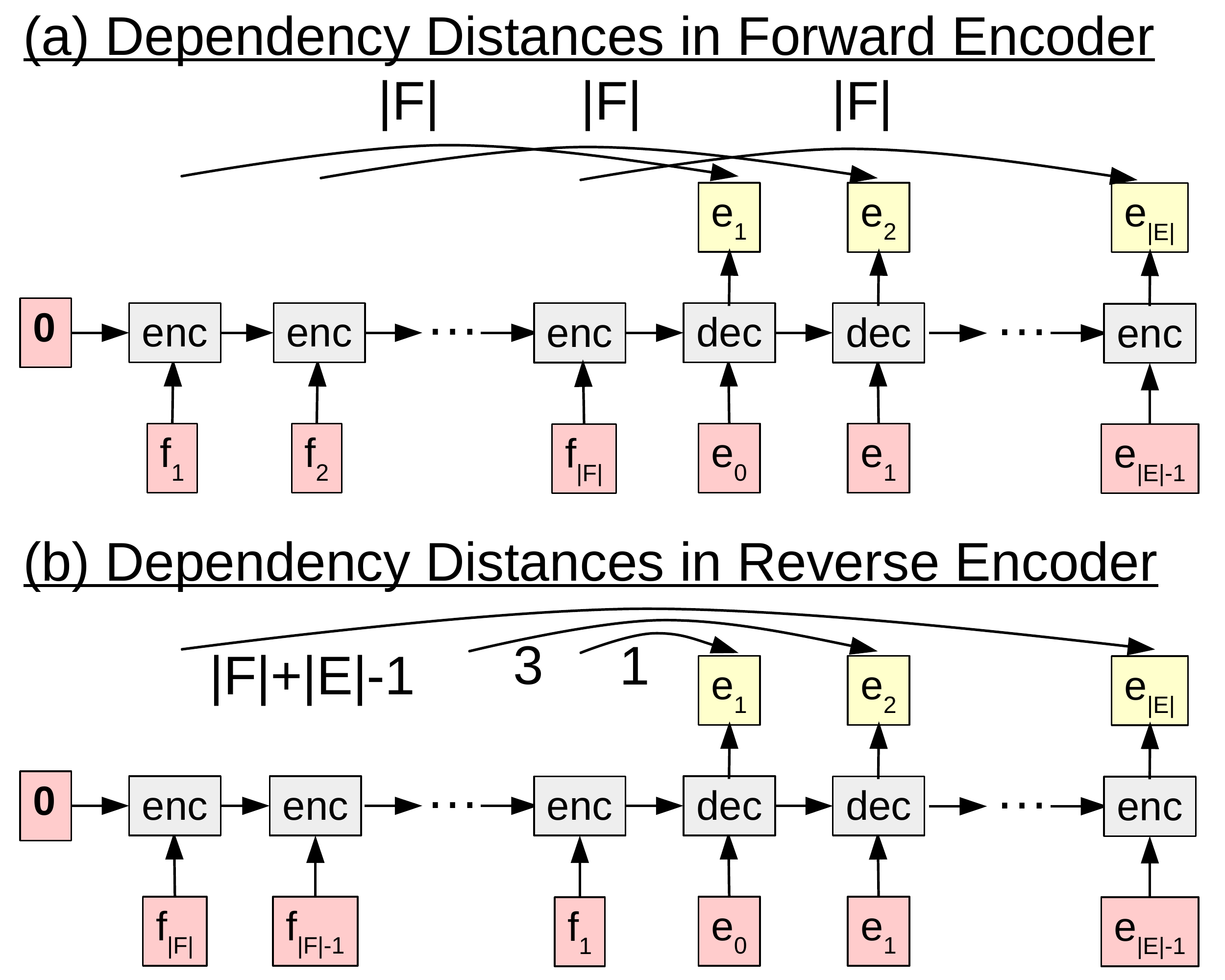}
 \caption{The distances between words with the same index in the forward and reverse decoders.}
 \label{fig:encdec:forvsrev}
\end{figure}

First, \cite{sutskever14sequencetosequence} have proposed a \term{reverse encoder}.
In this method, we simply run a standard linear encoder over $F$, but instead of doing so from left to right, we do so from right to left.
\begin{equation}
\overleftarrow{\bm{h}}^{(f)}_t = \begin{cases}
  \overleftarrow{\text{RNN}}^{(f)}(\bm{m}^{(f)}_t, \overleftarrow{\bm{h}}^{(f)}_{t+1}) & t \le |F|, \\
  \bm{0} & \mbox{otherwise}. \\
\end{cases}
\end{equation}
The motivation behind this method is that for pairs of languages with similar ordering (such as English-French, which the authors experimented on), the words at the beginning of $F$ will generally correspond to words at the beginning of $E$.
Assuming the extreme case that words with identical indices correspond to each-other (e.g. $f_1$ corresponds to $e_1$, $f_2$ to $e_2$, etc.), the distance between corresponding words in the linear encoding and decoding will be $|F|$, as shown in \figref{encdec:forvsrev}(a).
Remembering the vanishing gradient problem from \secref{rnnlm:lstm}, this means that the RNN has to propagate the information across $|F|$ time steps before making a prediction, a difficult feat.
At the beginning of training, even RNN variants such as LSTMs have trouble, as they have to essentially ``guess'' what part of the information encoded in their hidden state is being used without any prior bias.

Reversing the encoder helps solve this problem by reducing the length of dependencies for a subset of the words in the sentence, specifically the ones at the beginning of the sentences.
As shown in \figref{encdec:forvsrev}(b), the length of the dependency for $f_1$ and $e_1$ is 1, and subsequent pairs of $f_t$ and $e_t$ have a distance of $2t-1$.
During learning, the model can ``latch on'' to these short-distance dependencies and use them as a way to bootstrap the model training, after which it becomes possible to gradually learn the longer dependencies for the words at the end of the sentence.
In \cite{sutskever14sequencetosequence}, this proved critical to learn effective models in the encoder-decoder framework.

However, this approach of reversing the encoder relies on the strong assumption that the order of words in the input and output sequences are very similar, or at least that the words at the beginning of sentences are the same.
This is true for languages like English and French, which share the same ``subject-verb-object (SVO)'' word ordering, but may not be true for more typologically distinct languages.
One type of encoder that is slightly more robust to these differences is the \term{bi-directional encoder} \cite{bahdanau15alignandtranslate}. 
In this method, we use two different encoders: one traveling forward and one traveling backward over the input sentence
\begin{align}
\overrightarrow{\bm{h}}^{(f)}_t & = \begin{cases}
  \overrightarrow{\text{RNN}}^{(f)}(\bm{m}^{(f)}_t, \overrightarrow{\bm{h}}^{(f)}_{t+-}) & t \ge 1, \\
  \bm{0} & \mbox{otherwise}. \\
\end{cases} \\
\overleftarrow{\bm{h}}^{(f)}_t & = \begin{cases}
  \overleftarrow{\text{RNN}}^{(f)}(\bm{m}^{(f)}_t, \overleftarrow{\bm{h}}^{(f)}_{t+1}) & t \le |F|, \\
  \bm{0} & \mbox{otherwise}. \\
\end{cases}
\end{align}
which are then combined into the initial vector $\bm{h}^{(e)}_0$ for the decoder RNN.
This combination can be done by simply concatenating the two final vectors $\overrightarrow{\bm{h}}_{|F|}$ and $\overleftarrow{\bm{h}}_{1}$.
However, this also requires that the size of the vectors for the decoder RNN be exactly equal to the combined size of the two encoder RNNs.
As a more flexible alternative, we can add an additional parameterized hidden layer between the encoder and decoder states, which allows us to convert the bidirectional encoder states into an appropriately-sized state for the decoder:
\begin{equation}
\bm{h}^{(e)}_0 = \tanh(W_{\overrightarrow{f}e}\overrightarrow{\bm{h}}_{|F|} + W_{\overleftarrow{f}e}\overleftarrow{\bm{h}}_{1} + \bm{b}_e).
\end{equation}

\begin{figure}
 \centering
 \includegraphics[width=10cm]{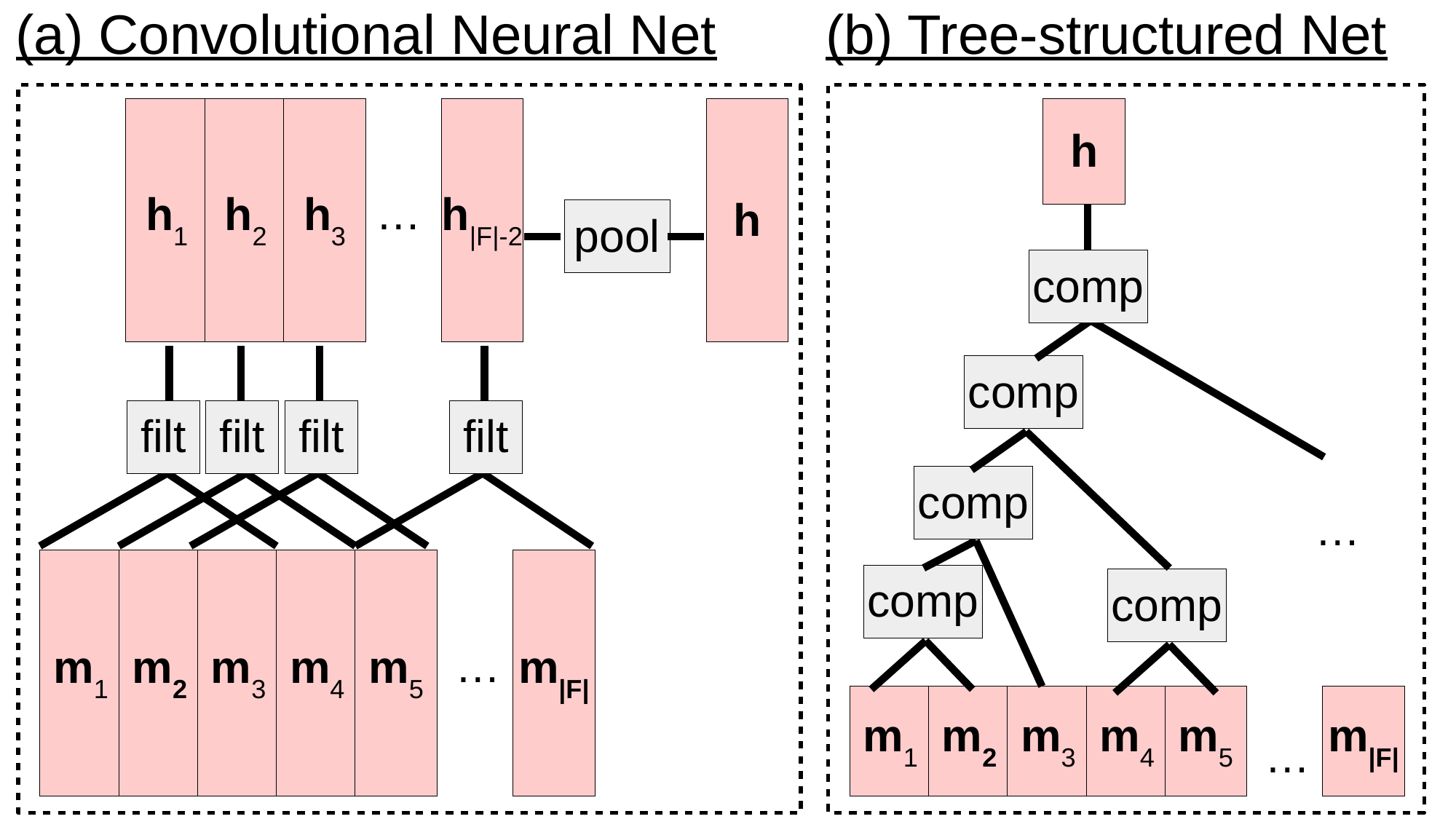}
 \caption{Examples of convolutional and tree-structured networks.}
 \label{fig:encdec:cnntree}
\end{figure}

\subsubsection{Convolutional Neural Networks}

In addition, there are also methods for decoding that move beyond a simple linear view of the input sentence.
For example, \term{convolutional neural networks} (CNNs; \cite{fukushima1988neocognitron,waibel1989phoneme,lecun1998gradient}), \figref{encdec:cnntree}(a)) are a variety of neural net that combines together information from spatially or temporally local segments.
They are most widely applied to image processing but have also been used for speech processing, as well as the processing of textual sequences.
While there are many varieties of CNN-based models of text (e.g. \cite{kalchbrenner14cnnsentence,lei2015moldingcnns,kalchbrenner2016neural}), here we will show an example from \newcite{kim2014cnntextcat}.
This model has $n$ \term{filters} with a width $w$ that are passed incrementally over $w$-word segments of the input.
Specifically, given an embedding matrix $M$ of width $|F|$, we generate a hidden layer matrix $H$ of width $|F|-w+1$, where each column of the matrix is equal to 
\begin{equation}
\bm{h}_t = W \text{concat}(\bm{m}_{t}, \bm{m}_{t+1}, \ldots, \bm{m}_{t+w-1})
\end{equation}
where $W \in \mathbb{R}^{n \times w|\bm{m}|}$ is a matrix where the $i$th row represents the parameters of filter $i$ that will be multiplied by the embeddings of $w$ consecutive words.
If $w=3$, we can interpret this as $\bm{h}_1$ extracting a vector of features for $f_1^3$, $\bm{h}_2$ as extracting a vector of features for $f_2^4$, etc. until the end of the sentence.

Finally, we perform a \term{pooling} operation that converts this matrix $H$ (which varies in width according to the sentence length) into a single vector $\bm{h}$ (which is fixed-size and can thus be used in down-stream processing).
Examples of pooling operations include average, max, and $k$-max \cite{kalchbrenner14cnnsentence}.

Compared to RNNs and their variants, CNNs have several advantages and disadvantages:
\begin{itemize}
\item On the positive side, CNNs provide a relatively simple way to detect features of short word sequences in sentence text and accumulate them across the entire sentence.
\item Also on the positive side, CNNs do not suffer as heavily from the vanishing gradient problem, as they do not need to propagate gradients across multiple time steps.
\item On the negative side, CNNs are not quite as expressive and are a less natural way of expressing complicated patterns that move beyond their filter width.
\end{itemize}
In general, CNNs have been found to be quite effective for text classification, where it is more important to pick out the most indicative features of the text and there is less of an emphasis on getting an overall view of the content \cite{kim2014cnntextcat}.
There have also been some positive results reported using specific varieties of CNNs for sequence-to-sequence modeling \cite{kalchbrenner2016neural}.

\begin{figure}
 \centering
 \includegraphics[width=7cm]{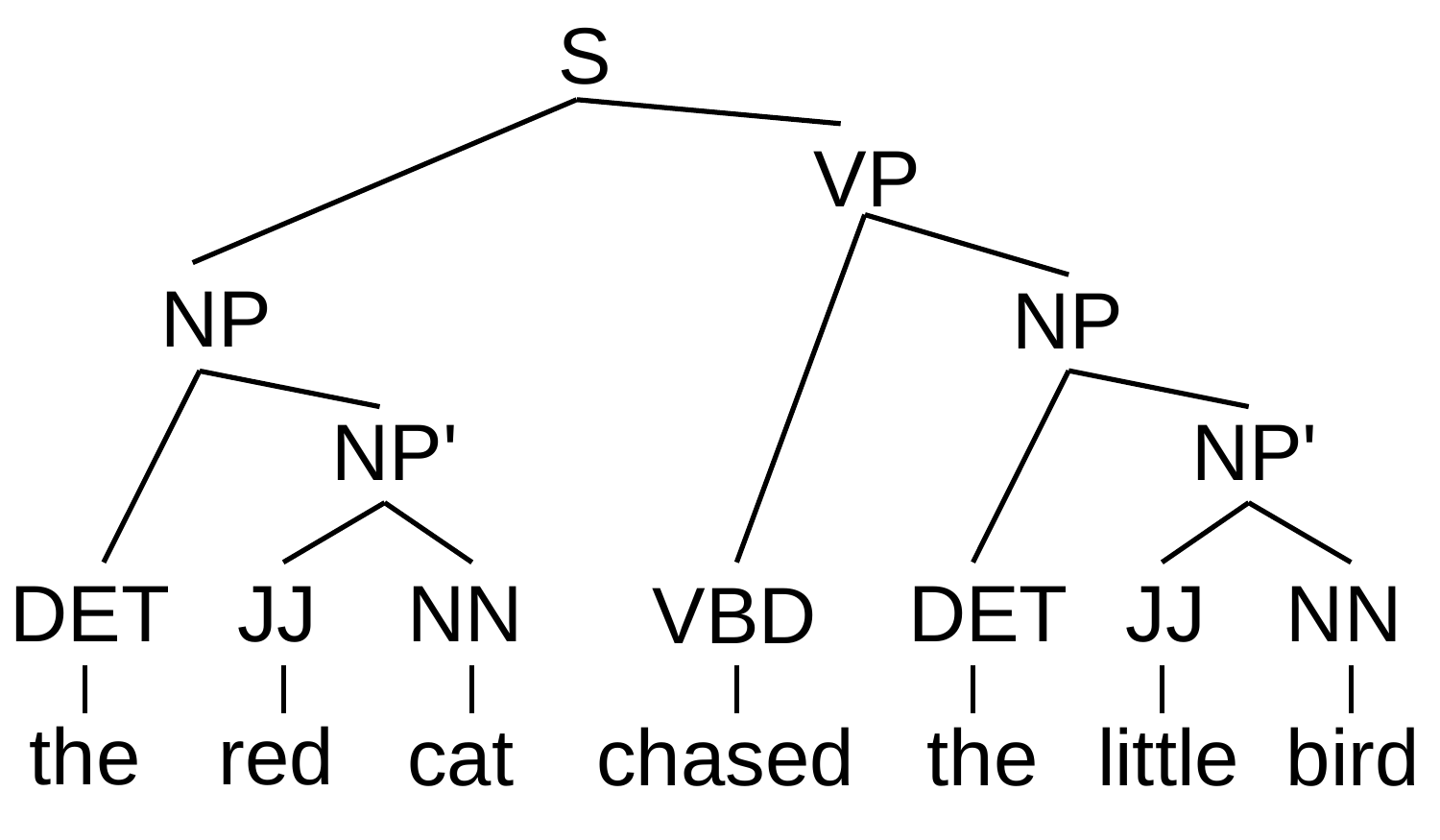}
 \caption{An example of a syntax tree for a sentence showing the sentence structure and phrase types (DET=``determiner'', JJ=``adjective'', NN=``noun'', VBD=``past tense verb'', NP=``noun phrase'', NP'=``part of a noun phrase'', VP=``verb phrase'', S=``sentence'').}
 \label{fig:encdec:parsetree}
\end{figure}

\subsubsection{Tree-structured Networks}

Finally, one other popular form of encoder that is widely used in a number of tasks are \term{tree-structured networks} (\cite{pollack1990recursive,socher11recursivenn}, \figref{encdec:cnntree}(b)).
The basic idea behind these networks is that the way to combine the information from each particular word is guided by some sort of structure, usually the syntactic structure of the sentence, an example of which is shown in \figref{encdec:parsetree}.
The reason why this is intuitively useful is because each syntactic phrase usually also corresponds to a coherent semantic unit. Thus, performing the calculation and manipulation of vectors over these coherent units will be more appropriate compared to using random substrings of words, like those used by CNNs.

For example, let's say we have the phrase ``the red cat chased the little bird'' as shown in the figure.
In this case, following a syntactic tree would ensure that we calculate vectors for coherent units that correspond to a grammatical phrase such as ``chased'' and ``the little bird'', and combine these phrases together one by one to obtain the meaning of larger coherent phrase such as ``chased the little bird''.
By doing so, we can take advantage of the fact that language is \term{compositional}, with the meaning of a more complex phrase resulting from regular combinations and transformation of smaller constituent phrases \cite{szabo2010compositionality}.
By taking this linguistically motivated and intuitive view of the sentence, we hope will help the neural networks learn more generalizable functions from limited training data.

Perhaps the most simple type of tree-structured network is the \term{recursive neural network} proposed by \newcite{socher11recursivenn}.
This network has very strong parallels to standard RNNs, but instead of calculating the hidden state $\bm{h}_t$ at time $t$ from the previous hidden state $\bm{h}_{t-1}$ as follows:
\begin{equation}
\bm{h}_t = \text{tanh}(W_{xh} \bm{x}_t + W_{hh} \bm{h}_{t-1} + \bm{b}_h),
\end{equation}
we instead calculate the hidden state of the parent node $\bm{h}_p$ from the hidden states of the left and right children, $\bm{h}_l$ and $\bm{h}_r$ respectively:
\begin{equation}
\bm{h}_p = \text{tanh}(W_{xp} \bm{x}_t + W_{lp} \bm{h}_{l} + W_{rp} \bm{h}_{r} + \bm{b}_p).
\end{equation}
Thus, the representation for each node in the tree can be calculated in a bottom-up fashion.

Like standard RNNs, these recursive networks suffer from the vanishing gradient problem.
To fix this problem there is an adaptation of LSTMs to tree-structured networks, fittingly called \term{tree LSTMs} \cite{tai15treelstm}, which fixes this vanishing gradient problem.
There are also a wide variety of other kinds of tree-structured composition functions that interested readers can explore \cite{socher13cvg,dyer2015stacklstm,dyer2016rnng}.
Also of interest is the study by \newcite{li2015treestructures}, which examines the various tasks in which tree structures are necessary or unnecessary for NLP.

\iffullbook
\else

\subsection{Ensembling Multiple Models}
\label{sec:encdec:ensembling}

One other method that is widely used in encoder-decoders, or other models of translation is \term{ensembling}: the combination of the prediction of multiple independently trained models to improve the overall prediction results.
The intuition behind ensembling is that different models will make different mistakes, and that on average it is more common for models to agree when the answer is correct than when it is mistaken.
Thus, if we combine multiple models together, it becomes possible to smooth over these mistakes, finding the correct answer more often.

The first step in ensembling encoder-decoder models is to independently train $N$ different models $P_1(\cdot), P_2(\cdot), \ldots, P_N(\cdot)$, for example, by randomly initializing the weights of the neural network differently before training.
Next, during search, at each time step we calculate the probability of the next word as the average probability of the $N$ models:
\begin{equation}
P(e_t \mid F, e_1^{t-1}) = \frac{1}{N} \sum_{i=1}^{N} P_i(e_t \mid F, e_1^{t-1}).
\end{equation}
This probability is used in searching for our hypotheses.

\fi

% \subsection{Further Reading}
% \label{sec:encdec:further}
% 
% TODO
% 
% Readers interested in history can take a look at
% 
% \cite{allen1987several}

\subsection{Exercise}
\label{sec:encdec:exercise}

In the exercise for this chapter, we will create an encoder-decoder translation model and make it possible to generate translations.

Writing the program will entail:
\begin{itemize}
\item Extend your RNN language model code to first read in a source sentence to calculate the initial hidden state.
\item On the training set, write code to calculate the loss function and perform training.
\item On the development set, generate translations using greedy search. 
\item Evaluate your generated translations by comparing them to the reference translations to see if they look good or not.
Translations can also be evaluated by automatic means, such as BLEU score \cite{papineni02bleu}.
A reference implementation of a BLEU evaluation script can be found here: \url{https://github.com/moses-smt/mosesdecoder/blob/master/scripts/generic/multi-bleu.perl}.
\end{itemize}

Potential improvements to the model include:
Implementing beam search and comparing the results with greedy search.
Implementing an alternative encoder.
Implementing ensembling.

  \section{Attentional Neural MT}
  \label{sec:attention}
  In the past chapter, we described a simple model for neural machine translation, which uses an encoder to encode sentences  as a fixed-length vector.
However, in some ways, this view is overly simplified, and by the introduction of a powerful mechanism called \term{attention}, we can overcome these difficulties.
This section describes the problems with the encoder-decoder architecture and what attention does to fix these problems.

\subsection{Problems of Representation in Encoder-Decoders}
\label{sec:attention:encdecproblems}

Theoretically, a sufficiently large and well-trained encoder-decoder model should be able to perform machine translation perfectly.
As mentioned in \secref{nnlm:nn}, neural networks are universal function approximators, meaning that they can express any function that we wish to model, including a function that accurately predicts our predictive probability for the next word $P(e_t \mid F, e_1^{t-1})$.
However, in practice, it is necessary to learn these functions from limited data, and when we do so, it is important to have a proper \term{inductive bias} -- an appropriate model structure that allows the network to learn to model accurately with a reasonable amount of data.

There are two things that are worrying about the standard encoder-decoder architecture.
The first was described in the previous section: there are long-distance dependencies between words that need to be translated into each other.
In the previous section, this was alleviated to some extent by reversing the direction of the encoder to bootstrap training, but still, a large number of long-distance dependencies remain, and it is hard to guarantee that we will learn to handle these properly.

The second, and perhaps more, worrying aspect of the encoder-decoder is that it attempts to store information sentences of any arbitrary length in a hidden vector of fixed size.
In other words, even if our machine translation system is expected to translate sentences of lengths from 1 word to 100 words, it will still use the same intermediate representation to store all of the information about the input sentence.
If our network is too small, it will not be able to encode all of the information in the longer sentences that we will be expected to translate.
On the other hand, even if we make the network large enough to handle the largest sentences in our inputs, when processing shorter sentences, this may be overkill, using needlessly large amounts of memory and computation time.
In addition, because these networks will have large numbers of parameters, it will be more difficult to learn them in the face of limited data without encountering problems such as overfitting.

The remainder of this section discusses a more natural way to solve the translation problem with neural networks: attention.

\subsection{Attention}
\label{sec:attention:attention}

The basic idea of attention is that instead of attempting to learn a single vector representation for each sentence, we instead keep around vectors for every word in the input sentence, and reference these vectors at each decoding step.
Because the number of vectors available to reference is equivalent to the number of words in the input sentence, long sentences will have many vectors and short sentences will have few vectors.
As a result, we can express input sentences in a much more efficient way, avoiding the problems of inefficient representations for encoder-decoders mentioned in the previous section.

First we create a set of vectors that we will be using as this variably-lengthed representation.
To do so, we calculate a vector for every word in the source sentence by running an RNN in both directions:
\begin{eqnarray*}
%    \label{eq:encode}
    \overrightarrow{\bm{h}}^{(f)}_j = \text{RNN}( \text{embed}(f_j), \overrightarrow{\bm{h}}^{(f)}_{j-1} ) \\
    \overleftarrow{\bm{h}}^{(f)}_j = \text{RNN}( \text{embed}(f_j), \overleftarrow{\bm{h}}^{(f)}_{j+1} ).
\end{eqnarray*}
Then we concatenate the two vectors $\overrightarrow{\bm{h}}^{(f)}_j$ and $\overleftarrow{\bm{h}}^{(f)}_j$ into a bidirectional representation $\bm{h}^{(f)}_j$
\begin{equation*}
  \bm{h}^{(f)}_j = [\overleftarrow{\bm{h}}^{(f)}_j; \overrightarrow{\bm{h}}^{(f)}_j].
\end{equation*}
We can further concatenate these vectors into a matrix:
\begin{equation*}
  H^{(f)} = \text{concat\_col}(\bm{h}^{(f)}_1,\ldots,\bm{h}^{(f)}_{|F|}).
\end{equation*}
This will give us a matrix where every column corresponds to one word in the input sentence.

\begin{figure}[t]
 \centering
 \includegraphics[width=8cm]{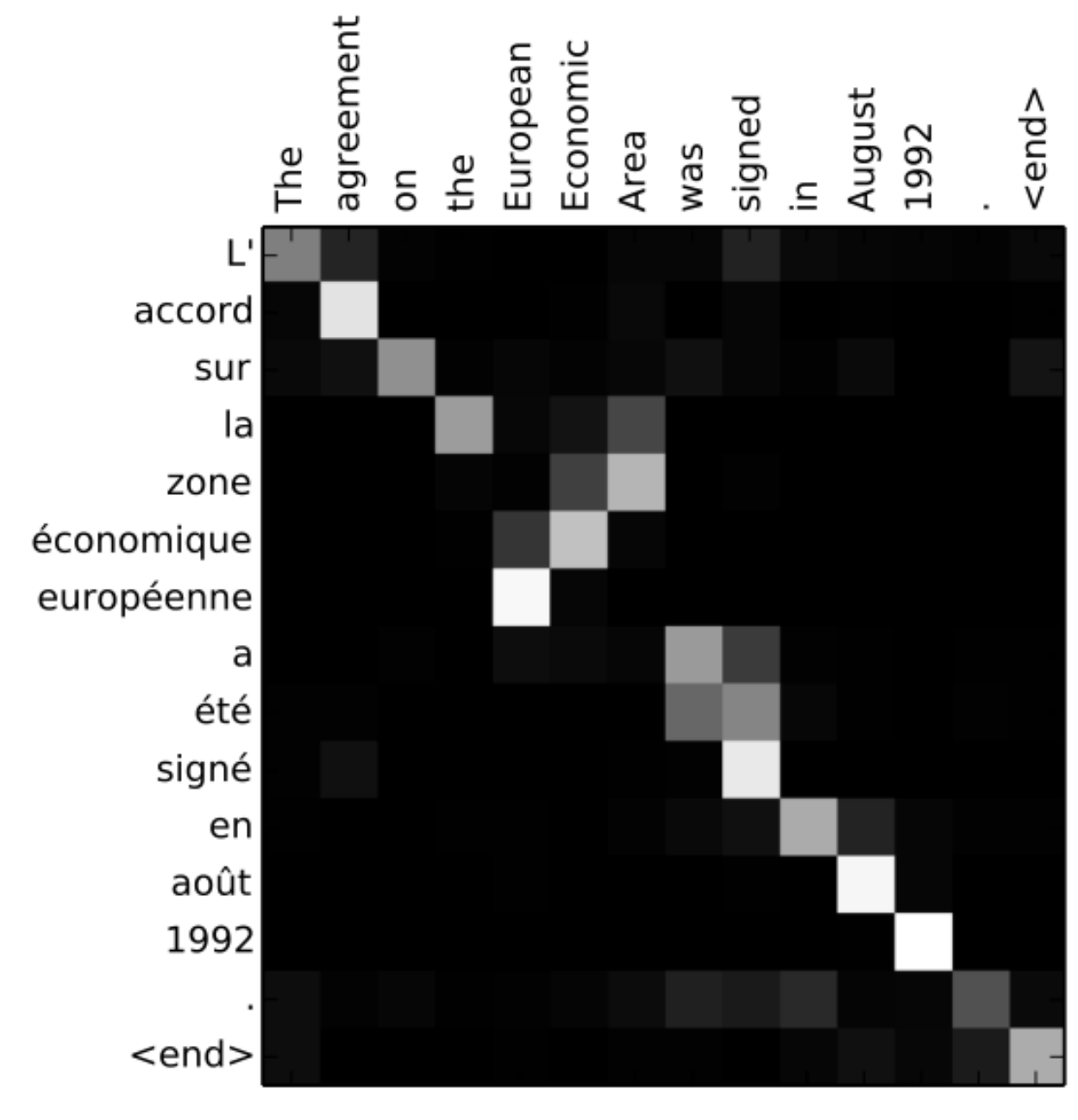}
 \caption{An example of attention from \cite{bahdanau15alignandtranslate}. English is the source, French is the target, and a higher attention weight when generating a particular target word is indicated by a lighter color in the matrix.}
 \label{fig:attention:attentionexample}
\end{figure}

However, we are now faced with a difficulty.
We have a matrix $H^{(f)}$ with a variable number of columns depending on the length of the source sentence, but would like to use this to compute, for example, the probabilities over the output vocabulary, which we only know how to do (directly) for the case where we have a vector of input.
The key insight of attention is that we calculate a vector $\bm{\alpha}_t$ that can be used to combine together the columns of $H$ into a vector $\bm{c}_t$
\begin{equation}
  \bm{c}_t = H^{(f)} \bm{\alpha}_t. 
  \label{eq:attention:context}
\end{equation}
$\bm{\alpha}_t$ is called the \term{attention vector}, and is generally assumed to have elements that are between zero and one and add to one.

The basic idea behind the attention vector is that it is telling us how much we are ``focusing'' on a particular source word at a particular time step.
The larger the value in $\bm{\alpha}_t$, the more impact a word will have when predicting the next word in the output sentence.
An example of how this attention plays out in an actual translation example is shown in \figref{attention:attentionexample}, and as we can see the values in the alignment vectors generally align with our intuition.

\subsection{Calculating Attention Scores}

The next question then becomes, from where do we get this $\bm{\alpha}_t$?
The answer to this lies in the \textit{decoder} RNN, which we use to track our state while we are generating output.
As before, the decoder's hidden state $\bm{h}^{(e)}_t$ is a fixed-length continuous vector representing the previous target words $e^{t-1}_1$, initialized as $\bm{h}^{(e)}_0 = \bm{h}^{(f)}_{\lvert F \rvert + 1}$.
This is used to calculate a context vector $\bm{c}_t$ that is used to summarize the source attentional context used in choosing target word $e_t$, and initialized as $\bm{c}_0 = \bm{0}$.

First, we update the hidden state to $\bm{h}^{(e)}_t$ based on the word representation and context vectors from the previous target time step
\begin{equation}
    \bm{h}^{(e)}_t = \text{enc}( [\text{embed}(e_{t-1}); \bm{c}_{t-1}], \bm{h}^{(e)}_{t-1} ).
\end{equation}

Based on this $\bm{h}^{(e)}_t$, we calculate an \term{attention score} $\bm{a}_t$, with each element equal to
\begin{equation}
a_{t,j} = \text{attn\_score}(\bm{h}^{(f)}_j,\bm{h}^{(e)}_t). \label{eq:attention:attnscore}
\end{equation}
$\text{attn\_score}(\cdot)$ can be an arbitrary function that takes two vectors as input and outputs a score about how much we should focus on this particular input word encoding $\bm{h}^{(f)}_j$ at the time step $\bm{h}^{(e)}_t$.
We describe some examples at a later point in \secref{attention:types}.

We then normalize this into the actual attention vector itself by taking a softmax over the scores:
\begin{equation}
\label{eq:attention:softmax}
\bm{\alpha}_t = \text{softmax}(\bm{a}_t).
\end{equation}
This attention vector is then used to weight the encoded representation $H^{(f)}$ to create a context vector $\bm{c}_t$ for the current time step, as mentioned in \eqref{attention:context}.

We now have a context vector $\bm{c}_t$ and hidden state $\bm{h}^{(e)}_t$ for time step $t$, which we can pass on down to downstream tasks.
For example, we can concatenate both of these together when calculating the softmax distribution over the next words:
\begin{equation}
\bm{p}^{(e)}_t = \text{softmax}(W_{hs} [ \bm{h}^{(e)}_t; \bm{c}_t ] + b_s).
\end{equation}
It is worth noting that this means that the encoding of each source word $\bm{h}^{(f)}_j$ is considered much more directly in the calculation of output probabilities.
In contrast to the encoder-decoder, where the encoder-decoder will only be able to access information about the first encoded word in the source by passing it over $|F|$ time steps, here the source encoding is accessed (in a weighted manner) through the context vector \eqref{attention:context}.

\begin{figure}[t]
 \centering
 \includegraphics[width=10cm]{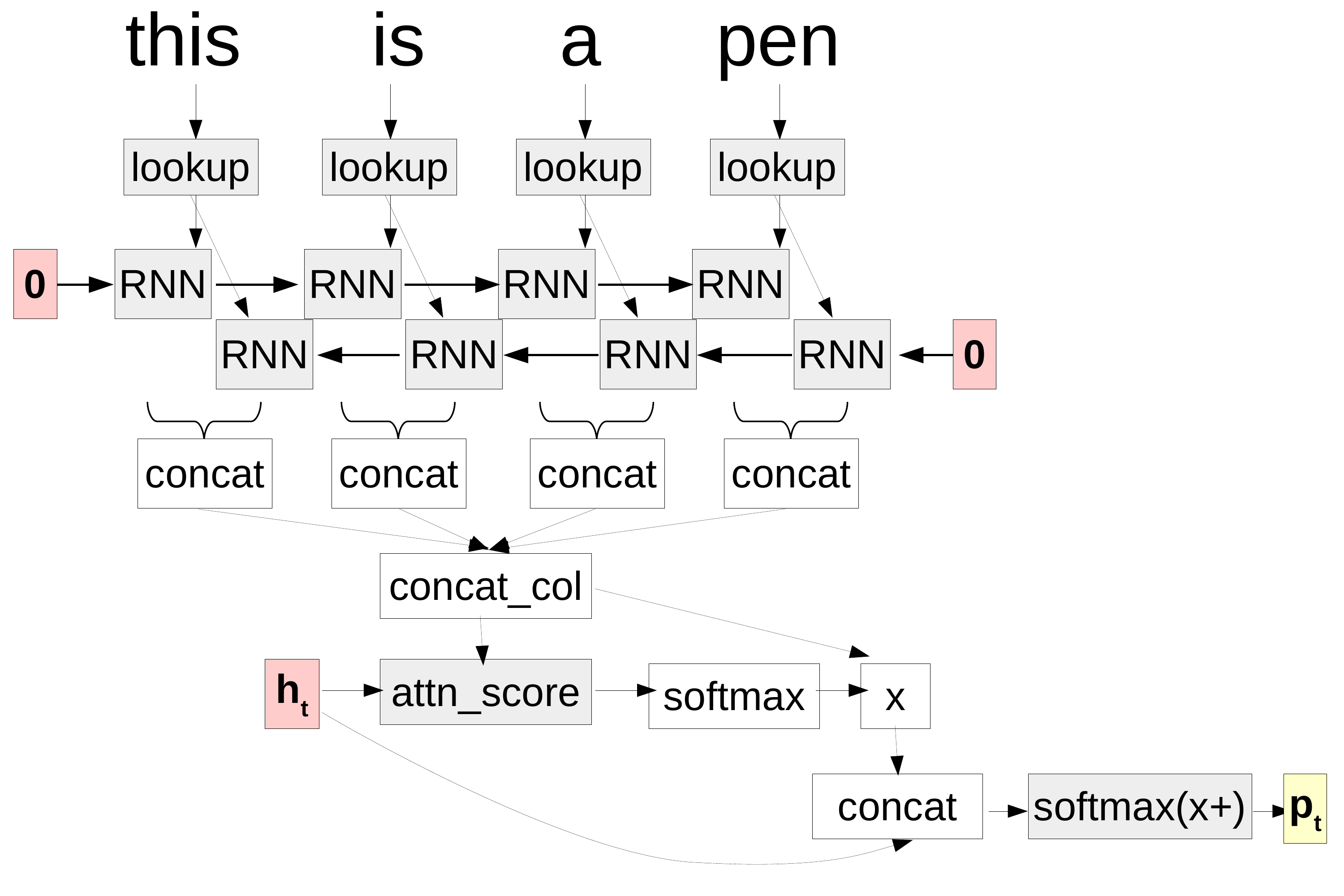}
 \caption{A computation graph for attention.}
 \label{fig:attention:attention}
\end{figure}

This whole, rather involved, process is shown in \figref{attention:attention}.

\subsection{Ways of Calculating Attention Scores}
\label{sec:attention:types}

As mentioned in \eqref{attention:attnscore}, the final missing piece to the puzzle is how to calculate the attention score $a_{t,j}$.

\newcite{luong15effectiveattentional} test three different attention functions, all of which have their own merits:
\begin{description}
\item[Dot product:]
This is the simplest of the functions, as it simply calculates the similarity between $\bm{h}^{(e)}_t$ and $\bm{h}^{(f)}_j$ as measured by the dot product:
\begin{equation}
\text{attn\_score}(\bm{h}^{(f)}_j, \bm{h}^{(e)}_t) := \bm{h}^{(f)\intercal}_j \bm{h}^{(e)}_t.
\end{equation}
This model has the advantage that it adds no additional parameters to the model.
However, it also has the intuitive disadvantage that it forces the input and output encodings to be in the same space (because similar $\bm{h}^{(e)}_t$ and $\bm{h}^{(f)}_j$ must be close in space in order for their dot product to be high).
It should also be noted that the dot product can be calculated efficiently for every word in the source sentence by instead defining the attention score over the concatenated matrix $H^{(f)}$ as follows:
\begin{equation}
\text{attn\_score}(H^{(f)}, \bm{h}^{(e)}_t) := H^{(f)\intercal}_j \bm{h}^{(e)}_t.
\end{equation}
Combining the many attention operations into one can be useful for efficient impementation, especially on GPUs.
The following attention functions can also be calculated like this similarly.\question{What do the equations look like for the combined versions of the following functions?}
\item[Bilinear functions:]
One slight modification to the dot product that is more expressive is the \term{bilinear function}.
This function helps relax the restriction that the source and target embeddings must be in the same space by performing a linear transform parameterized by $W_a$ before taking the dot product:
\begin{equation}
\text{attn\_score}(\bm{h}^{(f)}_j, \bm{h}^{(e)}_t) := \bm{h}^{(f)\intercal}_j W_a \bm{h}^{(e)}_t.
\end{equation}
This has the advantage that if $W_a$ is not a square matrix, it is possible for the two vectors to be of different sizes, so it is possible for the encoder and decoder to have different dimensions.
However, it does introduce quite a few parameters to the model ($|\bm{h}^{(f)}| \times |\bm{h}^{(e)}|$), which may be difficult to train properly.
\item[Multi-layer perceptrons:]
Finally, it is also possible to calculate the attention score using a multi-layer perceptron, which was the method employed by \newcite{bahdanau15alignandtranslate} in their original implementation of attention:
\begin{equation}
\text{attn\_score}(\bm{h}^{(e)}_t,\bm{h}^{(f)}_j) := \bm{w}_{a2}^\intercal \text{tanh}(W_{a1}[\bm{h}^{(e)}_t; \bm{h}^{(f)}_j]),
\end{equation}
where $W_{a1}$ and $\bm{w}_{a2}$ are the weight matrix and vector of the first and second layers of the MLP respectively.
This is more flexible than the dot product method, usually has fewer parameters than the bilinear method, and generally provides good results.
\end{description}

In addition to these methods above, which are essentially the defacto-standard, there are a few more sophisticated methods for calculating attention as well.
For example, it is possible to use recurrent neural networks \cite{yang2016recurrentattention},  tree-structured networks based on document structure \cite{yang2016hierarchicalattention}, convolutional neural networks \cite{allamanis2016convolutional}, or structured models \cite{kim2017structuredattention} to calculate attention.

\subsection{Copying and Unknown Word Replacement}
\label{sec:attention:unk}

One pleasant side-effect of attention is that it not only increases translation accuracy, but also makes it easier to tell which words are translated into which words in the output.
One obvious consequence of this is that we can draw intuitive graphs such as the one shown in \figref{attention:attentionexample}, which aid error analysis.

Another advantage is that it also becomes possible to handle unknown words in a more elegant way, performing \term{unknown word replacement} \cite{luong15rareword}.
The idea of this method is simple, every time our decoder chooses the unknown word token $\sentunk$ in the output, we look up the source word with the highest attention weight at this time step, and output that word instead of the unknown token $\sentunk$.
If we do so, at least the user can see which words have been left untranslated, which is better than seeing them disappear altogether or be replaced by a placeholder.

\iffullbook
It is also common to use alignment models such as those in \secref{ibmmodels} to obtain a translation dictionary, then use this to aid unknown word replacement even further.
\else
It is also common to use alignment models such as those described in \newcite{brown93smt} to obtain a translation dictionary, then use this to aid unknown word replacement even further.
\fi
Specifically, instead of copying the word as-is into the output, if the chosen source word is $f$, we output the word with the highest translation probability $P_{\text{dict}}(e \mid f)$.
This allows words that are included in the dictionary to be mapped into their most-frequent counterpart in the target language.

\subsection{Intuitive Priors on Attention}
\label{sec:attention:improvements}

Because of the importance of attention in modern NMT systems, there have also been a number of proposals to improve accuracy of estimating the attention itself through the introduction of intuitively motivated prior probabilities.
\newcite{cohn2016structural} propose several methods to incorporate biases into the training of the model to ensure that the attention weights match our belief of what alignments between languages look like.

\iffullbook
These take several forms, and are heavily inspired by the alignment models that will be explained in \secref{ibmmodels}.
\else
These take several forms, and are heavily inspired by the alignment models used in more traditional SMT systems such as those proposed by \newcite{brown93smt}.
\fi
These models can be briefly summarized as:
\begin{description}
\item[Position Bias:] If two languages have similar word order, then it is more likely that alignments should fall along the diagonal. This is demonstrated strongly in \figref{attention:attentionexample}. It is possible to encourage this behavior by adding a prior probability over attention that makes it easier for things near the diagonal to be aligned.
\item[Markov Condition:] In most languages, we can assume that most of the time if two words in the target are contiguous, the aligned words in the source will also be contiguous. For example, in \figref{attention:attentionexample}, this is true for all contiguous pairs of English words except ``the, European'' and ``Area, was''. To take advantage of this property, it is possible to impose a prior that discourages large jumps and encourages local steps in attention. A model that is similar in motivation, but different in implementation, is the \term{local attention} model \cite{luong15effectiveattentional}, which selects which part of the source sentence to focus on using the neural network itself.
\item[Fertility:] We can assume that some words will be translated into a certain number words in the other langauge. For example, the English word ``cats'' will be translated into two words ``les chats'' in French. Priors on fertility takes advantage of this fact by giving the model a penalty when particular words are not attended too much, or attended to too much. In fact one of the major problems with poorly trained neural MT systems is that they repeat the same word over and over, or drop words, a violation of this fertility constraint. Because of this, several other methods have been proposed to incorporate coverage in the model itself \cite{tu2016coverage,mi2016coverage}, or as a constraint during the decoding process \cite{wu2016google}.
\item[Bilingual Symmetry:] Finally, we expect that words that are aligned when performing translation from $F$ to $E$ should also be aligned when performing translation from $E$ to $F$. This can be enforced by training two models in parallel, and enforcing constraints that the alignment matrices look similar in both directions.
\end{description}
\newcite{cohn2016structural} experiment extensively with these approaches, and find that the bilingual symmetry constraint is particularly effective among the various methods.

\subsection{Further Reading}
\label{sec:attention:further}

This section outlines some further directions for reading more about improvements to attention:

\begin{description}
\item[Hard Attention:] As shown in \eqref{attention:softmax}, standard attention uses a soft combination of various contents. There are also methods for hard attention that make a hard binary decision about whether to focus on a particular context, with motivations ranging from learning explainable models \cite{lei2016rationalizing}, to processing text incrementally \cite{yu2016online,gu2017learning}.
\item[Supervised Training of Attention:] In addition, sometimes we have hand-annotated data showing us true alignments for a particular language pair. It is possible to train attentional models using this data by defining a loss function that penalizes the model when it does not predict these alignments correctly \cite{mi2016supervisedattention}.
\item[Other Ways of Memorizing Input:] Finally, there are other ways of accessing relevant information other than attention. \newcite{wang2016memory} propose a method using \term{memory networks}, which have a separate set of memory that can be written to or read from as the processing continues.
% \cite{zhang2016variational}
\end{description}

\subsection{Exercise}
\label{sec:attention:exercise}

In the exercise for this chapter, we will create code to train and generate translations with an attentional neural MT model.

Writing the program will entail extending your encoder-decoder code to add attention.
You can then generate translations and compare them to others.

\begin{itemize}
\item Extend your encoder-decoder code to add attention.
\item On the training set, write code to calculate the loss function and perform training.
\item On the development set, generate translations using greedy search. 
\item Evaluate these translations, either manually or automatically.
\end{itemize}
It is also highly recommended, but not necessary, that you attempt to implement unknown word replacement.

Potential improvements to the model include implementing any of the improvements to attention mentioned in \secref{attention:improvements} or \secref{attention:further}.

  \section{Conclusion}
  \label{sec:seq2seqend}
  This tutorial has covered the basics of neural machine translation and sequence-to-sequence models.
It gradually stepped through models of increasing sophistication, starting with $n$-gram language models, and culminating in attention, which now represents the state-of-the-art in many sequence-to-sequence modeling tasks.

It should be noted that this is a very active reserach field, and there are a number of advanced research topics that are beyond the scope of this tutorial, but may be of interest to readers who have mastered the basics and would like to learn more.
\begin{description}
\item[Handling large vocabularies:]
One difficulty of neural MT models is that they perform badly when using large vocabularies; it is hard to learn how to properly translate rare words with limited data, and computation becomes a burden.
One method to handle this is to break words into smaller units such as characters \cite{chung16character} or subwords \cite{sennrich16bpe}.
It is also possible to incorporate translation dictionaries with broad coverage to handle low-frequency phenomena \cite{arthur16emnlp}.
\item[Optimizing translation performance:] 
While the models presented in this tutorial are trained to maximize the likelihood of the target sentence given the source $P(E \mid F)$, in reality what we actually care about is the accuracy of the generated sentences.
There have been a number of works proposed to resolve this disconnect by directly considering the accuracy of the generated results when training our models.
These include methods that sample translation results from the current model and move towards parameters that result in good translations \cite{ranzato16sequence,shen16minrisk}, methods that optimize parameters towards partially mistaken hypotheses to try to improve robustness to mistakes in generation \cite{bengio2015scheduled,norouzi2016reward}, or methods that try to prevent mistakes that may occur during the search process \cite{wiseman16bso}.
\item[Multi-lingual learning:]
Up until now we assumed that we were training a model between two languages $F$ and $E$.
However, in reality there are many languages in the world, and some work has shown that we can benefit by using data from all these languages to learn models together \cite{firat16multiway,johnson16multilingual,ha2016toward}.
It is also possible to perform transfer across languages, training a model first on one language pair, then fine-tuning it to others \cite{zoph2016transfer}.
\item[Other applications:] Similar sequence-to-sequence models have been used for a wide variety of tasks, from dialog systems \cite{vinyals2015neural,shang2015neuralresponding} to text summarization \cite{rush2015neuralattention}, speech recognition \cite{chan2016listen}, speech synthesis \cite{van2016wavenet}, image captioning \cite{karpathy2015deep,vinyals2015show}, image generation \cite{gregor2015draw}, and more.
\end{description}

This is just a small sampling of topics from this exciting and rapidly expanding field, and hopefully this tutorial gave readers the tools to strike out on their own and apply these models to their applications of interest.

\subsection*{Acknowledgements}
I am extremely grateful to Qinlan Shen and Dongyeop Kang for their careful reading of these materials and useful comments about unclear parts. I also thank the students in the Machine Translation and Sequence-to-sequence Models class at CMU for pointing out various bugs in the materials when a preliminary version was used in the class.

\bibliographystyle{plain}
\bibliography{bib/myplain,bib/gneubig}

% \appendixfile{Quiz Questions}{900}{quiz}

\end{document}